\documentclass{article}

\usepackage{arxiv}          
\usepackage[utf8]{inputenc} 
\usepackage[T1]{fontenc}    
\usepackage{url}            
\usepackage{booktabs}       
\usepackage{amsfonts}       
\usepackage{nicefrac}       
\usepackage{microtype}      
\usepackage{lipsum}         
\usepackage{graphicx}
\usepackage{amsmath, amssymb}
\usepackage{algorithmic}
\usepackage{textcomp}
\usepackage{multirow}
\usepackage{subcaption}
\usepackage[numbers]{natbib}
\usepackage{doi}
\usepackage{xcolor}
\usepackage{hyperref}       

\title{IDA-UIE: An Iterative Framework for Deep Network-based Degradation Aware Underwater Image Enhancement}

\author{ \href{https://orcid.org/0000-0001-6854-9743}{\includegraphics[scale=0.06]{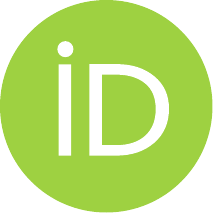}\hspace{1mm}Pranjali Singh}\\
	Centre for Intelligent Cyber Physical Systems, Indian Institute of Technology Guwahati,
	 \\
	\texttt{s.pranjali@iitg.ac.in} \\
	\And
	\href{https://orcid.org/0000-0003-2885-0026}{\includegraphics[scale=0.06]{orcid.pdf}\hspace{1mm}Prithwijit Guha} \\
	Dept. of Electronics and Electrical Engg., Indian Institute of Technology Guwahati,
 \\
	\texttt{pguha@iitg.ac.in} \\
}

\renewcommand{\shorttitle}{IDA-UIE}

\hypersetup{
pdftitle={A template for the arxiv style},
pdfsubject={q-bio.NC, q-bio.QM},
pdfauthor={David S.~Hippocampus, Elias D.~Striatum},
pdfkeywords={First keyword, Second keyword, More},
}

\begin{document}
\maketitle

\begin{abstract}
	Underwater image quality is affected by fluorescence, low illumination, absorption, and scattering. Recent works in underwater image enhancement have proposed different deep network architectures to handle these problems. Most of these works have proposed a single network to handle all the challenges. We believe that deep networks trained for specific conditions deliver better performance than a single network learned from all degradation cases. Accordingly, the first contribution of this work lies in the proposal of an iterative framework where a single dominant degradation condition is identified and resolved. This proposal considers the following eight degradation conditions -- low illumination, low contrast, haziness, blurred image, presence of noise and color imbalance in three different channels. A deep network is designed to identify the dominant degradation condition. Accordingly, an appropriate deep network is selected for degradation condition-specific enhancement. The second contribution of this work is the construction of degradation condition specific datasets from good quality images of two standard datasets (UIEB and EUVP). This dataset is used to learn the condition specific enhancement networks. The proposed approach is found to outperform nine baseline methods on UIEB and EUVP datasets.
\end{abstract}


\section{Introduction}
Poor visibility conditions in the world's oceans have limited our understanding of these environments. To address this challenge, underwater image enhancement techniques are employed \cite{sahu2014survey}. With approximately 70\% of the Earth's surface covered by water, there is increasing interest in exploring underwater realms. Clear images are essential for monitoring marine species, underwater mountains, and plants. Additionally, the effects of color in underwater images are significant. Light reflection varies greatly depending on the sea's structure, with water capable of bending light to create crinkle patterns or diffusing it. The quality of underwater photos is influenced by several factors, including restricted visibility range, uneven lighting, unwanted noise, and reduced color fidelity \cite{raveendran2021underwater}.

\subsection{Application}
Underwater image enhancement has numerous practical applications in various fields, including oceanography, underwater archaeology, underwater robotics, underwater exploration, and more \cite{r9}. Some specific applications are outlined below:

\begin{enumerate}
    \item Marine Life: Underwater image enhancement aids in the identification and tracking of marine life, such as fish, corals, and other organisms. This is crucial for scientific research on the health and behavior of underwater ecosystems \cite{raveendran2021underwater}.
    \item Oceanography: Enhanced underwater images improve the study of ocean currents, tides, and underwater topography.
    \item Underwater Archaeology: Enhancing images of submerged structures, like shipwrecks, assists in identifying and studying historical artifacts and structures.
    \item Underwater Security and Surveillance: The accuracy and effectiveness of security and surveillance systems in underwater environments are enhanced through underwater image enhancement. This aids in detecting and tracking intruders, suspicious objects, and potential threats to underwater infrastructure, such as pipelines and oil rigs \cite{raveendran2021underwater}.
    \item Underwater Robotics: Underwater robots, such as remotely operated vehicles (ROVs) and autonomous underwater vehicles (AUVs), are equipped with cameras and sensors for object detection and navigation. Underwater image enhancement improves the quality of images captured by these sensors, facilitating the detection and tracking of marine life, underwater structures, and potential hazards.
    \item Underwater Photography and Videography: Enhancing the quality of underwater images and videos makes them more appealing to audiences and enhances the immersive experience. This is particularly important for promoting dive sites and other underwater attractions.
    \item Underwater Mapping and Navigation: Image enhancement increases the accuracy and detail of maps and navigation systems used for underwater exploration and research, aiding in the discovery and exploration of new dive sites and other underwater environments.
    \item Underwater Tourism through Virtual Reality: Enhancing the quality of images and videos used for virtual reality (VR) experiences provides a more immersive and realistic experience for users, enabling safe and realistic exploration of underwater environments \cite{hasibuan2021contrast}.
\end{enumerate}
Improving the quality of underwater images can significantly impact our understanding of the underwater world and enhance our ability to explore and interact with it \cite{raveendran2021underwater}.

\begin{figure}[!h]
    \centering
    \includegraphics[width=10 cm,height=5cm,keepaspectratio]{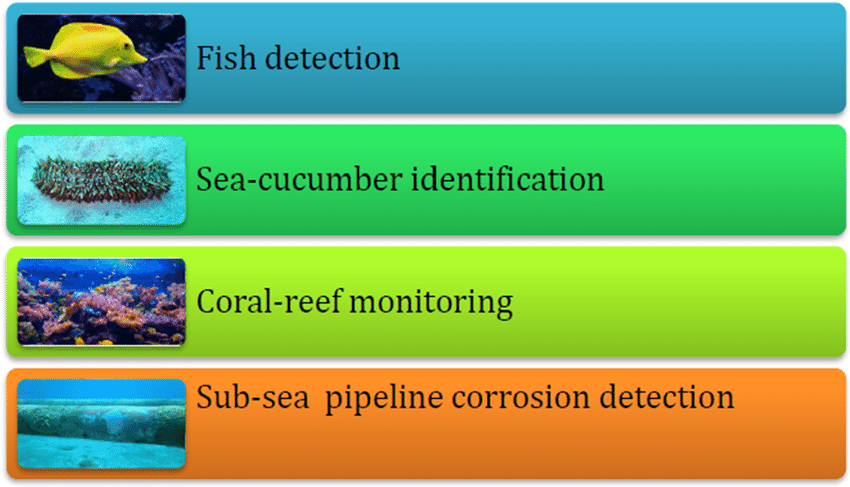}
    \caption{ Application areas of underwater image processing, highlighting its critical roles in marine life identification, oceanography, underwater archaeology, security and surveillance, robotics, photography and videography, mapping and navigation, and virtual reality tourism \cite{a3}.}
    \label{fig:application}
\end{figure}

\subsection{Challenges}
Light attenuation refers to the reduction in light intensity as it travels through a medium, resulting from absorption, scattering, and reflection by particles and molecules within that medium. The degree of light attenuation is influenced by the medium's properties, such as its composition, density, and scattering characteristics.

\begin{figure}[!h]
    \centering
    \includegraphics[width=10 cm,height=7cm,keepaspectratio]{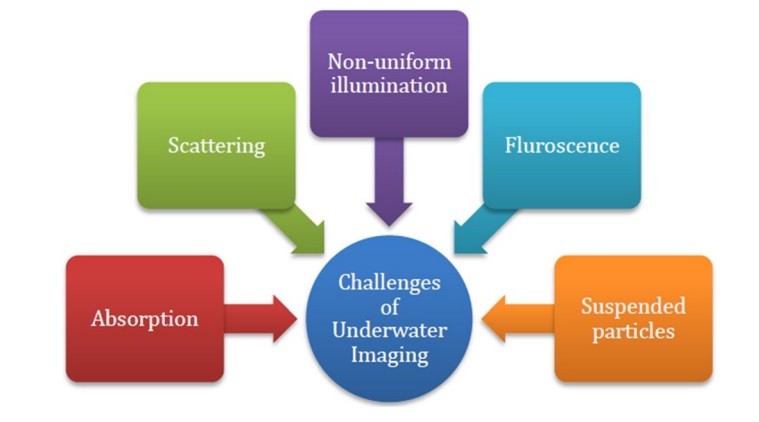}
    \caption{Challenges in underwater imaging include significant light attenuation due to absorption, scattering, and reflection by water molecules, suspended particles, and dissolved substances. The attenuation varies with wavelength, causing shorter wavelengths like blue and green to be absorbed and scattered more than longer wavelengths like red. Additionally, underwater images are affected by fluorescence, non-uniform illumination, and reduced visibility, making it essential to enhance image quality for better exploration and study of underwater environments \cite{a3}.}
    \label{fig:challenge}
\end{figure}

In water, light attenuation is significantly greater than in air due to the higher density and greater concentration of particles and molecules. Water molecules, suspended particles, and dissolved substances like salts and organic matter all contribute to the attenuation of light, as illustrated in Fig \ref{fig:challenge}.

The extent of light attenuation in water varies with the wavelength of the light. Shorter wavelengths, such as blue and green light, are attenuated more strongly than longer wavelengths, such as red and infrared light. This phenomenon explains why objects underwater appear bluer and darker compared to their appearance in air; blue light, having a shorter wavelength, is absorbed and scattered more than red light, which has a longer wavelength.

In contrast, light attenuation in air is much lower due to the lower density and smaller concentration of particles and molecules in the atmosphere. However, atmospheric conditions like fog, haze, and pollution can also contribute to light attenuation, especially for longer wavelengths of light, such as red and infrared.

The underwater environment encompasses areas submerged in water, whether in natural or artificial bodies such as oceans, seas, reservoirs, rivers, or aquifers. It is the cradle of life on Earth and is vital for sustaining diverse life forms, serving as a natural habitat for numerous organisms. Many human activities occur within accessible regions of the underwater environment. Consequently, understanding the characteristics of the underwater imaging model is essential for conducting research across various fields \cite{r5}.

\subsubsection{Absorption and Scattering}
\begin{figure}[!h]
    \centering
    \includegraphics[width=0.5 \textwidth]{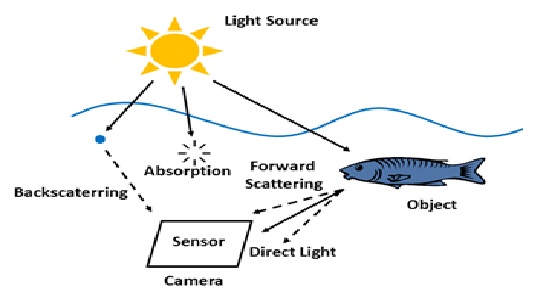}\includegraphics[width=0.5 \textwidth]{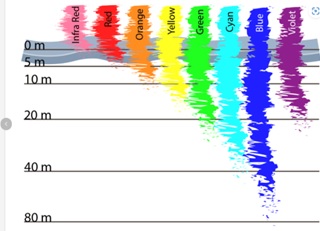}
    \caption{Light attenuation in underwater environments, illustrating the exponential decay of light intensity due to absorption and scattering. The diagram shows how red light, with the longest wavelength, is absorbed first, while blue light, with the shortest wavelength, penetrates the farthest, resulting in a bluish tint in underwater images \cite{raveendran2021underwater}}
    \label{fig:attenuation}
\end{figure}

The Lambert-Beer empirical law states that the decay in light intensity depends on the properties of the medium through which it travels. In water, light intensity decays exponentially through a process known as attenuation. Attenuation results from the combined effects of absorption and scattering, leading to a loss of light energy and a change in the direction of electromagnetic energy. This attenuation poses a significant challenge for underwater imaging by creating a hazy effect that complicates image processing applications in marine environments. In clear water, attenuation limits visibility to approximately 20 meters, whereas in turbid water, visibility is reduced to only 5 meters. Additionally, light absorption in water varies with wavelength; as depth increases, different colors of light are absorbed at different rates. Red, with the longest wavelength, is absorbed first, while blue, with the shortest wavelength, penetrates the farthest, resulting in a bluish tint in underwater images as shown in Fig \ref{fig:attenuation}.

\begin{figure}[!h]
    \centering
    \includegraphics[width=10 cm,height=7cm,keepaspectratio]{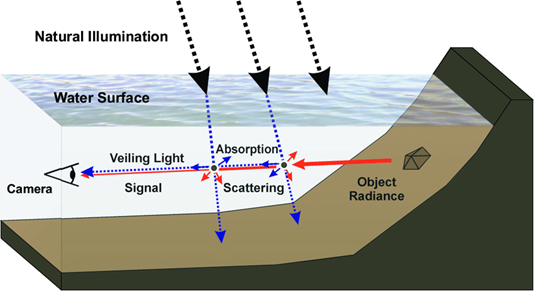}
    \caption{ Absorption and scattering in underwater environments, showing how light interacts with floating particles. The diagram illustrates the effects of forward scattering and backward scattering on the visibility and clarity of underwater images \cite{raveendran2021underwater}.}
    \label{absorption}
\end{figure}

In an underwater medium, the presence of dust particles leads to scattering phenomena. When light reflects off an object's external surface and reaches the camera, it interacts with the floating particles in the medium, causing a scattering effect. There are two types of scattering that affect underwater images: forward scattering and backward scattering as shown in Fig \ref{absorption}.

The model is based on the principles of linear superposition and the water medium modeling defined in the Jaffe–McGlamery model \cite{raveendran2021underwater}. The irradiance entering the camera is a linear combination of three distinct components: the direct component ($E_d$), the forward-scattered component ($E_f$), and the backscatter component ($E_b$). The total irradiance ($E_T$) can be expressed as follows:
\begin{equation}
    E_{T} = E_{d} + E_{f} + E_{b}
\end{equation}
The direct component, denoted as $E_d$, refers to the light that is reflected by an object and reaches the camera without undergoing any scattering. Forward scatter, represented by $E_f$, occurs when the light reflected from an object scatters in its direction before reaching the camera. In contrast, backscatter happens when the light scatters directly towards the camera after reflecting off particles in the water. These models are often used for image restoration but require high-speed computations and longer execution times.

$E_d$ signifies the light that is directly reflected by the object without any scattering in the water. This component is particularly beneficial for underwater imaging and can be expressed as:
\begin{equation}
    E_d(x,y)=E(x,y) e^{-cd(x,y)}
\end{equation}

The expression $E(x,y)$ represents the irradiance at position $(x,y)$. The total attenuation coefficient (c) of the medium quantifies the combined effects of scattering and absorption on light loss within the medium. The variable $d(x,y)$ denotes the distance between the object and the camera. Furthermore, $E_f$ refers to the forward scatter component, which is light reflected by an object and scattered at a small angle before reaching the camera:
\begin{equation}
    E_f(x,y)=E_d(x,y) * g(x,y)
\end{equation}
To denote the convolution operator, the symbol $\ast$ is used, and $g(x,y)$ represents the point spread function (PSF). To avoid the mathematically complex issue of solving the deconvolution through PSF estimation, researchers typically assume that the underwater scene is close to the camera, thereby neglecting the impact of forward scattering.

$E_b$ represents the backscattered light reflected by particles in the water. This component does not include light from the object itself, as it is primarily caused by the scattering of floating particles. $B_\infty$ denotes the underwater background light.
\begin{equation}
    E_b(x,y)=B_\infty(\lambda)(1-e^{-cd(x,y)})
\end{equation}

\subsubsection{Suspended Particles}
The presence of suspended particles in water can be mathematically modeled using the radiative transfer equation, which describes the interaction of light with matter. For underwater images, this equation can model the propagation of light through the water column, including the effects of scattering and absorption by suspended particles \cite{liu2014no}.

A common approach to enhancing underwater images involves using a dehazing algorithm that estimates the transmission map of the image. This map represents the fraction of light that has successfully transmitted through the water column. The transmission map can be estimated using the following equation:
\begin{equation}
    t(x) = e^{(-\beta d(x))}
\end{equation}

where \( t(x) \) is the transmission at pixel \( x \), \( d(x) \) is the distance between the camera and pixel \( x \), and \( \beta \) is the scattering coefficient of the water. The scattering coefficient depends on the concentration and size distribution of suspended particles in the water and can be estimated using empirical or theoretical models.

Once the transmission map is estimated, it can be used to remove the effects of haze and recover the original colors and contrast of the image using the following equation:
\begin{equation}
    I(x) = \frac{(I(x) - A)}{t(x)}  + A
\end{equation}
where \( I(x) \) is the intensity of the image at pixel \( x \), \( A \) represents the atmospheric light (the color of the light in the absence of scattering), and \( t(x) \) is the estimated transmission at pixel \( x \).

Color correction algorithms can also be employed to compensate for the color distortion caused by suspended particles. A common approach is to estimate the color of the ambient light in the underwater environment using a white-balancing algorithm and then adjust the color balance of the image accordingly.

In summary, the key to mathematically enhancing underwater images lies in modeling the effects of suspended particles on light transmission using the radiative transfer equation and applying appropriate image enhancement techniques to mitigate these effects.

\subsubsection{Non-Uniform Illumination}

\begin{figure}[!h]
    \centering
   \includegraphics[width=10 cm,height=5cm,keepaspectratio]{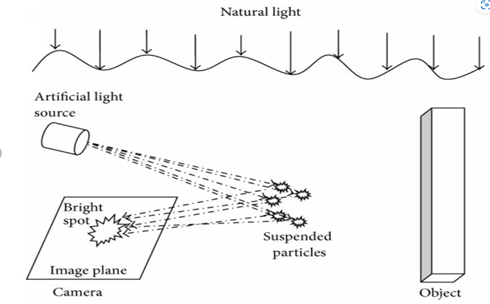}
    \caption{Non-uniform illumination and the presence of suspended particles in water, demonstrating how absorption and scattering lead to blurriness, reduced contrast, and loss of image quality. High turbidity and powerful artificial light sources exacerbate these effects, causing reflections and bright spots that obscure image details \cite{a5}}
    \label{non-ill}
\end{figure}

Absorption and scattering of light in water can lead to blurriness, reduced contrast, and an overall decline in image quality. These effects are further exacerbated in high-turbidity underwater conditions or when powerful artificial light sources are used \cite{h6}. Such light sources can cause non-uniform lighting in fluorescence, resulting in reflections that obscure image details and create bright spots as shown in Fig \ref{non-ill}.

A common method to model non-uniform illumination in underwater environments is by using the Beer-Lambert law. This law describes how light intensity attenuates as it travels through a medium, stating that the intensity of light decreases exponentially with distance:
\begin{equation}
    I = I_0 * e^{(-k*d)}
\end{equation}
where \( I \) is the intensity of the light after passing through the medium, \( I_0 \) is the initial intensity of the light, \( k \) is the extinction coefficient of the medium (a measure of how much the medium absorbs or scatters light), and \( d \) is the distance the light has traveled through the medium.

In underwater environments, the extinction coefficient can vary depending on factors such as water depth, water clarity, and the presence of suspended particles or plankton. Thus, the Beer-Lambert law can be used to model the non-uniformity of underwater illumination \cite{d2}.

Other factors contributing to non-uniform illumination in underwater environments include the angle of incidence of the light, the direction and intensity of light fluorescence, and the presence of shadows and reflections. Modeling these factors may require more complex mathematical formulas, such as ray tracing or radiative transfer models.

\subsubsection{Fluorescence}
\begin{figure}[!h]
    \centering
    \includegraphics[width=10 cm,height=7cm,keepaspectratio]{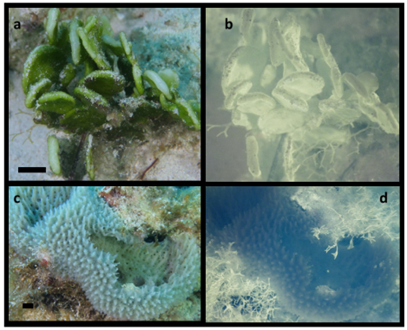}
    \caption{ Illustration of fluorescence, a phenomenon where certain materials absorb light at one wavelength and emit it at a longer wavelength, commonly affecting underwater images by causing color distortions. \cite{a1}}
    \label{fluor}
\end{figure}

Fluorescence is a phenomenon where certain materials absorb light at one wavelength and emit it at a longer wavelength. However, underwater image processing provides methods to overcome these challenges. As shown in Fig \ref{fluor}, visual information can be combined with quantitative assessment to effectively address these issues.

To address these challenges, various techniques and algorithms have been developed for underwater image enhancement. These include classical methods such as histogram equalization and Retinex, as well as deep learning-based approaches utilizing CNNs, GANs, and U-Net \cite{g3}. These techniques aim to improve contrast, sharpness, and color balance in images while minimizing the effects of scattering, absorption, and other factors. However, there is still significant work required to further enhance the quality and clarity of underwater images, particularly under challenging conditions \cite{yussof2013performing}.

\section{Major Contribution}
Most existing works have designed a single deep network for image quality improvement. In contrast, this work proposes an Iterative Framework for Degradation Aware Underwater Image Enhancement (IDA-UIE).

IDA-UIE identifies a dominant degradation condition and appropriately enhances it. Correction of one degradation may reveal another degradation condition. Thus, the enhanced image is further subjected to degradation identification and subsequent enhancement. This system attempts to improve the image quality through degradation-aware enhancement iterations.

This section details the significant contributions made in this project, which focus on enhancing underwater images through an innovative framework and specialized deep networks.

\begin{enumerate}
    \item \textbf{Iterative Framework for Degradation Aware Underwater Image Enhancement :}
One of the primary contributions is the proposal of an iterative framework specifically designed for degradation-aware underwater image enhancement. Traditional methods often employ a single deep network to improve image quality. However, these approaches can fall short when dealing with complex and varied degradation types found in underwater images.

Our iterative framework, named Iterative Degradation Aware Underwater Image Enhancement (IDA-UIE), addresses this by identifying the dominant degradation condition in an image and enhancing it accordingly. The process is iterative because enhancing one type of degradation can reveal another underlying issue. Thus, after the initial enhancement, the image is re-evaluated for additional degradations, which are then corrected in subsequent iterations. This iterative approach ensures a comprehensive enhancement process, gradually improving the image quality through multiple refinement steps.
    \item \textbf{Deep Network for Identifying Dominant Degradation :}
To support the iterative framework, we designed a deep network, denoted as $\mathbf{\Phi}_{DC}$, for identifying the dominant degradation in underwater images. This network is critical as it drives the entire enhancement process by accurately pinpointing the most significant degradation affecting the image.

The $\mathbf{\Phi}_{DC}$ network is trained to recognize eight specific types of degradation: low illumination, low contrast, haziness, blur, noise, and color imbalances in the red, green, and blue channels. Additionally, it can identify if an image is not degraded. This identification step is crucial for ensuring that each image receives the appropriate type of enhancement.

    \item \textbf{Eight Deep Networks for Condition-Specific Underwater Image Enhancement :}
Following the identification of the dominant degradation, the framework employs one of eight specialized deep networks to enhance the image. Each of these networks is tailored to address a specific type of degradation:

\textbf{$\mathbf{\Phi}_{IC}$:} Illumination Correction - Enhances images with low illumination, improving visibility and detail.

\textbf{$\mathbf{\Phi}_{CE}$ :} Contrast Enhancement - Increases the contrast in images, making features more distinguishable.

\textbf{$\mathbf{\Phi}_{DH}$:} Removes haziness to clarify images.

\textbf{$\mathbf{\Phi}_{DB}$: } Sharpens images to correct blur.

\textbf{$\mathbf{\Phi}_{DN}$:} Reduces noise to produce cleaner images.

\textbf{$\mathbf{\Phi}_{CBR}$ :} Color Balance for Red Channel - Corrects color imbalances in the red channel.

\textbf{$\mathbf{\Phi}_{CBB}$:} Color Balance for Blue Channel - Corrects color imbalances in the blue channel.

\textbf{$\mathbf{\Phi}_{CBG}$: }Color Balance for Green Channel - Corrects color imbalances in the green channel.

Each network has been meticulously designed and trained to excel at its specific enhancement task, ensuring that the iterative framework can effectively improve various aspects of underwater images.

    \item \textbf{Construction of Two Datasets with Condition-Specific Degradations :}
To train the nine deep networks (one for degradation identification and eight for specific enhancements), we constructed two extensive datasets: UIEB-D8 and EUVP-X-D8. These datasets are based on standard underwater image datasets but have been augmented with condition-specific degradations to simulate real-world underwater conditions more accurately.

Each image in these datasets has been systematically degraded to reflect one of the eight targeted conditions. This detailed and condition-specific dataset construction ensures that the networks are well-trained to recognize and correct each type of degradation effectively.

\textbf{UIEB-D8 Dataset}
The UIEB-D8 dataset is derived from the UIEB dataset ~\cite{li2019underwater}, with images subjected to controlled degradations to create training examples for each of the eight conditions. This dataset provides a robust foundation for training the enhancement networks.

\textbf{EUVP-X-D8 Dataset}
Similarly, the EUVP-X-D8 dataset is based on the EUVP dataset ~\cite{9001231} and includes images with various degradations. By using these two diverse datasets, the networks are trained to handle a wide range of underwater image conditions, enhancing their generalizability and effectiveness.

\end{enumerate}
 
\begin{figure}[ht!]
\centering
\includegraphics[width=1.0\columnwidth,height=0.4\columnwidth]{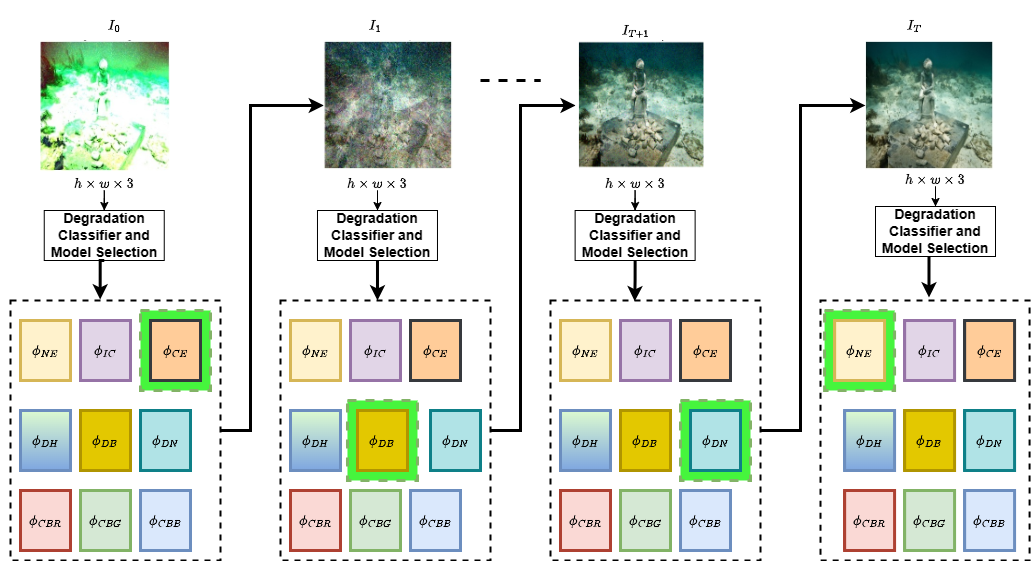}
\caption{The functional block diagram of the Iterative Framework for Degradation Aware Underwater Image Enhancement (IDA-UIE), illustrating the process of identifying dominant degradations and applying condition-specific enhancements iteratively to improve overall image quality.}
\label{fig:IllustrationModel}
\end{figure}

\section{Related Work}
\begin{figure}[!h]
    \centering
    \includegraphics[width=\textwidth,height=10cm,keepaspectratio]{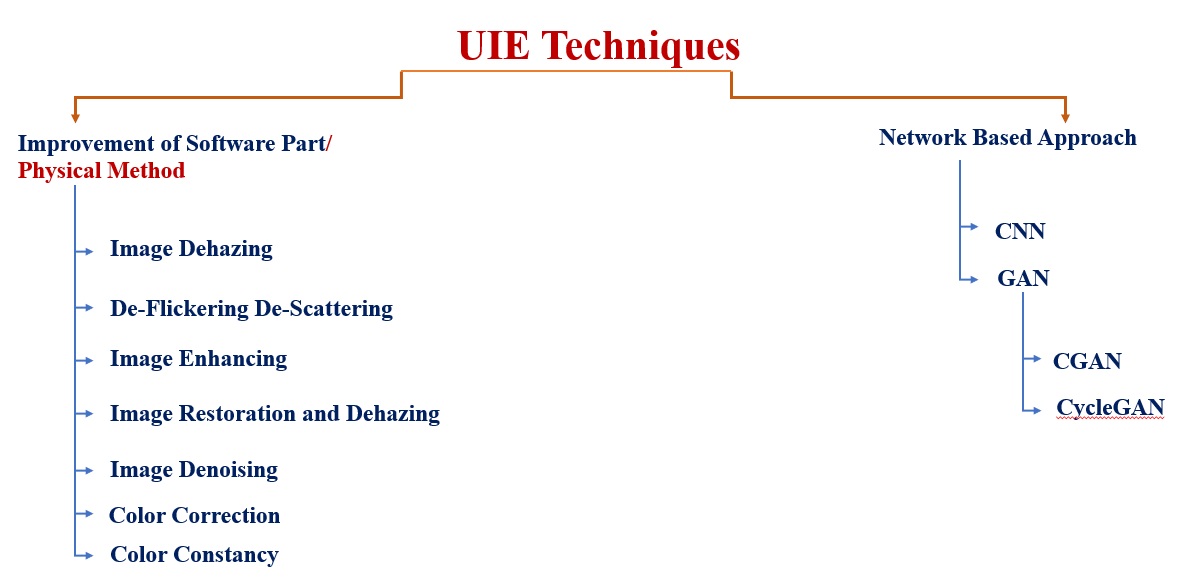}
    \caption{Techniques of Underwater Image Enhancement (UIE), categorized into traditional methods and deep learning-based methods, illustrating the various approaches used to improve underwater image quality. \cite{raveendran2021underwater}}
    \label{fig:technique}
\end{figure}

Here, presents a classification and summary of existing techniques for enhancing underwater images, mainly categorized into traditional and deep learning-based methods. The underwater image enhancement (UIE) techniques are broadly categorized in Figure \ref{fig:technique}.
\subsection{Traditional Methods}

Traditional methods include both model-based and non-model methods. Non-model methods, such as the histogram algorithm, enhance visual effects through pixel adjustments without considering imaging principles. Model-based methods, also known as image restoration techniques, estimate the relationship between clear, blurry, and transmission images based on an imaging model to produce clear images. An example of a model-based method is the dark channel prior (DCP) algorithm, as shown in Figure {fig:technique} \cite{raveendran2021underwater}.

\subsubsection{Image Denoising}
Image denoising is a technique used to reduce or remove noise from digital images. It aims to improve the visual quality of an image by suppressing unwanted noise while preserving important details and structures \cite{hu2022overview}.

\subsubsection{Contrast Enhancement Techniques}
Image quality is often evaluated based on the level of contrast in the image. Contrast refers to the difference in luminance reflected from two adjacent planes and is a key factor in making objects distinguishable from the background. Vision is more sensitive to contrast than absolute luminance, which allows us to perceive the world despite variations in illumination conditions. If an image has highly concentrated contrast in a particular range, such as being very dark, critical information may be lost in those areas. Therefore, optimizing the contrast is necessary to represent all the details in the input image. To address issues related to contrast in underwater image processing, numerous algorithms for achieving contrast enhancement have been developed \cite{raveendran2021underwater}.

\subsubsection{Color Correction Techniques}
The colors present in underwater images are mainly blue and green due to their shorter wavelengths. The histogram distribution of these images indicates that the green channel's mean is more significant than that of the red channel, and the RGB channels' distribution range does not cover the full range of [0, 255]. To correct the issue of color cast, color correction techniques are used to improve the visual information content of underwater images. A manual correction approach is found to be better than automatic enhancement techniques in terms of the significance level. An enhancement method that uses fuzzy logic and bacterial foraging optimization is proposed to remove the color cast, which gives better results than existing algorithms. Additionally, a method for non-uniform illumination correction is proposed, which uses maximum-likelihood estimation to map the image to Rayleigh distribution. An adaptive linear stretch method that adjusts regions with low light distributions with a threshold depending on the histogram is also proposed. 

\subsubsection{Histogram Equalization Method}
Underwater images often require image enhancement for improved quality. As such, there are several methods available in the literature to address this issue. In this study, a new underwater image enhancement method is proposed. This method employs the HSV, V transform algorithm, and histogram equalization techniques. Initially, the RGB image is separated into its R, G, and B components, and then converted into the HSV color space. The V element is then extended within a specified interval before converting the image back to the RGB color space. Histogram equalization is then applied to each of the R, G, and B components, and the components are combined to form a color image. Finally, a Gaussian low-pass filter is applied to the image. The performance of the proposed method is compared to that of other studies using mean value and entropy metric, which demonstrate that the proposed method significantly improves underwater image quality \cite{deperlioglu2018underwater}.

\subsubsection{CLAHE}
Ordinary AHE tends to over-amplify the contrast in near-constant regions of the image. It is originally developed for the enhancement of low-contrast images \cite{a1}. CLAHE is a variant of adaptive histogram equalization in which contrast amplification is limited to reduce this problem of noise amplification. In order to limit noise amplification, CLAHE is used \cite{f8}. In CLAHE, the contrast-limited procedure is applied to each neighborhood from which the transformation function is derived. Rather than taking the whole image, CLAHE prevents over-amplification by dividing the image into small data regions called tiles and performing contrast enhancement \cite{hasibuan2021contrast}. These tiles are then rejoined to get an overall enhanced image. It is applied to both grayscale and colored images \cite{hassan2021retinex} \cite{hasibuan2021contrast}.

\subsubsection{Retinex Based Method}
Underwater images often suffer from low contrast and color distortion due to the variable attenuation of light and non-uniform absorption of red, green, and blue components. To address these issues, a Retinex-based approach for underwater image enhancement has been proposed. The approach involves using contrast-limited adaptive histogram equalization (CLAHE) to enhance the contrast of the darker components of the underwater image while limiting noise, which may blur visual information. Next, a Retinex-based enhancement is performed on the CLAHE-processed image to restore distorted colors \cite{hassan2021retinex} \cite{hasibuan2021contrast}. To restore distorted edges and achieve smoothing of the blurred parts of the image, bilateral filtering is performed on the Retinex-processed image. To optimize the individual strengths of CLAHE, Retinex, and bilateral filtering algorithms within a single framework, suitable parameter values are determined. Comparing the performance of the proposed approach with existing methods, both qualitatively and quantitatively, indicates that it results in better enhancement of underwater images \cite{yussof2013performing}.

\subsubsection{Dark Channel Prior}
Haze arises from particles suspended in bodies of water such as sand, minerals, and plankton. This phenomenon disrupts the clarity of underwater images by reducing contrast, causing poor visibility, absorbing natural light, and limiting color variation. Enhancing the quality and visibility of underwater images requires the dehazing process \cite{kumarunderwater}. This research introduces the Dark Channel Prior (DCP) algorithm, which capitalizes on the observation that most local patches in haze-free outdoor images contain pixels with very low intensity in at least one color channel. By utilizing DCP, underwater images exhibit significantly improved visibility and superior color accuracy. Moreover, this approach reduces computational complexity and enhances dehazing efficiency. Underwater images experience distortions primarily due to light dispersion and color effects. The dispersion of light and its scattering in water reduces the visibility and contrast of captured images. Additionally, color changes caused by the presence of particles such as sand, minerals, and plankton in the water, along with the absorption and scattering of natural light, further impact underwater images. When light reflects from objects in the water, it encounters suspended particles, leading to light absorption and scattering \cite{kumarunderwater}. To address these issues, the Dark Channel Prior (DCP) method is applied. This method estimates the atmospheric light and utilizes a mathematical function to handle both sky and non-sky regions. It identifies affected patches in the images, estimates the scene depth, and removes the haze to enhance the clarity of the image. To improve the accuracy of the depth map generated by the block-based dark channel prior, image matting is employed. This combination of techniques enhances accuracy and enables more precise identification of object contours \cite{kumarunderwater}. The application of image matting to the underwater depth map, derived through the general dark-channel methodology, represents a novel approach. Subsequently, the following section presents a list of existing works in this field.

\subsubsection{Other Methods}
Underwater images often suffer from low contrast and poor visibility, making it crucial to enhance them before further processing. Image enhancement techniques aim to improve the quality and contrast of degraded underwater photos and videos. Standard cameras used for capturing underwater scenes face challenges such as limited available light, low resolution, and blurriness, necessitating the improvement of the initial images or videos obtained from image processing equipment. Researchers have proposed various solutions to address these challenges.

One commonly used approach for enhancing underwater images is the dark channel prior (DCP), which aims to improve the Peak Signal to Noise Ratio (PSNR). However, DCP has significant drawbacks, including the tendency to darken images, reduce contrast, and introduce halo effects. To overcome these limitations, the suggested technique incorporates contrast-limited adaptive histogram equalization (CLAHE) and the Adaptive Color Correction technique.

To evaluate the proposed approach, experiments were conducted using photographs obtained from the Japan Agency for Marine-Earth Science and Technology (JAMSTEC) as well as from the internet. Performance measures such as entropy (MOE), enhancement (EME), mean square error (MSE), and PSNR were used during the evaluation. The results demonstrate that the proposed framework outperforms other methods in terms of MSE and PSNR, achieving values of 0.26 and 32, respectively. 

\textbf{Mean Filter}
The mean filter is a method employed to decrease image noise. It involves performing a local averaging operation, making it one of the most basic linear filters. In this technique, each pixel's value is substituted with the average value of all the pixels in its surrounding neighborhood. If we denote a noisy image as \( f(i,j) \), the resulting smoothed image can be obtained as \( g(x,y) \) by following this process.

\begin{equation}
    g(x,y)=\frac{1}{n}_{i,j \in S} \sum f(i,j)
\end{equation}

\textbf{Bilateral Filter}
A bilateral filter is a non-linear image-smoothing filter that preserves edges while reducing noise. It operates by replacing the intensity of each pixel with a weighted average of the intensities of nearby pixels. The weights are determined using a Gaussian distribution.
\begin{equation}
    BF[I]_p = \sum_{q \in S} G_{\sigma_s}(||p-q||) G_{\sigma_r}(|I_p-I_q|)I_q
\end{equation}
\textbf{Gaussian Filter}
A Gaussian Filter serves as a low-pass filter employed to diminish noise (high-frequency components) and blur specific areas within an image. This filter is constructed as an Odd-sized Symmetric Kernel (a Matrix in Digital Image Processing terms), which is applied to each pixel in the Region of Interest to achieve the intended outcome. The kernel is designed to be gentle regarding significant colour changes (edges), as the pixels near the centre of the kernel hold more significance in determining the final value compared to those at the edges.
\begin{equation}
    G(x,y)=\frac{1}{2 \pi \sigma^2} e ^{-\frac{x^2+y^2}{2\sigma^2}}
\end{equation}

\textbf{Median Filter}
The median filter, frequently employed in digital filtering, is a non-linear technique aimed at eliminating noise from images or signals. It serves as a common pre-processing step to enhance subsequent processing outcomes, such as image edge detection.

\subsection{Deep Learning}
Here, proposes a CNN-based network for enhancing underwater images, which can learn a mapping to estimate the color-corrected image and transmission map without requiring extra labels on the target source. The report employs a pixels-disrupting strategy to suppress the interference of tiny textures in local patches, resulting in improved convergent speed and accuracy during the learning process. The proposed framework is trained on a synthesis dataset of 200,000 underwater images using the underwater imaging model presented in this report and demonstrates superior generalization ability on real-source underwater images.

Deep underwater image enhancement algorithms can be categorized into two primary types: CNN-based and GAN-based algorithms. The CNN algorithms focus on preserving the authenticity of the original underwater image, while the GAN-based algorithms strive to enhance the visual quality of the images. However, this classification is simplistic, so classify the networks based on their architectural distinctions.

\subsubsection{Encoder-Decoder Models}
\begin{figure}
    \centering
    \includegraphics[width=\textwidth,height=9cm,keepaspectratio]{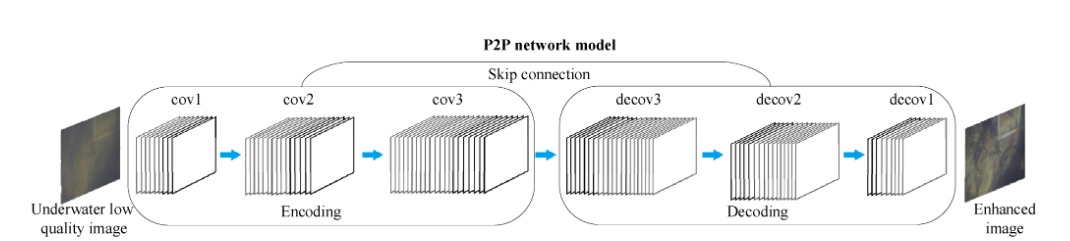}
    \caption{Overview of CNN-based and GAN-based algorithms for underwater image enhancement. The CNN-based network learns to estimate color-corrected images and transmission maps without extra labels, employing a pixels-disrupting strategy for improved convergence and accuracy. Encoder-Decoder models, such as the P2P Network, utilize a symmetric architecture with convolutional and deconvolutional layers to enhance image quality. \cite{sun2019deep}}
    \label{fig:p2p}
\end{figure}

The following models benefit from the well-known encoder–decoder
architecture to advance underwater image enhancement research.
\textbf{P2P Network}
Recently, \cite{sun2019deep} proposed an approach to improve the quality of underwater images using pixel-to-pixel (P2P) networks. Their model, resembling REDNet \cite{mao2016image}, adopts a symmetric architecture consisting of an encoder and a decoder. The encoder is constructed with three convolutional layers, while the decoder is formed by three deconvolutional layers. ReLU activation is applied to each network element except for the last one as shown in Fig \ref{fig:p2p}.

To train the model, the authors utilized a dataset of 3359 real-world underwater images. They introduced degradation levels by adding 30, 50, and 70 ml of milk to 1 $m^3$ of water, representing low, medium, and high degradation, respectively. Among the dataset, 10,000 images were chosen for training purposes, and an additional 2000 images were reserved for testing.

 To achieve a data-driven image enhancement model, the super-parameters of our network play a crucial role. The convolutional part retains the first three layers while discarding the fully connected layers. The reason behind this decision is that the full connection layers are designed for feature mapping from two dimensions to one, primarily for input to classifiers. However, the objective is to create a pixel-to-pixel network for image enhancement, which differs from classification tasks. Utilizing full connection layers would result in the loss of important two-dimensional information, making it unsuitable for underwater image enhancement\cite{sun2019deep}. Additionally, chose to abandon the pooling layers. Although pooling and unpooling layers can enhance object recognition and semantic segmentation by sharpening object edges, they are unnecessary and detrimental to image enhancement and denoising tasks. This is primarily because pooling layers lead to denser feature graphs during the multi-to-one mapping operation, causing the loss of spatial information within a receptive field. Furthermore, the corresponding unpooling layers introduce considerable noise information. During the unpooling mapping, only one value originates from the original feature map, while the remaining values are artificially generated (typically filled with zeros) \cite{sun2019deep}.

\subsubsection{U-Net}
\begin{figure}
    \centering
    \includegraphics[width=\textwidth,height=9cm,keepaspectratio]{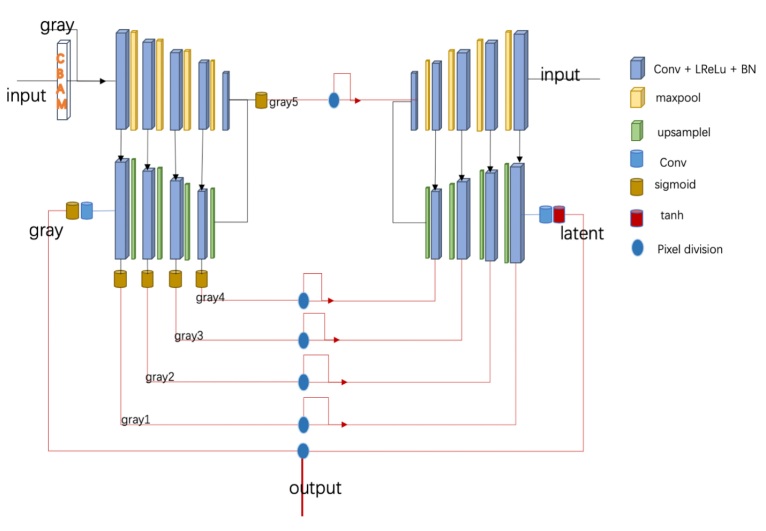}
    \caption{ Diagram of the improved U-Net structure for underwater image enhancement. The convolutional block attention module (CBAM) is integrated as an attention mechanism to enhance feature extraction by applying attention to both channel and spatial dimensions. This structure includes an encoder stage with five layers, each containing two convolutional layers with a kernel size of 3, followed by LeakyReLU activation and BatchNorm2d functions. The first U-Net estimates the latent image after compensating for red light, while another U-Net estimates the transmission image from the input grayscale image \cite{wang2020underwater} \cite{wang2020underwater}}
    \label{fig:unet}
\end{figure}

The improvement of U-Net is based on  network structure. The specific structure diagram is shown in Fig. \ref{fig:unet}. The convolutional block attention module (CBAM) was added to the first U-Net as an attention mechanism module that combines spatial and channel. By applying attention to both the channel and spatial dimensions, it can be embedded into most current mainstream networks, and the feature extraction ability of the network model can be improved without significantly increasing the amount of computation and the number of parameters.
A latent image representing the underwater image after compensating for the red light was estimated by using a U-Net, and another U-Net was used to estimate the transmission image from the input grey-scale image. To avoid losing details during network mapping, the CBAM was added \cite{wang2020underwater}.
The first U-Net consists of an encoder stage and a decoder stage. The encoder stage consists of five network layers, with each layer containing two convolution layers. A kernel size of 3 is used in each convolutional layer, and each convolutional layer is followed by a LeakyReLU activation function and a BatchNorm2d function.
A combination of multi-scale structure similarity and L1 is used for the loss function. To calculate SSIM, the appropriate selection of the size of the Gaussian kernel to compute the image mean value and variance is particularly crucial. If the selection is small, the local structure of the image cannot be ll-maintained by the calculated SSIM loss, and artifacts will appear. If the selection is large, the noise will be generated by the network at the edge of the image.

\subsubsection{Conditional Generative Adversarial Network}
Underwater images are crucial for obtaining and interpreting information about the underwater environment. The reliability of underwater intelligent systems depends on high-quality underwater images. Unfortunately, these images often suffer from low contrast, color casts, blurring, low light, and uneven illumination, which severely limit their usefulness. To address this issue, numerous methods have been proposed, including those that utilize deep learning technologies. Hover, the performance of these methods is often unsatisfactory due to a lack of sufficient training data and effective network structures \cite{yang2020underwater}.

To tackle these challenges, this report proposes a conditional generative adversarial network (cGAN) for enhancing underwater images. The proposed approach uses a multi-scale generator to produce clear underwater images and a dual discriminator to capture local and global semantic information, ensuring that the generated results are both realistic and natural. Experimental results, obtained from both real-world and synthetic underwater images, show that the proposed method outperforms existing state-of-the-art underwater image enhancement methods \cite{yang2020underwater}.

\begin{figure}
    \centering
    \includegraphics[width=\textwidth,height=14cm,keepaspectratio]{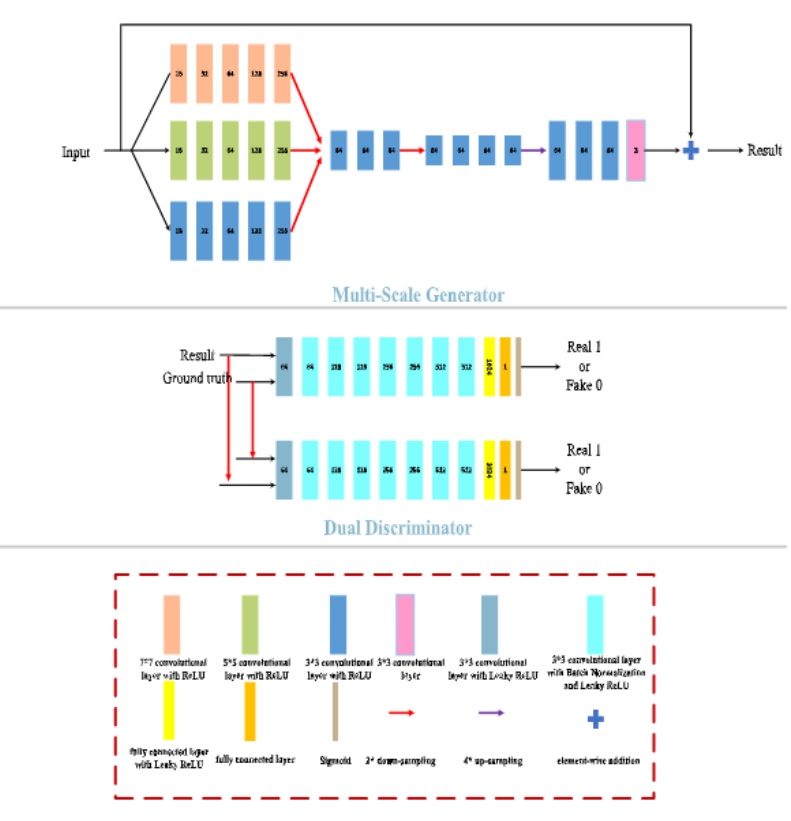}
    \caption{ Illustration of the proposed conditional generative adversarial network (cGAN) for enhancing underwater images. This approach employs a multi-scale generator to produce clear underwater images and a dual discriminator to capture both local and global semantic information, ensuring realistic and natural results. Experimental results demonstrate that this method outperforms existing state-of-the-art underwater image enhancement techniques \cite{yang2020underwater}\cite{yang2020underwater}.}
    \label{fig:my_label}
\end{figure}
\textbf{Multi-Scale Generator}. cGAN's multi-scale generator comprises three main components: a multi-scale feature extraction unit, a feature refinement unit, and a residual map estimation unit. The multi-scale feature extraction unit is constructed using three sets of multi-scale convolutions with different kernel sizes (7x7, 5x5, and 3x3), each set consisting of five convolutional layers with increasing filter numbers ranging from 16 to 256 \cite{yang2020underwater}. A non-linear activation ReLU follows each convolutional layer. The multi-scale feature extractor aims to obtain statistical information from inputs on various scales by acquiring different receptive fields. The multi-scale features are then down-sampled by half of their original size, concatenated, and fed to the feature refinement unit to capture global features and reduce computational costs. The refined features are processed through successive convolutional layers before being down-sampled and fed to three successive convolutional layers, each with 64 filters, and then up-sampled to their original size. Finally, a residual map is estimated by a convolutional layer without non-linear activity, which is used to achieve the final enhanced result via element-wise addition. Zero padding is applied to each convolutional layer to maintain input and output sizes. With the exception of the multi-scale feature extractor's convolutional layers, all convolutional layers have 3x3 kernel sizes. Unlike the common encoder-decoder and cGAN network structures, the generator includes a multi-scale feature extraction unit designed to enhance network capability and adapt to varying underwater sources. Additionally,  the generator has a shallow and lightweight structure and does not use skip connections.

\textbf{Dual Discriminator}. The dual discriminator comprises two sub-discriminators with identical network structures but different weights. Additionally, the inputs to these sub-discriminators have different sizes - one is the original size, while the other is half the original size. The dual discriminator aims to guide the generator in producing realistic images at both the global semantic and local detail levels. This design is necessary because the existing discriminator cannot effectively guide the generator to create realistic details. By providing multi-resolution inputs to different discriminators, the visual quality of the results can be improved. Specifically, the sub-discriminator contains eight convolutional layers with an increasing number of 3x3 filters, increasing from 64 to 512 by a factor of 2. Stridden convolutions are used to reduce the image resolutions, and the 512 feature maps are fed to two fully connected layers to predict the probability of the inputs being real or fake. Unlike the multi-scale generator, the first convolution in the sub-discriminator is followed by Leaky ReLU non-linear activation, while the other convolutions are followed by batch normalization and Leaky ReLU. The last fully connected layer uses the Sigmoid non-linear activation to predict the probability, which is commonly used in image classification tasks. These two sub-discriminators are employed to guide the multiscale generator \cite{yang2020underwater}.

\subsubsection{Cycle GAN}
A variation of the standard GAN network structure is the cycle-consistent adversarial network (CycleGAN), which uses two mirror-symmetric GAN generators and two matching discriminators arranged in a ring network. The CycleGAN framework involves training two GAN networks, denoted as G and F, along with two discriminators, $D_x$ and $D_y$. The generators G and F are utilized to discover the mapping relationships between the X and Y domains and the Y and X domains, respectively. The necessary conditions for the input picture and the produced image to correlate are F(G(x)) $\approx$ x  and G(F(y)) $\approx$ y. 
To achieve cyclic consistency, Cyc1e GAN is suggested as the cyclic consistency loss function. This network structure overcomes the challenge faced by GANs, which require paired data for training, and performs all with underwater photos that do not have paired data \cite{r5}.
The CycleGAN is a GAN designed for unpaired image-to-image translation, where the task is 
to translate images from a Source domain X to a target domain Y. It consists of two GANs, one 
for translating from domain X to Y and one from Y to X \cite{r4}. 
The two discriminators represent the functions:
\begin{equation}
    D_A: X \to R; 
    D_B: Y \to R
\end{equation}
the two generators represent the function:
\begin{equation}
    G_A: X \to Y;
    G_B: Y \to X
\end{equation}

\textbf{Discriminator}
The structure of the discriminators used in cycleGAN is rather conventional: fully 
convolutional neural networks with five-layer blocks, each of which has an instance 
normalization layer, a leaky reLU layer, and a 2D convolution layer with a kernel size of 4x4 
and stride of 2. (except the output block which uses a Sigmoid Layer as activation) \cite{h7}. Each of the 
first, five-layer blocks will reduce the size of the picture by half and increase the number of 
channels every time an image is fed into a discriminator. The input will thus have 512 channels 
and a size of 16x16 after the fifth layer (The model input has a size of 256x256). The output 
layer block will finally combine all 512 channels into a single 16x16 channel.
\begin{figure}[!h]
    \centering
    \includegraphics[width=\textwidth,height=14cm,keepaspectratio]{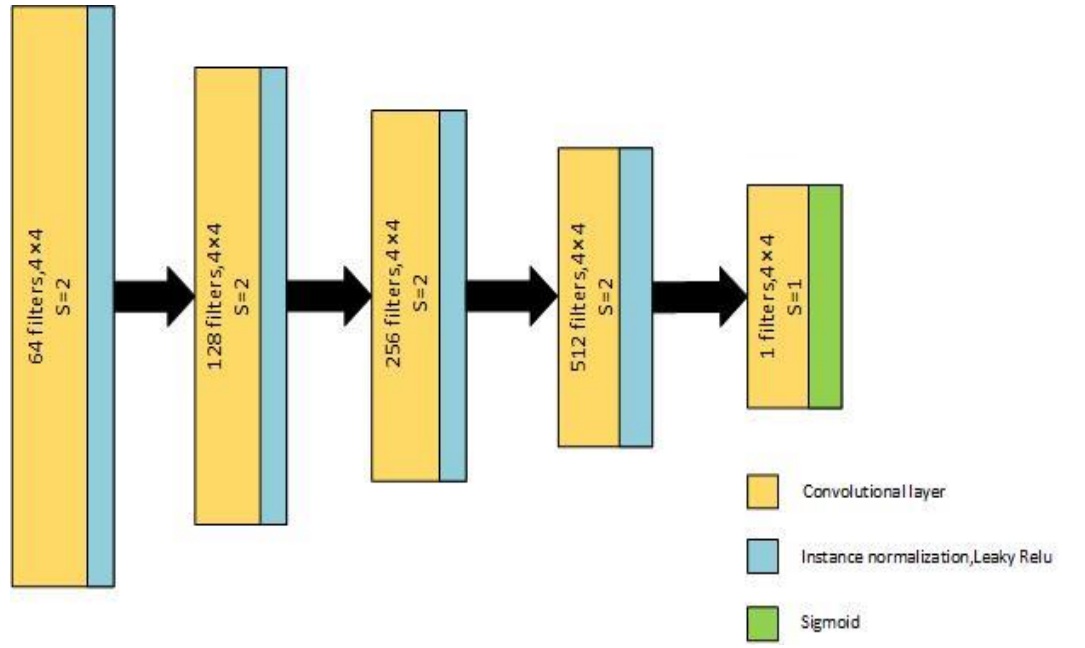}
    \caption{Structure of the PatchGAN discriminator used in CycleGAN for underwater image enhancement. The discriminator comprises five convolutional blocks, each with instance normalization, leaky ReLU, and 2D convolution layers. The final block uses a Sigmoid activation. Input image size is halved and channels increased at each block, resulting in a 16x16 output with 512 channels \cite{wang2020underwater}}.
    \label{fig:patch}
\end{figure}

\textbf{Generator}
A generator's goal is to alter the input picture and produce it as the output. A neural network 
structure is made up of three parts an encoder, a transformer, and a decoder. While increasing the 
number of channels, the encoder reduces the size of the input pictures. It is made up of 3 layers
of blocks, similar to the Discriminator, with a 2D Convolution Layer, an Instance Normalization 
Layer, and a Leaky ReLU Layer in each block. The first layer block just adds 64 input channels; 
it has no effect on the image's size. However, each of the next 2 layers of blocks reduces the input 
size by 50\% while increasing the number of channels. The transformer then receives the altered 
input \cite{g2} \cite{g3}.
The transformer maintains the input's size while adding the needed characteristics \cite{f3}. It has six 
ResNet blocks, also known as residual netblocks. Each ResNet has two layers blocks: the first 
layer block has a Leaky ReLU Layer, an Instance Normalization Layer, and a 2D Convolution 
Layer (with stride=1). A 2D Convolution Layer (with stride=1) and an Instance Normalization 
Layers are both included in the second layer block. The decoder receives the modified input after 
that \cite{f1}.
To create the final output image, the decoder shrinks the input to its original size and collapses 
all channels into RGB. Two Transpose Convolution Layers are stacked together to accomplish 
the enlargement operation. A transpose convolution layer might be thought of as a simple 
combination of a 2D Up-sampling layer and a 2D convolution layer with stride=1. In general, 
it will reduce the number of channels while increasing the size of the input. The output layer 
will eventually receive the 256x256 pixel data with 64 channels generated by the two transpose 
convolution layers, collapse the channels into RGB, and output it as the final output image \cite{e1} \cite{e2}.

\begin{figure}[!h]
    \centering
    \includegraphics[width=\textwidth]{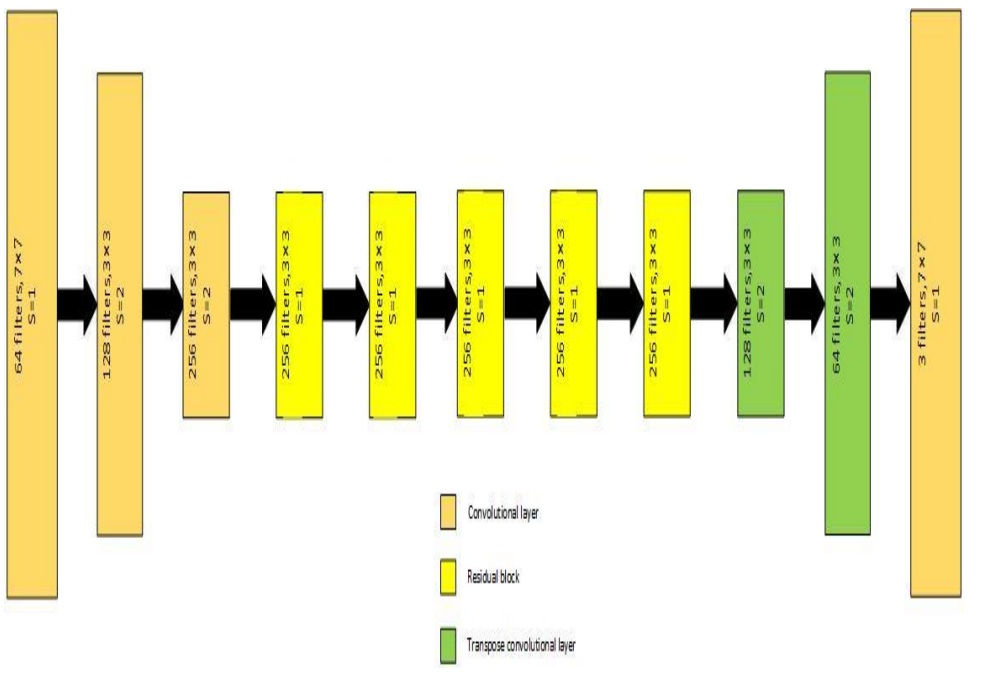}
    \caption{Structure of the generator in CycleGAN for underwater image enhancement. It consists of an encoder, transformer, and decoder. The encoder reduces image size and increases channels, the transformer adds features with six ResNet blocks, and the decoder restores the original size and converts channels to RGB for the final output \cite{wang2020underwater}.}
    \label{fig:generator}
\end{figure}

 Deep learning techniques have shown promise in enhancing underwater images, but there are gaps in the research that need to be addressed. One challenge is the high number of parameters involved, leading to overfitting and reduced generalization ability. There is a need for research to develop efficient deep-learning models with fewer parameters that still achieve good performance. Another gap is the interpretability of deep learning models, which are often considered "black-box" models, making it difficult to understand how they make decisions. There is a need for developing methods to interpret these models to identify strengths and weaknesses and improve performance. Overall, the research gap in deep learning for underwater image enhancement is in developing efficient and interpretable models with good performance.

Several methods have been proposed to tackle challenges in Underwater Image Enhancement (UIE). Challenges like light attenuation and scattering often result in color casts and diminished visibility \cite{yi2024no}. One particularly noteworthy approach introduced a novel quality assessment method centered around colorfulness, contrast, and visibility metrics, providing an effective means to evaluate UIE outcomes \cite{yi2024no}. However, the diverse underwater landscapes pose a challenge to existing color constancy methods. To address this, an adaptive UIE technique leveraging hue channel statistics and deep learning networks trained on authentic datasets with ground truth annotations was developed in~\cite{li2024underwater}. 

Texture and color enhancement are pivotal for effective underwater image enhancement, and the Texture-Aware and Color-Consistent Network (TACC-Net) has emerged as a standout performer in this regard. By decoupling features to enhance texture and ensure color consistency, TACC-Net has significantly improved visual quality \cite{hu2024texture}. Meanwhile, issues such as light absorption and turbulence continue to impair image quality in underwater target imaging, affecting clarity and resolution. To address these challenges, a study has proposed a block mixed filter denoising technique and underscored the importance of objective quality evaluation for image enhancement methods \cite{xiao2024underwater}.


\subsection{Baseline Methods:}

\subsubsection{Fusion Based}
This paper \cite{6247661} introduces a novel strategy to enhance underwater videos and images using fusion principles. The unique aspect of this strategy is that it derives both the inputs and the weight measures solely from the degraded version of the image, without the need for specialized hardware or prior knowledge of underwater conditions or scene structure.

The approach involves the derivation of two inputs from the original underwater image or frame. The first input is a color-corrected version that addresses the color distortion commonly caused by underwater environments. The second input is a contrast-enhanced version, which aims to improve the visibility of details often lost in the hazy underwater images. These inputs help to mitigate the color and contrast issues inherent in underwater imaging \cite{6247661}.

Additionally, four weight maps are defined to increase the visibility of distant objects, which are usually degraded due to medium scattering and absorption in underwater environments. These weight maps help in selectively emphasizing important features in the image, enhancing the overall clarity.

The fusion framework integrates these inputs and weight maps to produce an enhanced image. This approach ensures that the finest details and edges in the image are significantly improved. The enhanced images and videos are characterized by a reduced noise level, as effective edge-preserving noise reduction strategies are applied to minimize noise while retaining important details. Dark areas in the image are better exposed, making hidden details more visible. The overall contrast of the image is enhanced, making it more visually appealing and informative.

For videos, the framework also ensures temporal coherence between adjacent frames. This means that the enhancement process maintains consistency across frames, preventing flickering or abrupt changes that can distract viewers.

The utility of this enhancement technique is demonstrated across several challenging applications, showing its versatility and effectiveness in various underwater imaging scenarios.

\subsubsection{UGan-Based }
Autonomous underwater vehicles (AUVs) rely on a variety of sensors, including acoustic, inertial, and visual sensors, for intelligent decision-making. Among these, vision is particularly attractive due to its non-intrusive, passive nature and high information content, especially at shallower depths. However, several factors adversely affect the quality of visual data obtained underwater. Light refraction and absorption, suspended particles in the water, and color distortion all contribute to producing noisy and distorted images. Consequently, AUVs that depend on visual sensing face significant challenges and often exhibit poor performance on vision-driven tasks \cite{8460552}.

This paper \cite{8460552} proposes a method to enhance the quality of visual underwater scenes using Generative Adversarial Networks (GANs). The goal is to improve the visual input for vision-driven behaviors further down the autonomy pipeline of AUVs. GANs are well-suited for this task because of their ability to generate high-quality images that closely resemble real-world scenes, making them ideal for underwater image restoration \cite{8460552}.

The key challenges in underwater visual data include light refraction and absorption, suspended particles, and color distortion. Underwater environments significantly alter light paths, causing refraction and absorption that lead to reduced clarity and visibility. Particles in the water scatter light, creating a hazy appearance and further degrading image quality. The underwater medium absorbs different wavelengths of light at different rates, causing color distortions that affect the accuracy of visual data \cite{8460552}.

The proposed method leverages the power of GANs to address these challenges. GANs consist of two neural networks: a generator and a discriminator. The generator creates enhanced images from the degraded input, while the discriminator evaluates the authenticity of the generated images, driving the generator to produce increasingly realistic enhancements. This adversarial process results in images that are not only visually appealing but also more useful for subsequent vision-driven tasks.

To train the GANs effectively, a dataset specifically tailored for underwater image restoration is required. Recently proposed methods allow for the generation of such datasets by simulating various underwater conditions and degradations. This synthetic dataset includes images with different types of distortions commonly found in underwater environments, providing a comprehensive training set for the GANs.

For visually-guided underwater robots, improving the quality of visual data can lead to increased safety and reliability. Enhanced visual perception enables AUVs to perform better in tasks such as navigation, object detection, and diver tracking. The proposed GAN-based approach not only generates visually appealing images but also enhances the accuracy of vision-driven algorithms.

The effectiveness of the proposed method is demonstrated through both quantitative and qualitative evaluations. Enhanced images show significant improvements in clarity, color accuracy, and detail preservation compared to the original degraded images. Additionally, these improvements translate to increased accuracy for a diver tracking algorithm, showcasing the practical benefits of the enhanced visual data.

\subsubsection{ FUnIE-GAN }
In this work, \cite{9001231} a conditional generative adversarial network-based model is presented for real-time underwater image enhancement. The model's adversarial training is supervised by an objective function that evaluates perceptual image quality based on global content, color, local texture, and style information. A large-scale dataset, EUVP, is introduced, consisting of paired and unpaired collections of underwater images of varying quality, captured using seven different cameras under various visibility conditions during oceanic explorations and human-robot collaborative experiments.

Several qualitative and quantitative evaluations were performed, demonstrating that the proposed model effectively learns to enhance underwater image quality from both paired and unpaired training datasets. The enhanced images improve the performance of standard models for underwater object detection, human pose estimation, and saliency prediction. These results validate the suitability of the proposed model for real-time preprocessing in the autonomy pipeline of visually-guided underwater robots\cite{9001231}.

\begin{figure}
    \centering
    \includegraphics[width=1.0\columnwidth,height=0.45\columnwidth]{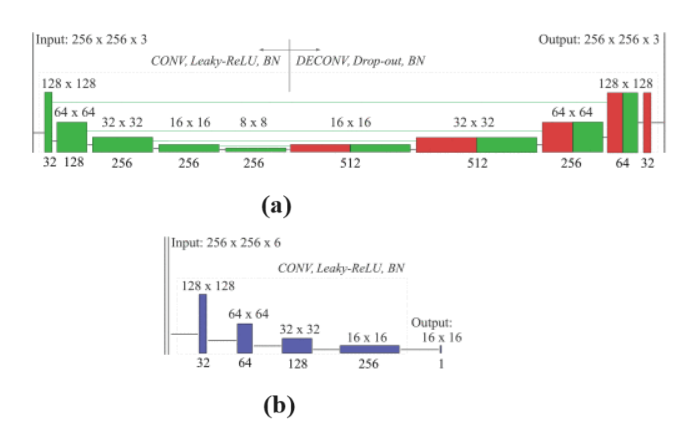}
    \caption{Network architecture of the proposed model, FUnIE-GAN, used for real-time underwater image enhancement. (a) The Generator improves image quality by focusing on global content, color, local texture, and style information. (b) The Discriminator supervises adversarial training using an objective function that evaluates perceptual image quality. This model is trained on the large-scale EUVP dataset, which includes paired and unpaired underwater images captured under various visibility conditions \cite{9001231}.}
    \label{fig:enter-label}
\end{figure}

\subsubsection{Deep- SESR}
In this paper \cite{islam2020simultaneous}, the simultaneous enhancement and super-resolution (SESR) problem for underwater robot vision is introduced and tackled, providing an efficient solution for near real-time applications. The proposed solution, Deep SESR, is a generative model based on a residual-in-residual network that learns to restore perceptual image qualities at 2x, 3x, or 4x higher spatial resolution. The model is trained using a multi-modal objective function that addresses chrominance-specific underwater color degradation, lack of image sharpness, and loss in high-level feature representation. Additionally, the model is supervised to learn salient foreground regions in the image, which guides it to enhance global contrast.

An end-to-end training pipeline is designed to jointly learn saliency prediction and SESR on a shared hierarchical feature space for fast inference. This approach ensures that the model can process images quickly, making it suitable for near real-time applications \cite{islam2020simultaneous}.

The paper \cite{islam2020simultaneous} also introduces UFO-120, the first dataset designed to facilitate large-scale SESR learning, containing over 1500 training samples and a benchmark test set of 120 samples. Experimental evaluations on UFO-120 and other standard datasets demonstrate that Deep SESR outperforms existing solutions for underwater image enhancement and super-resolution. The model's generalization performance is validated on several test cases, including underwater images with diverse spectral and spatial degradation levels and terrestrial images with unseen natural objects.

Furthermore, the computational feasibility of Deep SESR for single-board deployments is analyzed, demonstrating its operational benefits for visually-guided underwater robots. The model's ability to enhance and super-resolve images in near real-time provides significant advantages for underwater robotics, enabling more accurate and reliable visual perception in challenging underwater environments.

\subsubsection{WaterNet }
Underwater image enhancement is vital for marine engineering and aquatic robotics, but existing algorithms are mainly tested on synthetic datasets or limited real-world images. To evaluate these algorithms' real-world performance, a comprehensive perceptual study using large-scale real-world images is conducted. This study introduces the Underwater Image Enhancement Benchmark (UIEB), containing 950 real-world underwater images, with 890 having corresponding reference images and 60 considered challenging due to the lack of satisfactory references \cite{li2019underwater}.

The study also proposes Water-Net, an underwater image enhancement network trained on the UIEB \cite{8917818}. The benchmark evaluations and Water-Net demonstrate the strengths and limitations of current algorithms, providing insights for future research. This work advances the assessment and benchmarking of underwater image enhancement algorithms, contributing to the field's progress \cite{li2019underwater}.

\subsubsection{MSSCE-GAN}
Enhancing underwater images is crucial for applications such as underwater exploration. Traditional methods often rely on paired underwater and reference images for training, which are challenging to acquire. These methods frequently suffer from information loss, resulting in blurred details and limited applicability across diverse underwater conditions \cite{10455905}.

This paper \cite{10455905} introduces a novel approach using the Multi-Scale Structural and Color Enhanced Generative Adversarial Network (MSSCE-GAN) for unpaired underwater image enhancement. The method includes modules for detail feature recovery and attention enhancement, addressing various distortions prevalent in underwater imagery.

Key to this approach is its ability to generate superior enhanced images without requiring paired training data. Experimental evaluations demonstrate significant improvements over existing techniques in terms of effectiveness and generalizability across multiple underwater image datasets.

\begin{figure}
    \centering
    \includegraphics[width=0.8\columnwidth,height=0.45\columnwidth]{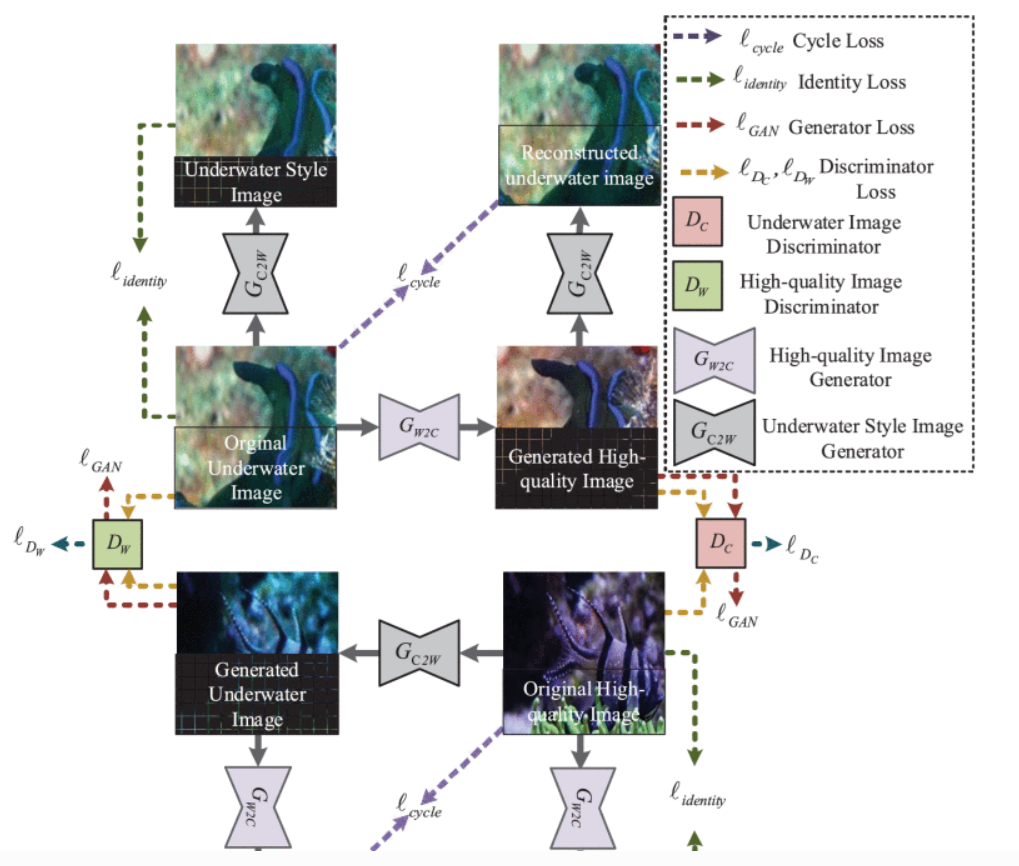}
    \caption{Network architecture of the Multi-Scale Structural and Color Enhanced Generative Adversarial Network (MSSCE-GAN) for unpaired underwater image enhancement. The model includes detail feature recovery and attention enhancement modules, generating high-quality enhanced images without paired training data. It shows significant improvements over existing methods in multiple underwater image datasets \cite{10455905}}
    \label{fig:mssc}
\end{figure}

\subsubsection{Deep WaveNet}
Underwater images typically suffer from low contrast and significant color distortions due to varying light attenuation as it travels through water. This phenomenon affects different colors asymmetrically, complicating image restoration tasks. Despite numerous attempts using deep learning for underwater image restoration (UIR), existing methods often overlook this asymmetry in network design \cite{sharma2023wavelength}.

This article introduces two novel contributions to address these challenges in UIR. Firstly, it proposes adapting receptive field sizes based on the wavelength-dependent attenuation of color channels, aiming for improved performance. Secondly, it incorporates an attentive skip mechanism to refine multi-contextual features effectively, enhancing model representational power while suppressing irrelevant features.

The proposed framework, Deep WaveNet, is optimized using pixel-wise and feature-based cost functions. Extensive experiments demonstrate its superiority over state-of-the-art methods on benchmark datasets. Furthermore, the study validates the enhanced images through various high-level vision tasks, such as underwater image semantic segmentation and diver's 2D pose estimation \cite{sharma2023wavelength}.

\begin{figure}
    \centering
    \includegraphics[width=0.8\columnwidth,height=0.45\columnwidth]{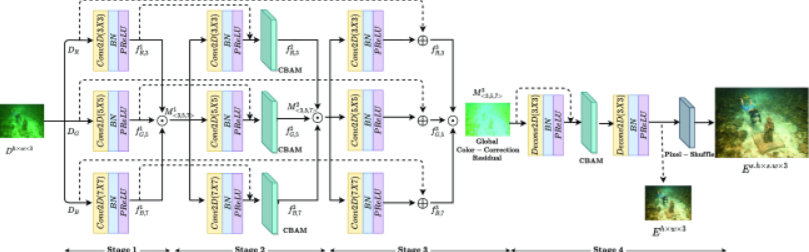}
    \caption{The proposed model aims to enhance underwater images and achieve super-resolution simultaneously. Section 3 details the integration of CBAM and Pixel-shuffle operations within the model . It accepts degraded underwater images as input and produces images that are enhanced both visually and spatially \cite{sharma2023wavelength}. }
    \label{fig:Deep wavenet}
\end{figure}

\section{Dataset}
\subsection{UIEB}

The Underwater Image Enhancement Benchmark (UIEB) dataset comprises 950 real-world underwater images, each with a size of $256 \times 256$ pixels. Among these, 890 images have corresponding reference images available for evaluation, while the remaining 60 images lack satisfactory reference images, presenting a challenge for analysis, showing in Table \ref{tab:uieb_dataset}. This dataset serves as a crucial resource for conducting comprehensive studies on underwater image enhancement algorithms, enabling both qualitative and quantitative assessments of algorithm performance.
\begin{table}[htbp]
\centering
\caption{Summary of Underwater Dataset UIEB~\cite{li2019underwater}}
\label{tab:uieb_dataset}
\begin{tabular}{|c|c|}
\hline
\textbf{Dataset Characteristics}      & \textbf{Details}                                                                                                       \\ \hline
Number of Real-world Images           & 950                                                                                                                    \\ \hline
Number of Images with Reference       & 890                                                                                                                    \\ \hline
Number of Challenging Images          & 60                                                                                                                     \\ \hline
\end{tabular}
\end{table}

\begin{figure}[ht!]
\centering
\includegraphics[width=1.0\columnwidth,height=0.45\columnwidth]{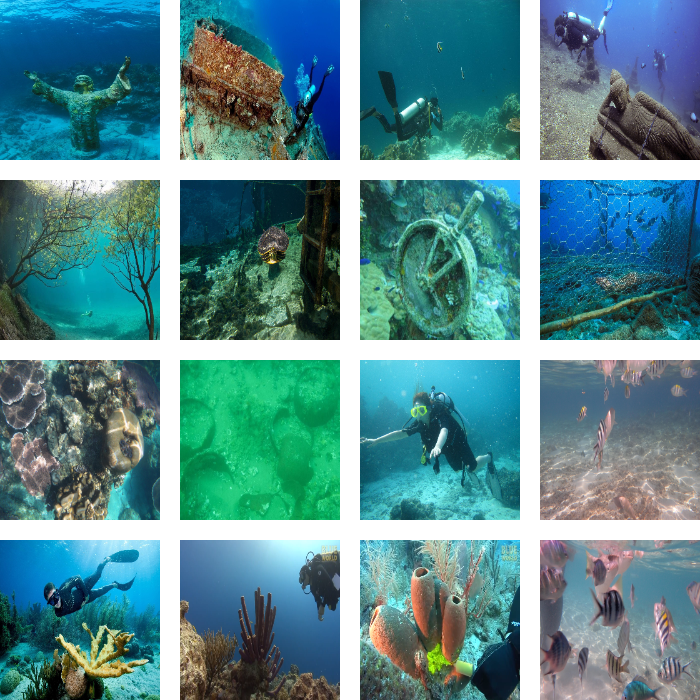}
\caption{Sample images from the UIEB dataset, showcasing a diverse range of underwater scenes. }
\label{fig:UIEB-Data}
\end{figure}

\subsection{EUVP}
\subsubsection{\textbf{Paired Dataset}}

\textbf{Underwater Dark:}

This dataset comprises 5550 pairs of images for training, each with a size of $256 \times 256$ pixels. Each pair consists of two images, one contains poor-quality or gray images, and the other contains enhanced or colored images. The filenames for each pair are identical. Additionally, 570 images are set aside for validation. In total, the dataset contains 11,670 images as shown in Table \ref{tab:underwater_datasets} .

\begin{figure}[ht!]
\centering
\includegraphics[width=1.0\columnwidth,height=0.45\columnwidth]{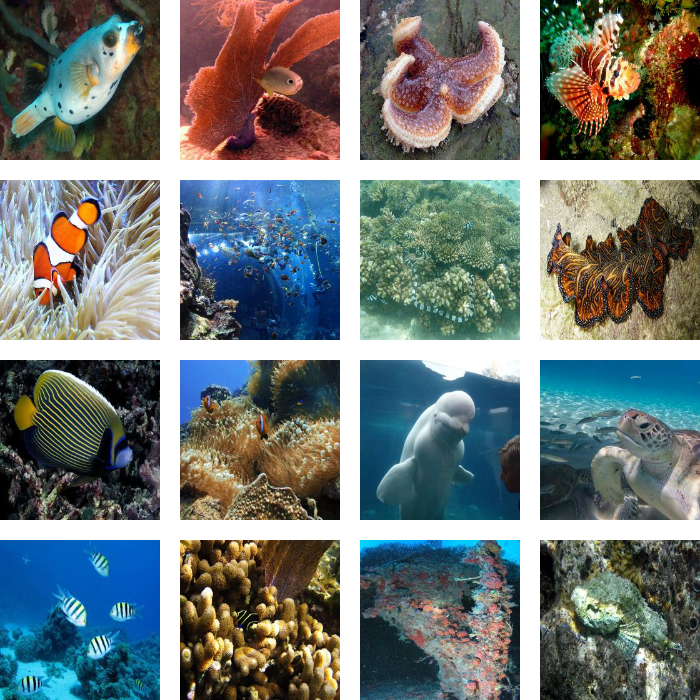}
\caption{Examples of images from the EUVP\_Underwater\_Dark dataset}
\label{fig:EUVP_Dark_data}
\end{figure}
\textbf{Underwater ImageNet:}

The Underwater ImageNet dataset consists of 3700 pairs of images for training, each with a size of $256 \times 256$ pixels. Similar to the Underwater Dark dataset, one contains poor-quality images and the other contains enhanced or better-quality images. The filenames for corresponding pairs match. The dataset also includes 1270 images for validation, resulting in a total of 8670 images.

\begin{figure}[ht!]
\centering
\includegraphics[width=1.0\columnwidth,height=0.45\columnwidth]{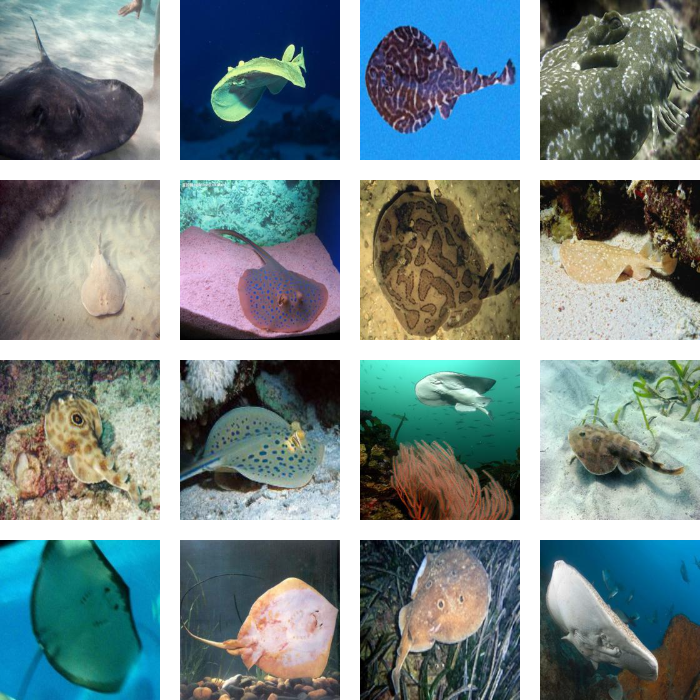}
\caption{Sample images from the EUVP\_Underwater\_Imagenet dataset}
\label{fig:EUVP_Dark_data}
\end{figure}

\textbf{Underwater Scenes:}

This dataset comprises 2185 pairs of images, each with a size of $320 \times 240$ pixels for training, with each pair containing a poor-quality image and a corresponding enhanced or better-quality image. The filenames for corresponding pairs are consistent. Additionally, 130 images are allocated for validation purposes. In total, the dataset encompasses 4500 images.

\begin{figure}[ht!]
\centering
\includegraphics[width=1.0\columnwidth,height=0.45\columnwidth]{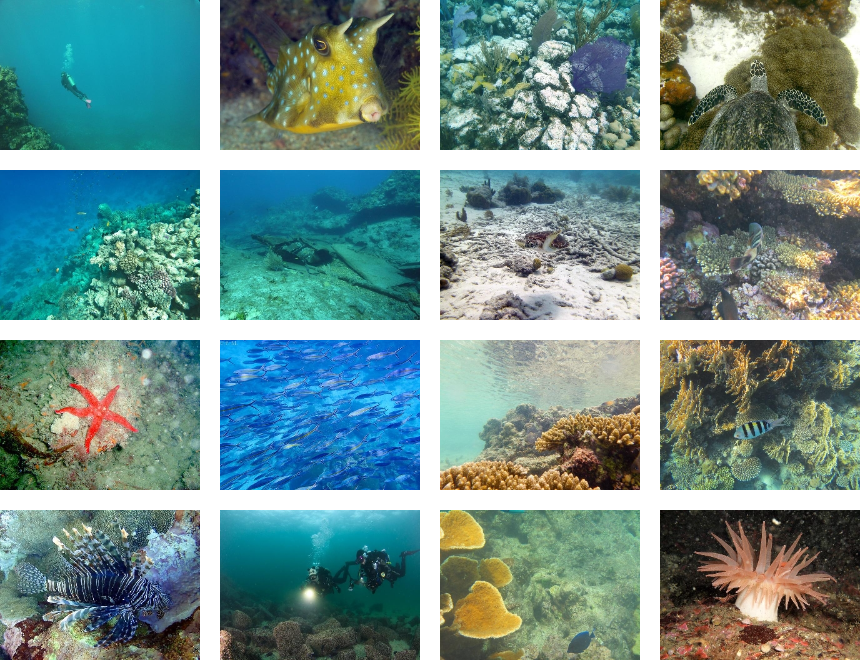}
\caption{Sample images from the EUVP\_Underwater\_Scenes dataset}
\label{fig:EUVP_Dark_data}
\end{figure}

\begin{table}[htbp]
\centering
\caption{Summary of Underwater Datasets EUVP (paired data) ~\cite{9001231}}
\label{tab:underwater_datasets}
\begin{tabular}{|c|c|c|c|}
\hline
\textbf{Dataset Name} & \textbf{Training Pairs} & \textbf{Validation} & \textbf{Total Images} \\ \hline
\textbf{Underwater Dark }      & 5550 pairs               & 570                  & 11670                 \\ \hline
\textbf{Underwater ImageNet}   & 3700 pairs               & 1270                 & 8670                  \\ \hline
\textbf{Underwater Scenes }    & 2185 pairs               & 130                  & 4500                  \\ \hline
\end{tabular}
\end{table}

\subsubsection{\textbf{Unpaired Data}}
In the dataset for unpaired training, there are 3195 images representing poor quality images, while the set comprises 3140 images of enhanced or better quality. These images come in sizes of $960\times540$, $640\times480$, and $320\times240$ pixels. Additionally, there are 330 images allocated for validation purposes. These images are not paired, meaning that there is no one-to-one correspondence between the poor-quality and enhanced-quality images. This dataset arrangement allows for the training of models aimed at enhancing image quality without relying on direct paired examples as shown in Table \ref{tab:unpaired}.

\begin{figure}[!h]
\centering
\includegraphics[width=1.0\columnwidth,height=0.45\columnwidth]{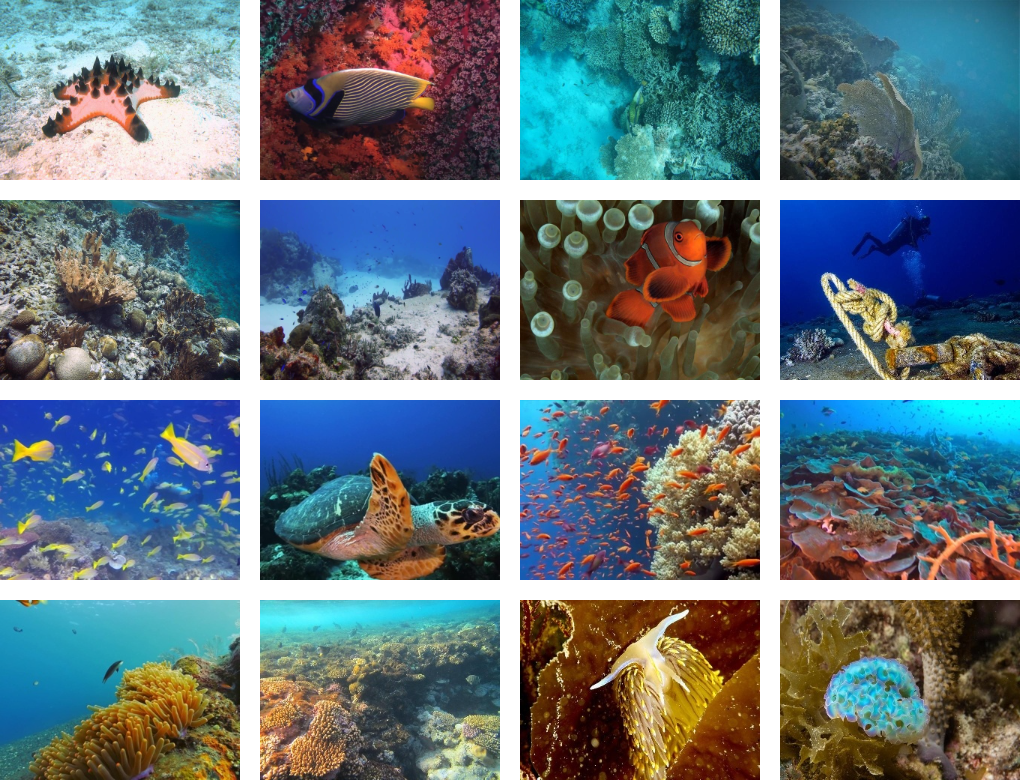}
\caption{Sample images from the EUVP\_Unpaired dataset}
\label{fig:EUVP_Unpaired}
\end{figure}

\begin{table}[!h]
\caption{Distribution of images in the dataset (Unpaired data)}
    \label{tab:unpaired}
    \centering
    \begin{tabular}{|c|c|c|c|}
        \hline
        \textbf{Poor quality} & \textbf{Good quality} & \textbf{Validation} & \textbf{ Total Images} \\ 
        \hline
        3195 & 3140 & 330 & 6665 \\ 
        \hline
    \end{tabular}
\end{table}

\section{Performance Evaluation}
Underwater image quality assessment is a challenging task that is used to evaluate the 
quality of the image accurately and automatically. Image quality assessment (IQA) methods are employed to automatically evaluate the quality of images. IQA approaches are 
broadly classified into (a) objective and (b) subjective image quality assessment. Subjective image quality assessments are 
expensive and time-consuming and hence not suitable for real-time applications. Objective 
assessment techniques use statistical and mathematical models based on the human visual system (HVS) to automatically estimate image quality.
Based on the availability of the original image, objective IQA methods can be classified 
into three categories (1) full reference IQA (FR) where the reference image is available, (2) reduced reference IQA (RR) where partial information of 
the reference image is available and (3) no reference IQA (NR) in which the reference image is 
not available. In addition to the standard performance evaluation parameters, to assess 
underwater image quality effectively, specialized metrics are proposed in the literature.

The performance of various underwater image enhancement and restoration techniques is analyzed using different qualitative and quantitative parameters. The qualitative evaluation involves the visual enhancement of the image by comparing histograms. 
The quantitative performance framework deals with various quality metric parameters 
which include:
\begin{itemize}
\item \textbf{Mean square error (MSE)}: MSE computes the cumulative squared error between the 
enhanced and the original image. The lower the MSE, the better the quality (low error) and 
is given as:

\begin{equation}
    MSE =\frac{1}{MN} \sum_{i=1}^{M} \sum_{j=1}^{N} \big[F(i,j)-E(i,j)\big]^2
\end{equation}
where F(i, j) is the original image, E(i, j) is the enhanced image, and M × N is image size.
\item  \textbf{Peak-signal-to-noise ratio (PSNR):} : It is the measure of the peak error and computed 
as 

\begin{equation}
    PSNR= 20 \log_{10} \Big(\frac{MAX_{F}}{\sqrt{MSE}}\Big)
\end{equation}

where maximum pixel value of the image is represented by $MAX_{F}$ and is 255 for gray 
level image.

\item \textbf{Entropy :} Entropy is a measure of information content present in the image and is 
given as: 

\begin{equation}
    H(F)= - \sum_{i=0}^{255} p_{i} \log_2 p_{i}
\end{equation}

where $p_{i}$ is the probability of occurrence of intensity i at a pixel in image F

\item \textbf{Structure similarity index measure (SSIM):}  SSIM measures the similarity between 
original image patches and enhanced patches at locations x and y from three aspects: 
brightness, contrast, and structure

\begin{equation}
    SSIM(F,E) = \frac{\big( 2 \mu_x\mu_y +C_1\big)\big(2\sigma_{xy}+C_2\big)}{\big(\mu_{x}^2 + \mu_{y}^2+C_1\big)\big(\sigma_{x}^2 + \sigma_{y}^2+C_2 \big)}
\end{equation}

where  $\mu_x, \mu_y$ are the mean values and $\sigma_x, \sigma_y$ are the standard deviation values of 
the pixels in patch x and y respectively.  $\sigma_{xy}$ is the covariance of patches x and y and 
$C1 = (k_1L)^2$ and $C2 = (k_2L)^2$
are small constants to avoid instability while the denominator is close to zero. L is the dynamic range of pixel values, $k_{1} = 0.01$ and $k_{2}$ = 0.03. The 
higher the SSIM value, the smaller the distortion and the better the enhancement.

\item \textbf{Colour enhancement factor (CEF):}  It helps in the representation of the effect of 
enhancement and is given as

\begin{equation}
    CEF=\frac{CM(\tilde{I)}}{CM(I)}
\end{equation}

$CM(I)=\sqrt{\sigma_\alpha^2 +\sigma_\beta^2}+0.3\sqrt{\mu_\alpha^2+\mu_\beta^2}$ where $\sigma_\alpha^2 , \sigma_\beta^2$ represent the standard 
deviations and $\mu_\alpha^2$ and $\mu_\beta^2$  are the average values of $\alpha$ and $\beta$ respectively. $CM(\tilde{I})$ is used to 
denote enhanced image and CM(I) the original image.

\item \textbf{Contrast to noise ratio (CNR):} This metric describes the amplitude of the signal relative to the surrounding noise in an image. CNR is computed by using 

\begin{equation}
    CNR(I,I')=\frac{(\mu_i - \mu_n)}{\sigma_n}
\end{equation}

$\mu_i$ represents the mean value of original image and $\mu_n$ is mean value of enhanced image and $\sigma_n$ denotes the standard deviation.

\item \textbf{Image enhancement metric (IEM):} This metric gives information about the sharpness and the improvement in the contrast after the process of enhancement. It is 
computed as follows
\begin{equation}
    IEM = \frac{\sum_{l=1}^{k1}\sum_{m=1}^{k2}\sum_{n=1}^8|I_{e,c}^{m,l}-I_{e,n}^{m,l}|}{\sum_{l=1}^{k1}\sum_{m=1}^{k2}\sum_{n=1}^8|I_{o,c}^{m,l}-I_{o,n}^{m,l}|}
\end{equation}

k1 and k2 denote the non-overlapping blocks. o and e represent the original and 
enhanced images respectively. The intensities of the centre pixel is denoted by $I_{o,c}^{m,l},I_{e,c}^{m,l}, I_{e,n}^{m,l}, I_{o,n}^{m,l}$
 are the intensities of the neighbours from the centre pixel.
 
\item \textbf{Absolute mean brightness error(AMBE):} AMBE helps to compute the brightness 
content that is preserved after the process of image enhancement. It is given as 

\begin{equation}
    AMBE(o,e)=|\mu_o-\mu_e|
\end{equation}

where F(i, j) is the original image, E(i, j) is the enhanced image, and M × N is the image 
size, the equation represents the absolute difference between the mean of original $\mu_o$ and 
enhanced images $\mu_e$. Median values of the AMBE metric indicate good preservation of 
brightness

\item \textbf{Spatial spectral entropy based quality index (SSEQ):} SSEQ is a highly efficient no reference (NR) IQA model proposed by. SSEQ can assess the quality of an image that is distorted across various distortion categories. SSEQ can be 
calculated by 
\begin{equation}
    E = - \sum_{i} \sum_{j} P_{i,j} \log_2 P_{i,j} 
\end{equation}

where P(i, j) is the spectral probability map given as

\begin{equation}
    P(i,j)=\frac{C(i,j)^2}{\sum_i \sum_j C(i,j)^2}
\end{equation}
C is a coefficient matrix computed on (i,j) pixels.

\item \textbf{Measure of enhancement (EME):} EME calculates the contrast of the images and aids 
in the optimum selection of processing parameters. It is computed as:
\begin{equation}
    EME_{m_1m_2}= max \Big(\frac{1}{m_1m_2}\sum_{l=1}^{m_1} \sum_{n=1}^{m_2}20 log \frac{X_{max;n,l}^{\omega}}{X_{min;n,l}^{\omega}} \Big)
\end{equation}

where $X_{max;n,l}^{\omega} and {X_{min;n,l}^{\omega}}$ represent the maximum value and minimum value of the 
image within the block $\omega_{n,l}$

\item \textbf{Root mean square error (RMSE)}: RMSE is computed by calculating the square root of MSE. It is given as 

\begin{equation}
    RMSE=\sqrt{\frac{1}{MN}\sum_{i=1}^M \sum_{j=1}^N \big[F(i,j)-E(i,j)]^2}
\end{equation}

\item \textbf{Measure of enhancement by entropy (EMEE):} EMEE is computed by
\begin{equation}
    EMEE_{m_1m_2}= max \Big(\frac{1}{m_1m_2}\sum_{l=1}^{m_1} \sum_{n=1}^{m_2}\alpha\frac{X_{max;n,l}{\theta}^\alpha}{X_{min;n,l}{\theta}} \frac{X_{max;n,l}{\theta}}{X_{min;n,l}{\theta}} \Big)
\end{equation}

Good image quality is indicated by the high value of EMEE. m1 and m2 represent the 
blocks in which the image is divided.

\item \textbf{Underwater color image quality evaluation metric (UCIQE):} UCIQE was specifically designed to quantify the effects of non-uniform color cast, low contrast and 
issues of blurring that affect underwater images. UCIQE for an image X in CIELab space is calculated as:

\begin{equation}
    UCIQE = c1 * \sigma_{chroma} + c2 * {contrast_l} + c3 * \mu_{saturation}
\end{equation}

where c1 c2 c3 represents the weighted coefficients, $\sigma_{chroma}$ denotes the standard deviation, $contrast_l$ is the contrast and the average value of saturation is denoted by 
$\mu_saturation$. Higher values of UCIQE signify that the image 
has good equilibrium among chroma, contrast, and saturation.

\item \textbf{Underwater Image Colorfulness Measure (UICM):} Underwater images often exhibit a color-casting problem wherein colors are gradually attenuated based on their wavelength as the water depth increases. The color red, which has the shortest wavelength, disappears first, resulting in a bluish or greenish appearance of the images. In addition, inadequate lighting conditions can also lead to significant color de-saturation. To address this, an effective algorithm for enhancing underwater images must ensure good color rendition. The human visual system (HVS) captures colors in the opponent color plane, and hence, the chrominance RG and YB components, which are associated with the two opponent color planes, are utilized in the UICM technique as illustrated in the reference.
\begin{equation}
    RG-R-G
\end{equation}

\begin{equation}
    YB=\frac{R+G}{2}-B
\end{equation}

Due to the heavy noise in underwater images, the traditional statistical values are not suitable for measuring their colorfulness. As a result, asymmetric alpha-trimmed statistical values are used instead. The mean can be expressed as:
\begin{equation}
    \mu_{\alpha,RG}=\frac{1}{K-T_{\alpha L}- T_{\alpha R}} \sum_{i=T_{\alpha L+1}}^{K-T_{\alpha R}}intensity_{RG,i}
\end{equation}

The second-order statistic variance $\sigma^2$ in:
\begin{equation}
    \sigma^2_{\alpha, RG}=\frac{1}{N}\sum_{p=1}^N (Intensity_{RG,p}-\mu_{\alpha, RG})^2
\end{equation}

The overall colorfulness metric used for measuring underwater image colorfulness is demonstrated in
\begin{equation}
    UICM=-0.2868\sqrt{\mu^2_{\alpha, RG}+\mu^2_{\alpha, YB}}+0.1586\sqrt{\sigma^2_{\alpha, RG}+\sigma^2_{\alpha, YB}}
\end{equation}

\item \textbf{ Underwater Image Sharpness Measure (UISM):}
Sharpness pertains to the quality of preserving fine details and edges in an image. In underwater images, forward scattering often causes significant blurring, resulting in a loss of image sharpness. To quantify sharpness on edges, the Sobel edge detector is initially applied to each RGB color component, and the resulting edge map is multiplied with the original image to generate a grayscale edge map. This preserves only the pixels on the edges of the original underwater image. To measure the sharpness of these edges, the enhancement measure estimation (EME) method is suitable for images with uniform backgrounds and exhibits non-periodic patterns. Hence, EME is utilized to calculate the sharpness of edges. The UISM is:
\begin{equation}
    UISM=\sum_{c=1}^3 \lambda_c EME(grayscale   edge_c)
\end{equation}

\begin{equation}
    EME=\frac{2}{k1k2}\sum_{l=1}^{k1}\sum_{k=1}^{k2} \log \frac{I_{max,k,l}}{I_{min,k,l}}
\end{equation}

\item \textbf{ Underwater Image Contrast Measure (UIConM):}
Studies have demonstrated a correlation between contrast and underwater visual capabilities, including stereoscopic acuity. In the case of underwater imagery, contrast deterioration is typically attributed to backward scattering. The intensity image is evaluated using the logAMEE measure to determine the contrast.
\begin{equation}
    UIConM =\log AMEE(intensity)
\end{equation}
The logAMEE in

\begin{equation}
    logAMEE = \frac{2}{k1k2}\sum_{l=1}^{k1}\sum_{k=1}^{k2}\frac{I_{max,k,l}, \ominus I_{min,k,l}}{I_{max,k,l} \oplus I_{min,k,l}} * \log \frac{I_{max,k,l}, \ominus I_{min,k,l}}{I_{max,k,l} \oplus I_{min,k,l}}
\end{equation}

\item \textbf{Underwater image quality measure (UIQM): }UIQM is based on the human visual system model and works without a reference image. UIQM 
comprises three main measures, UICM the underwater image colorfulness measure, UISM the underwater image sharpness measure, and UIConM the underwater 
image contrast measure. UIQM is calculated as follows:

\begin{equation}
    UIQM = Coeff_1 * UICM + Coeff_2 * UISM + Coeff_3
* UIConM
\end{equation}

Higher values of UIQM indicate good levels of enhancement.

\item \textbf{Colourfulness contrast fog density index (CCF):} No-reference IQA method is proposed to predict underwater color image quality. using CCF 
metric. CCF metric is a weighted combination of colorfulness index, contrast index, and fog density index which is computed as,
\begin{equation}
    CCF = \omega_1 * Colorfulness + \omega_2 * Contrast + \omega_3 * Fogdensity
\end{equation}

Colorfulness index due to absorption, blurring because of forward scattering and fog 
density due to backward scattering is examined in the CCF computation.

\item\textbf{Average gradient (AG):} Average gradient is a full reference metric that is used to 
define the sharpness of the given image. It represents the change in the rate of minute 
details present in the image. It is computed as, 

\begin{equation}
    AG=\frac{1}{(L-1)(M-1)} \sum_{i=1}^{L-1} \sum_{j=1}^{M-1} \sqrt{\big(\nabla_x I(i,j)\big)^2 + \sqrt{\big(\nabla_y I(i,j)\big)}}
\end{equation}

where L and M denote the width and height of the image and $\nabla_x$ and $\nabla_y$ represent the 
the gradient in the x and y directions respectively \cite{raveendran2021underwater}.

\item\textbf{Patch based contrast quality index (PCQI): } PCQI is defined as,
\begin{equation}
    PCQI(i,j)=\frac{1}{P}\sum_{k=1}^{P} l_r(i_k,j_k)l_s(i_k,j_k)l_t(i_k,j_k)
\end{equation}

where P is the number of patches present in the image and $l_r$, $l_s$, and $l_t$
 represent the 
comparison functions. Higher values of PCQI indicate good contrast.
\end{itemize}

In this section,  conducted experiments on two datasets, UIEB and EUVP, to evaluate the performance of various underwater image enhancement methods in terms of both qualitative and quantitative metrics. The UIEB dataset comprises 890 real underwater images, while the EUVP dataset contains paired and unpaired compilations of underwater images.  selected five images from each dataset for evaluation purposes.  used several typical methods for underwater image enhancement, including AHE, CLAHE, ICM, UCM, Gray World, Wavelet fusion, and the Recursive adaptive histogram modification method.

\section{Datasets: UIEB-D8 and EUVP-X-D8}
This work has used the two standard datasets UIEB~\cite{li2019underwater} and EUVP~\cite{9001231} that are available in the public domain and are widely used in UIE research. The UIEB dataset has 890 paired images where each pair consists of a good quality image along with a degraded one. EUVP dataset has both paired and unpaired images. EUVP has three different paired datasets -- Underwater Dark, Underwater ImageNet and Underwater Scenes.

\subsection{Formation of Datasets}
To diversify the dataset, 8 different degradation techniques were applied to the ground truth images:
\subsubsection{Illumination Degradation}
\begin{figure}[!h]
    \centering
    \includegraphics[width=1.0\columnwidth,height=0.35\columnwidth]{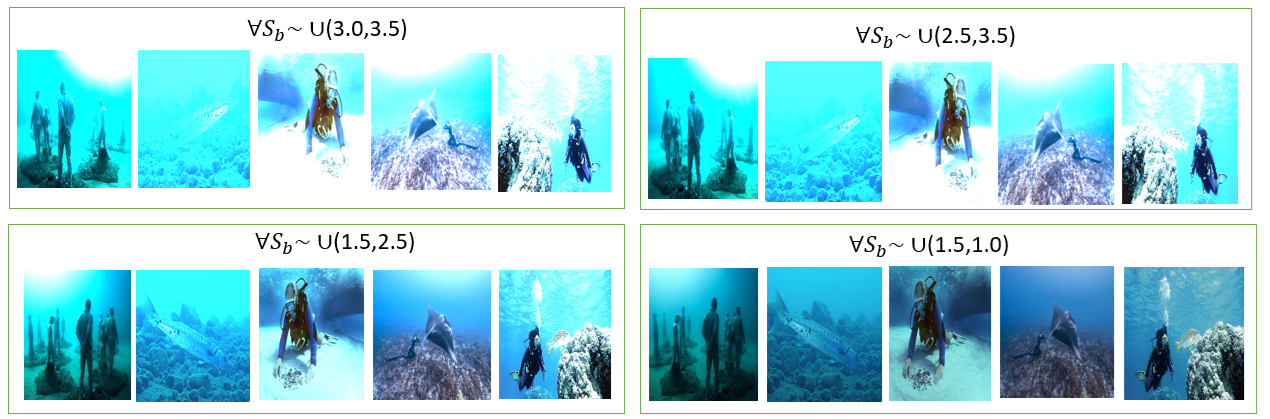}
    \includegraphics[width=1.0\columnwidth,height=0.35\columnwidth]{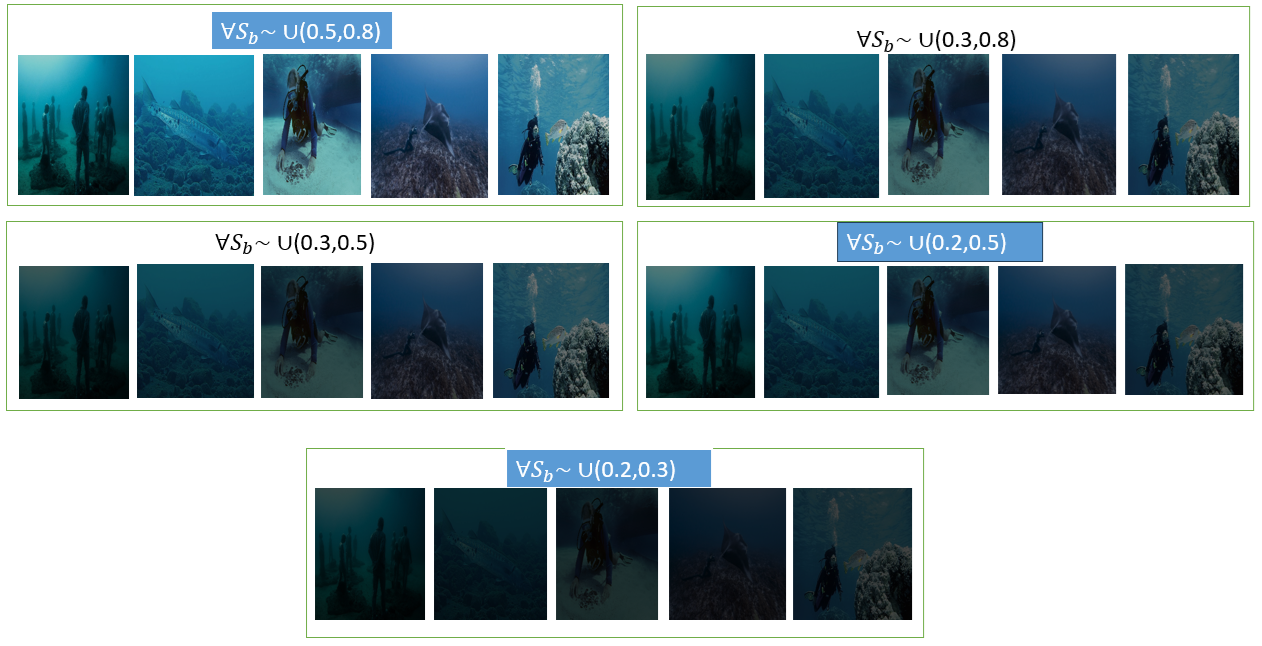}
    \caption{Image exhibiting illumination degradation, characterized by low lighting and diminished clarity due to varying levels of illumination conditions}
    \label{fig:Aba_Ill}
\end{figure}

Low illumination in images can result from various factors such as poor lighting conditions. Simulating low illumination is crucial for testing the robustness of image processing algorithms in real-world scenarios. The degradation is achieved by reducing the overall brightness of the image, mimicking the effect of dim lighting conditions. This reduction in brightness can lead to loss of details and visibility of objects in the image and variation of illumination shown in Fig \ref{fig:Aba_Ill}.

The equation to simulate low illumination is as follows:
\begin{equation}
    I_{ID} (x,y) = s_b \times I (x,y), \quad \forall s_b \sim \cup(a, b)
\end{equation}
where $I_{ID}$ is the modified image after applying varying illumination. This factor determines the extent of brightness reduction. A lower value results in a darker image.

\vspace{0.2cm}

\subsubsection{Contrast Degradation}
High contrast simulates images with intense differences between light and dark areas. This effect is achieved by adjusting the pixel values to increase the contrast.The equation for increasing contrast is:

 \begin{equation}
     I_{CD}(x,y) = \alpha \times I (x,y) + \beta, \quad \forall \alpha \sim U(a, b), \quad \beta = m 
 \end{equation}

\begin{figure}[!h]
    \centering
    \includegraphics[width=1.0\columnwidth,height=0.35\columnwidth]{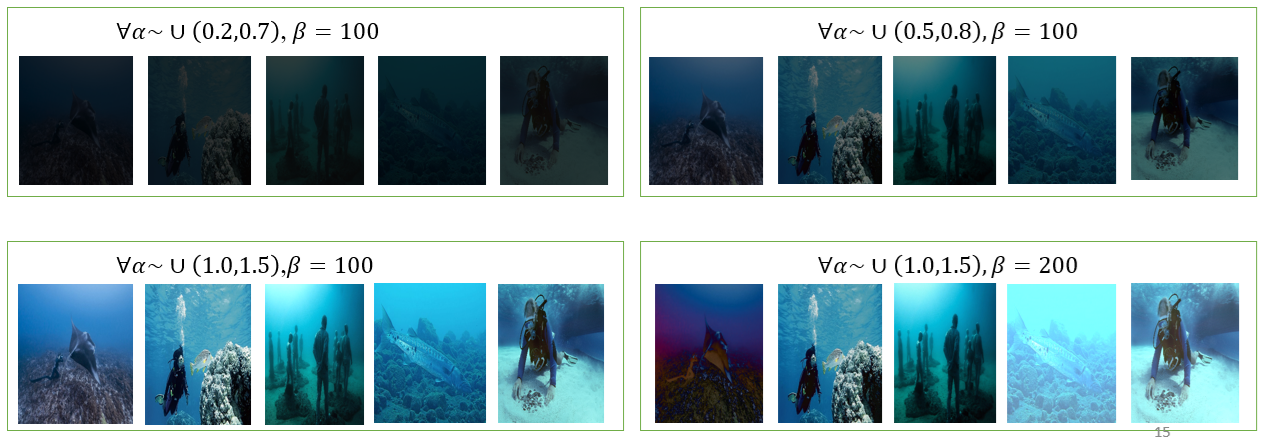}
\includegraphics[width=1.0\columnwidth,height=0.35\columnwidth]{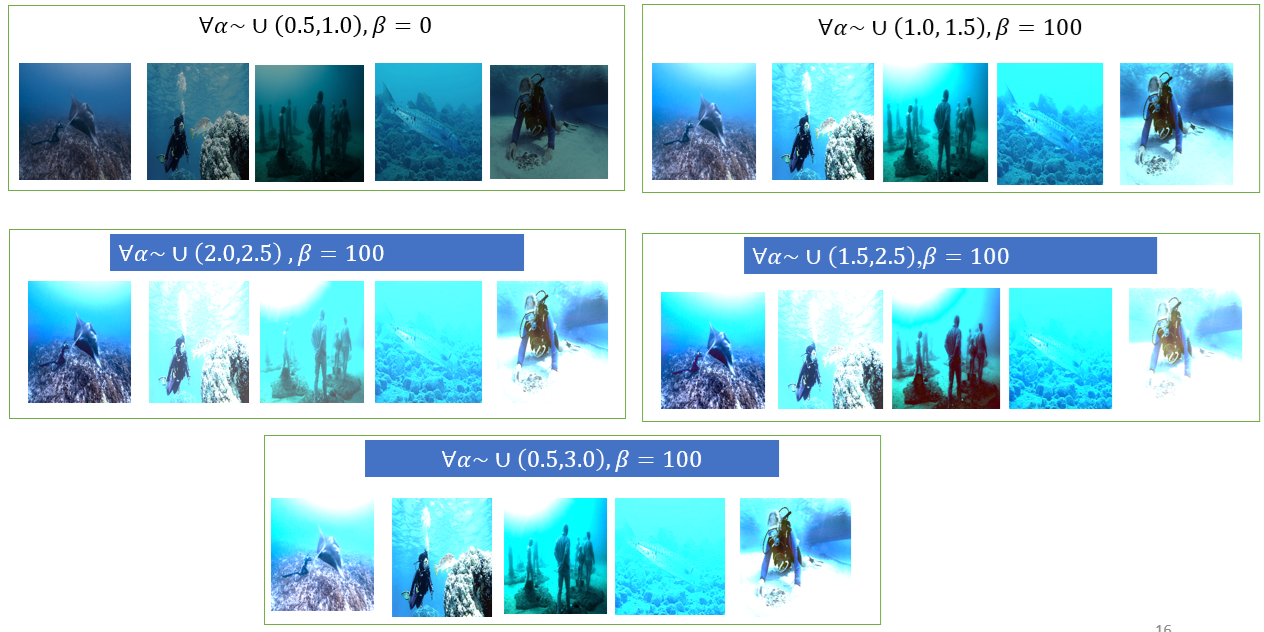}
    \caption{Image showcasing high contrast degradations. The pronounced contrast creates sharp distinctions, intensifying visual impact while potentially causing loss of detail in certain areas}
    \label{fig:Aba_con}
\end{figure}

By multiplying the pixel values by alpha and adding beta, the contrast of the image is increased while also adjusting its brightness as shown in Fig \ref{fig:Aba_con}. This results in an image $I_{CD}(x,y)$ with intensified differences between light and dark areas, creating a high contrast effect.

\vspace{0.2cm}

 \subsubsection{Hazy Degradation}
The hazy effect simulates the presence of haze or fog in the image. It is achieved by adding a semi-transparent haze layer over the original image.
The mathematical expression for applying the haze effect to the image is:
    \begin{equation}
        I_{DH}(x,y) = (1 - \gamma) \times I(x,y) + \gamma \times \gamma_c(x,y), \quad \forall \gamma \sim U(a, b), \quad \forall \gamma_c \sim U(l, m)
    \end{equation}
    
Here, $\gamma_L(x,y)$ creates a haze layer with the same dimensions as the degraded image, where each pixel is set to the randomly generated haze color.

\begin{figure}
    \centering
    \includegraphics[width=1.0\columnwidth,height=0.35\columnwidth]{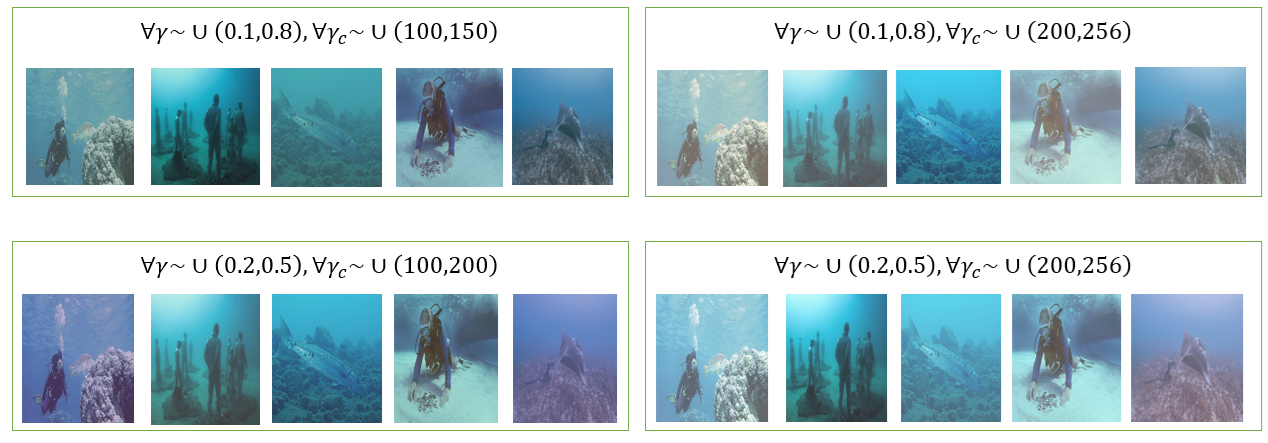}
\includegraphics[width=1.0\columnwidth,height=0.35\columnwidth]{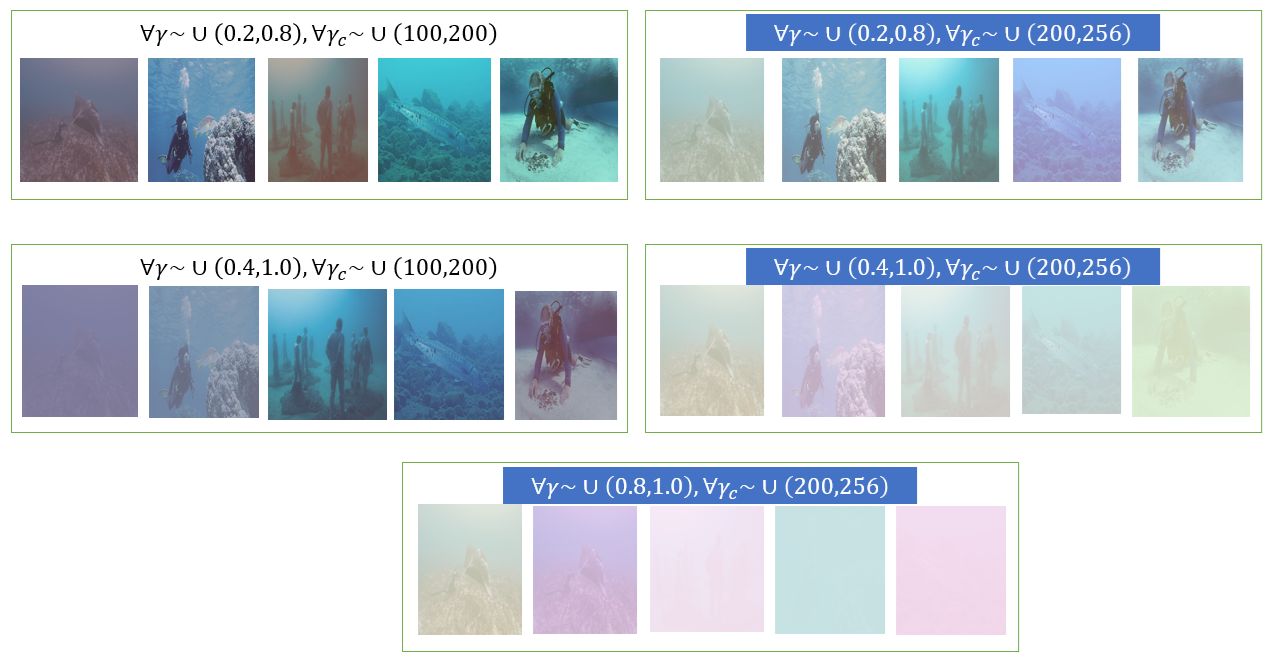}
    \caption{Image showing hazy degradations, the scene appears obscured due to atmospheric haze, resulting in reduced visibility and loss of fine details.}
    \label{fig:Aba_hazy}
\end{figure}

Blend the original image and the haze layer using the formula above that generate $I_{DH}$ . The original image is multiplied by $(1 - \gamma)$ to reduce its intensity, and the haze layer is multiplied by $\gamma$ to control the strength of the haze effect as shown in Fig \ref{fig:Aba_hazy}.
The resulting image represents the original scene with the added haze effect.
This process mimics the visual appearance of images captured in hazy conditions, where distant objects appear less distinct due to scattering of light by haze particles in the atmosphere.

\vspace{0.2cm}

\subsubsection{Blurry Degradation}
\begin{figure}[!h]
    \centering
    \includegraphics[width=1.0\columnwidth,height=0.35\columnwidth]{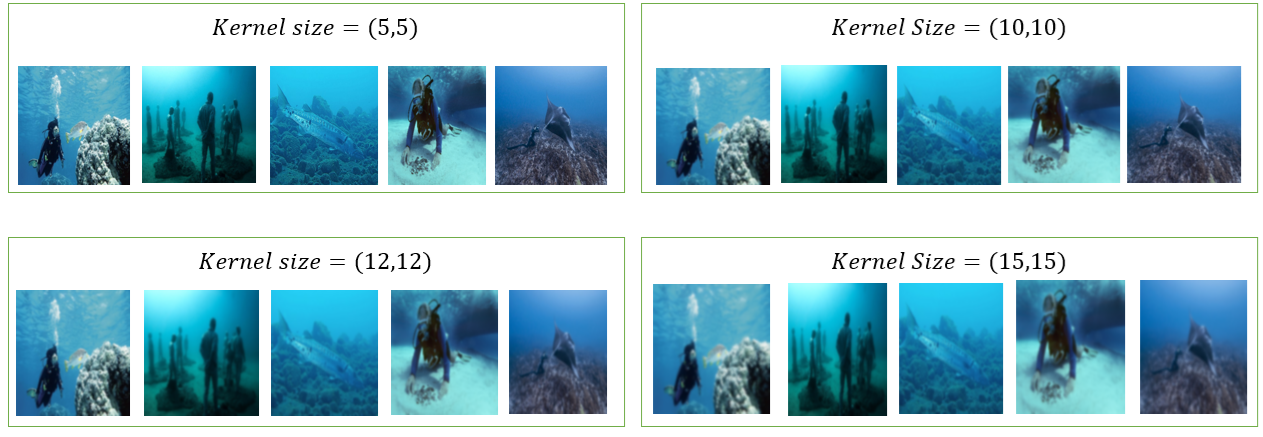}
\includegraphics[width=1.0\columnwidth,height=0.35\columnwidth]{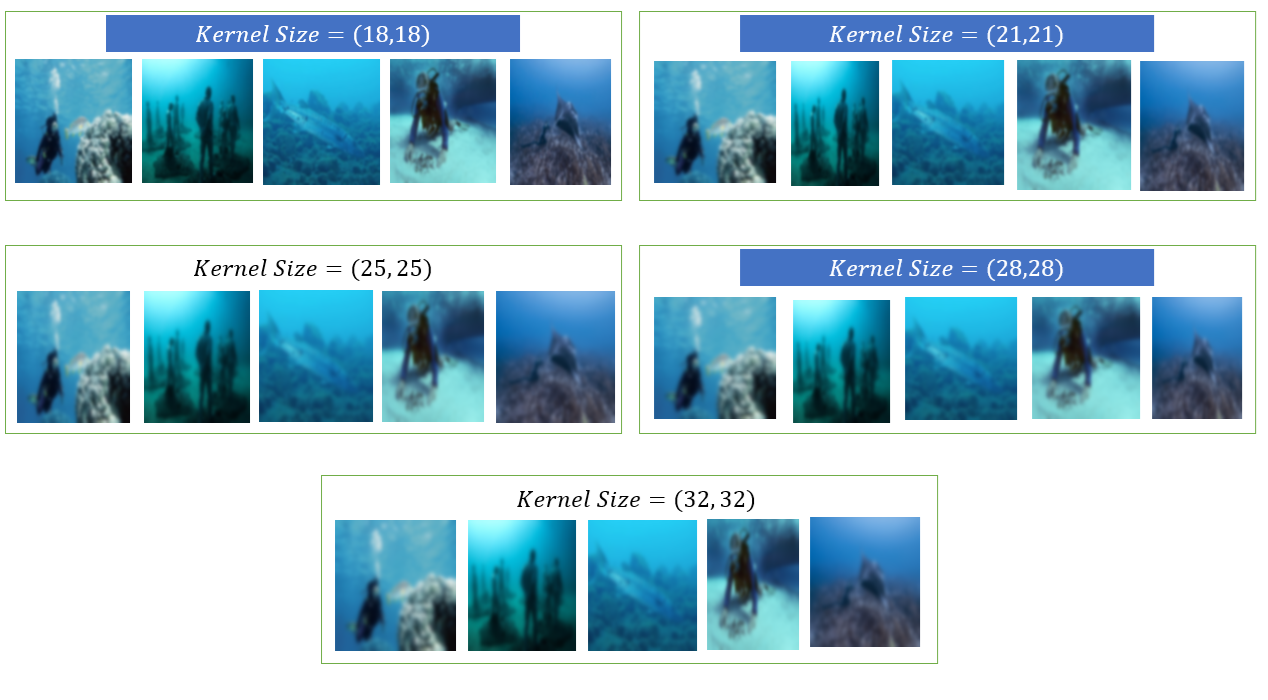}
    \caption{Blurry Degradatiom}
    \label{fig:Aba_Blurry}
\end{figure}

The blurry effect is simulated using a Gaussian blur filter applied to the entire image. The equation for applying Gaussian blur to an image is as follows:
\begin{equation}
    G(x, y) = \frac{1}{2\pi \sigma^2} \exp\left(-\frac{x^2 + y^2}{2\sigma^2}\right)
\end{equation}
where $(x,y)$ are the coordinates in the kernel, and 
$\sigma$ is the standard deviation of the Gaussian distribution. The Gaussian kernel $G(x, y)$ is typically normalized so that the sum of all elements equals 1.
The convolution operation between the input image $I$ and the Gaussian kernel $G$ is denoted by $I * G $. It's defined as:
\begin{equation}
    ( I * G )(x,y)
 = \sum_{i=-k}^{k} \sum_{j=-l}^{l} I(i,j) \cdot G(x-i,y-j) 
\end{equation}
where $(x,y)$ are the coordinates of the output pixel, 
$(i,j)$ are the coordinates of the input pixel, and $I(i,j)$ is the intensity of the input pixel. The convolution operation involves sliding the Gaussian kernel over each pixel of the input image and computing a weighted sum of pixel intensities in the neighborhood defined by the kernel as shown in Fig \ref{fig:Aba_Blurry}.
$I_{DB} (x,y)$ is blurred image and defined as: 
\begin{equation}
    I_{DB} (x,y) = (I * G )(x,y)
\end{equation}
The GaussianBlur function convolves the image with a $G(x,y)$ to compute the blurred result. The standard deviation of the $G(x,y)$ is implicitly determined by the kernel size. Overall, the Gaussian blur operation smooths out the sharp transitions between pixel values in the input image, resulting in a blurred version of the original image. The degree of blurring is controlled by the standard deviation $\sigma$ of the Gaussian kernel, with larger values of $\sigma$ resulting in more significant blurring.

\vspace{0.2cm}

\subsubsection{Noisy Degradation}  It is modeled as Gaussian (normal) distribution and is added to the pixel values of the image to simulate the effect of random fluctuations in the image acquisition process or transmission. Mathematically, Gaussian noise $\mathcal{N}(x, \mu, \sigma)$ can be expressed as:

\begin{equation}
    \mathcal{N}(x,y;\mu, \sigma) = \frac{1}{2\pi\sigma^2} e^{-\frac{(x^2 + y^2)^2}{2\sigma^2}}
\end{equation}
$x$ is the random variable representing the noise amplitude. $\mu$ is the mean (average) of the distribution, indicating the central tendency of the noise values. It's typically set to 0 for zero-mean noise. $\sigma$ is the standard deviation of the distribution, which controls the spread or variability of the noise values around the mean. It determines the scale of the noise.

Additive Gaussian noise $\mathcal{N}(0,1)$ is the pixel-wise Gaussian noise at coordinates (x,y), drawn from a Gaussian distribution with mean $\mu=0$ and standard deviation $\sigma=1$.
The degraded image $I_{ND}(x,y)$ resulting from the addition of Gaussian noise to the original image, can be mathematically represented as: The function \( I_{ND}(x,y) \) is defined as:
\begin{equation}
I_{ND}(x,y) =
\begin{cases}
    0 & \text{if } I(x,y) + N(x,y) < 0 \\
    I(x,y) + N(x,y) & \text{if } 0 \leq I(x,y) + N(x,y) \leq 255 \\
    255 & \text{if } I(x,y) + N(x,y) > 255
\end{cases}
\end{equation}
where \( N(x,y) \sim \mathcal{N}(0, \sigma^2) \).

\begin{figure}[!h]
    \centering
    \includegraphics[width=1.0\columnwidth,height=0.35\columnwidth]{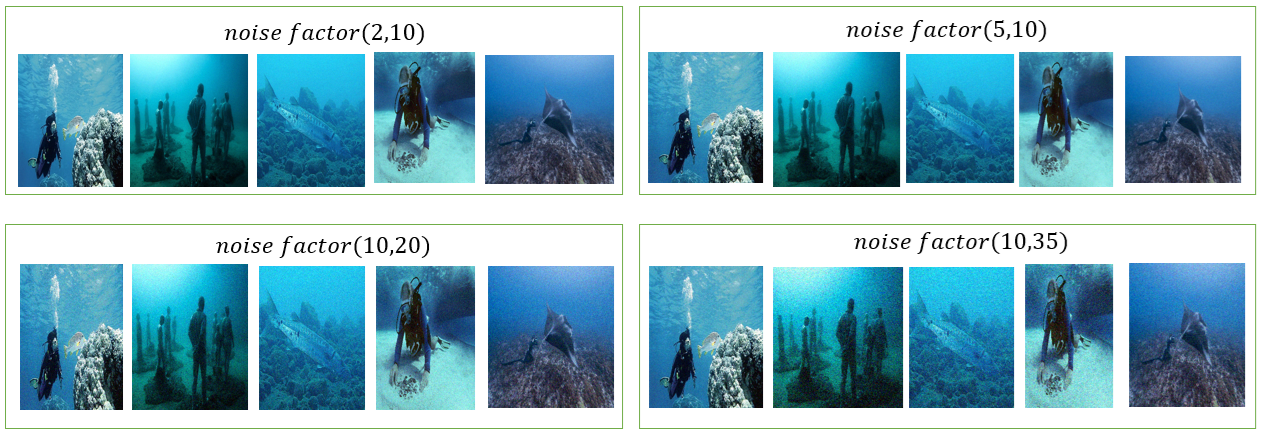}
\includegraphics[width=1.0\columnwidth,height=0.35\columnwidth]{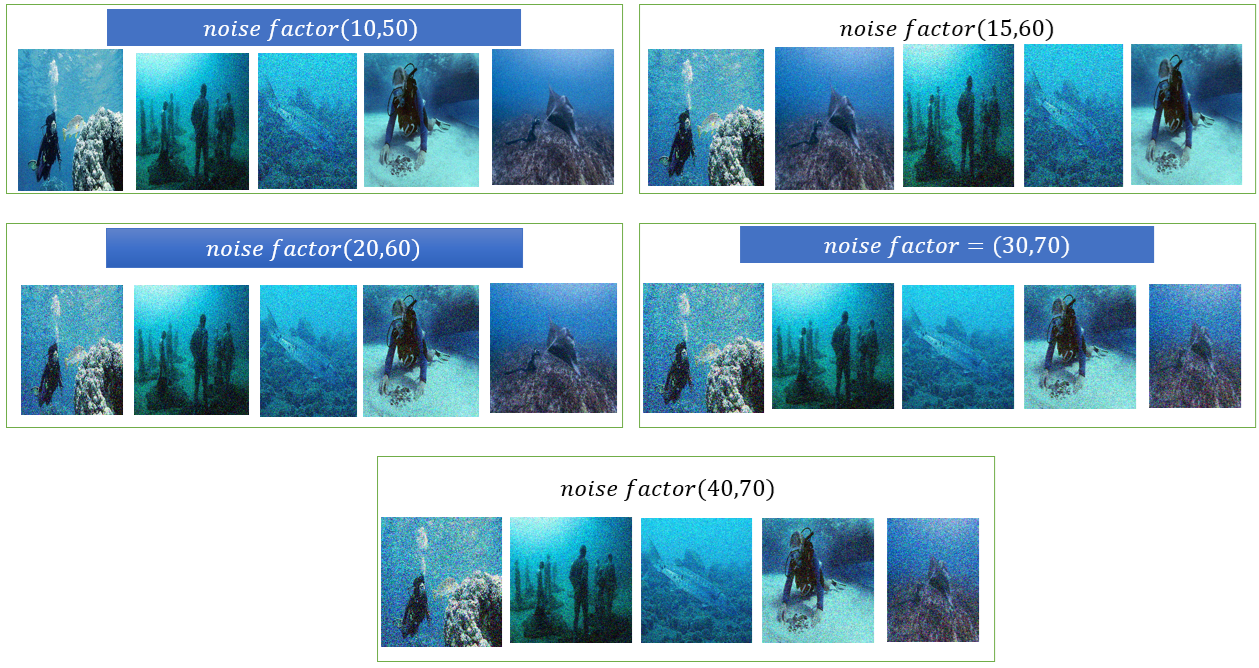}
    \caption{The image exhibits noisy degradation, characterized by the presence of unwanted random variations in pixel intensity.}
    \label{fig:Aba_noisy}
\end{figure}

This equation illustrates how additive Gaussian noise alters the pixel values of the original image, resulting in a degraded image with stochastic fluctuations that mimic real-world imaging artifacts. Adjusting the standard deviation parameter $\sigma$ controls the intensity and spread of the noise, influencing the perceptual quality of the degraded image as shown in Fig \ref{fig:Aba_noisy}.

\vspace{.3cm}


\subsubsection{Color Balance Degradation}
Consider an original image $I$ represented as a three-dimensional array where each pixel contains intensity values for red, green, and blue channels.
\[
\begin{bmatrix}
    R_{11} & G_{11} & B_{11} \\
    R_{12} & G_{12} & B_{12} \\
    \vdots & \vdots & \vdots \\
    R_{mn} & G_{mn} & B_{mn}
\end{bmatrix}
\]
where $R_{ij}$,$G_{ij}$ and $B_{ij}$ represent the intensity values of red, green, and blue channels respectively at pixel (i,j), and m and n denote the dimensions of the image.

\textbf{1. Reddish Tint:}
The reddish tint degradation replicates a deviation in color balance within the image, biasing the color distribution towards red tones. This deviation can occur due to several factors such as environmental lighting, white balance inaccuracies, or sensor characteristics. When a reddish tint afflicts an image, the prominence of red hues intensifies while the contributions of green and blue hues decrease. The blue and green channels are attenuated to induce a reddish tint, while the red channel remains unaltered.
\[
\begin{bmatrix}
    R_{11} & 0 & 0 \\
    R_{12} & 0 & 0 \\
    \vdots & \vdots & \vdots \\
    R_{mn} & 0 & 0
\end{bmatrix}
\]
$R_{ij}$ represents the intensity value of the blue channel at pixel (i,j) and the $I_{\text{reddish}}(x, y, 0)$ as:

\begin{equation}
I_{\text{reddish}}(x, y, 0) = 
\begin{cases}
0, & \text{if } I(x, y, 0) \times (1 - \text{Factor}) + 255 \times \text{Factor} < 0 \\
I(x, y, 0) \times (1 - \text{Factor}) + 255 \times \text{Factor}, & \text{if } 0 \leq I(x, y, 0) \times (1 - \text{Factor}) + 255 \times \text{Factor} \leq 255 \\
255, & \text{if } I(x, y, 0) \times (1 - \text{Factor}) + 255 \times \text{Factor} > 255
\end{cases}
\end{equation}

\begin{align*}
  \forall \, \text{Factor} \sim \cup (a, b); \\
  I_{\text{reddish}}(x, y, 1) = I(x, y, 1);  \\
  I_{\text{reddish}}(x, y, 2) = I(x, y, 2); 
\end{align*}

\begin{figure}[!h]
\centering
\includegraphics[width=1.0\textwidth,height=0.35\textwidth]{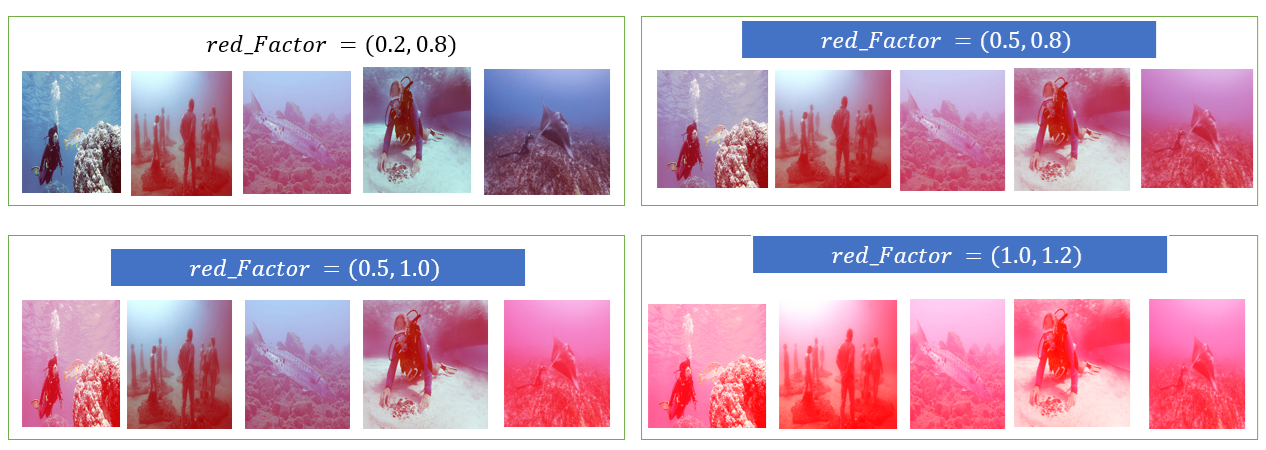}
\caption{Image displaying a reddish tint, the overall color tone of the scene is tinged with red, imparting a warm or rosy hue to the entire image with different variations of factor. }
\label{fig:reddish}
\end{figure}

\textbf{2. Greenish Tint:}
The greenish tint degradation emulates an imbalance in color distribution within the image, favoring green hues. This effect can arise from various factors such as environmental lighting conditions, inaccuracies in white balance, or characteristics of the imaging sensor. When an image is affected by a greenish tint, the intensity of green hues is accentuated while the contributions of red and blue hues diminish. To introduce a greenish tint, the red and blue color channels are suppressed, while the green channel remains unaffected and $I_{\text{greenish}}(x, y, 1)$ is defined as:
\[
\begin{bmatrix}
    0 & G_{11} & 0 \\
    0 & G_{12} & 0 \\
    \vdots & \vdots & \vdots \\
    0 & G_{mn} & 0
\end{bmatrix}
\]
$G_{ij}$ represents the intensity value of the blue channel at pixel (i,j).

\begin{equation}
I_{\text{greenish}}(x, y, 1) = 
\begin{cases}
0, & \text{if } I(x, y, 1) \times (1 - \text{Factor}) + 255 \times \text{Factor} < 0 \\
I(x, y, 1) \times (1 - \text{Factor}) + 255 \times \text{Factor}, & \text{if } 0 \leq I(x, y, 1) \times (1 - \text{Factor}) + 255 \times \text{Factor} \leq 255 \\
255, & \text{if } I(x, y, 1) \times (1 - \text{Factor}) + 255 \times \text{Factor} > 255
\end{cases}
\end{equation}

\begin{align*}
  \forall \, \text{Factor} \sim \cup (a, b); \\
  I_{\text{greenish}}(x, y, 0) = I(x, y, 0); \\
  I_{\text{greenish}}(x, y, 2) = I(x, y, 2);
\end{align*}

\begin{figure}[ht!]
\centering
\includegraphics[width=1.0\textwidth,height=0.35\textwidth]{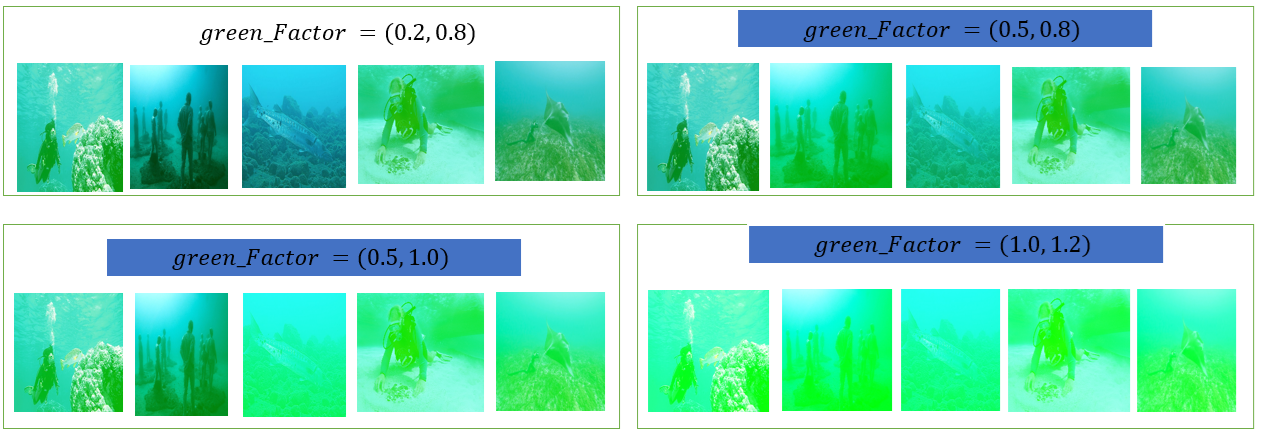}
\caption{The image exhibits a greenish tint, imparting a subtle green hue to the overall color palette with varying $I_{\text{greenish}}(x, y, 1)$.}
\label{fig:greenish}
\end{figure}

\textbf{3. Bluish Tint :}
The bluish tint degradation emulates a color imbalance in the image, skewing the color distribution towards blue hues. This phenomenon can occur due to various factors such as lighting conditions, white balance inaccuracies, or sensor characteristics. When an image is affected by a bluish tint, the intensity of blue color dominance increases while the contribution of red and green colors diminishes. To introduce a bluish tint, the red and green color channels are attenuated, while the blue channel remains unchanged.

\[
\begin{bmatrix}
    0 & 0 & B_{11} \\
    0 & 0 & B_{12} \\
    \vdots & \vdots & \vdots \\
    0 & 0 & B_{mn}
\end{bmatrix}
\]
$B_{ij}$ represents the intensity value of the blue channel at pixel (i,j).

\begin{equation}
I_{\text{bluish}}(x, y, 2) = 
\begin{cases}
0, & \text{if } I(x, y, 2) \times (1 - \text{Factor}) + 255 \times \text{Factor} < 0 \\
I(x, y, 2) \times (1 - \text{Factor}) + 255 \times \text{Factor}, & \text{if } 0 \leq I(x, y, 2) \times (1 - \text{Factor}) + 255 \times \text{Factor} \leq 255 \\
255, & \text{if } I(x, y, 2) \times (1 - \text{Factor}) + 255 \times \text{Factor} > 255
\end{cases}
\end{equation}

\begin{align*}
  \forall \, \text{Factor} \sim \cup (a, b); \\
  I_{\text{bluish}}(x, y, 0) = I(x, y, 0); \\
  I_{\text{bluish}}(x, y, 1) = I(x, y, 1);
\end{align*}

\begin{figure}[ht!]
\centering
\includegraphics[width=1.0\textwidth,height=0.35\textwidth]{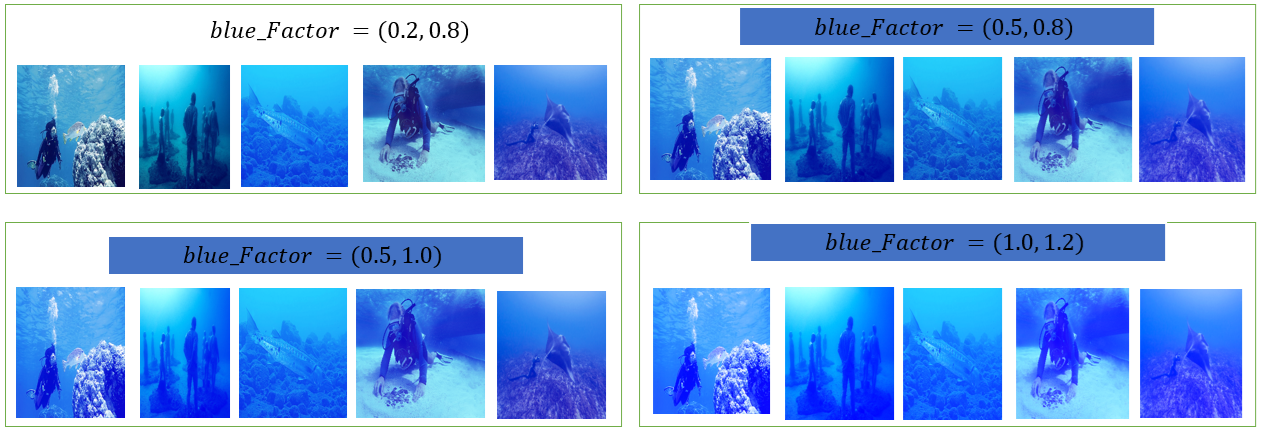}
\caption{Image displaying a bluish tint, noticeable hues of blue saturate the image.}
\label{fig:bluish}
\end{figure}

Here, $I_{CB_{R,G,B}}$ is modified image with an amplified Red, blue, green hue.
This process increases the intensity of Red, blue, green hues across the image, effectively introducing a bluish cast.

\subsection{Dataset Distribution:}
The tables provide statistics for your datasets, UIEB-D8 and EUVP-X-D8, detailing the distribution of images with eight types of degradation across different subcategories.

In the first table, the columns represent the datasets (UIEB and various EUVP subsets) and their respective total image counts. Each type of image degradation—Illumination, Contrast, Hazy, Blurry, Noisy, Reddish/Greenish/Bluish—is split into three subcategories (a, b, and c) for different type of degradation. The total number of images in each dataset is listed at the end.

For the UIEB dataset, which contains 890 images, the images are evenly distributed among the three subcategories (a, b, c) for each degradation type, with approximately 296-298 images per subcategory. The total count for all images across all degradation types is 5340.

The EUVP\_P (U\_Dark) dataset consists of 3138 images, with each subcategory within the degradation types having exactly 1046 images, leading to a total image count of 18,828.

The EUVP\_P (U\_ImageNet) dataset contains 3700 images, with each subcategory for the degradation types having exactly 1233-1234 images, making the total number of images 22,200.

The EUVP\_P (U\_Scenes) dataset has 2185 images, with each subcategory for the degradation types having exactly 728-729 images, resulting in a total of 13,110 images.

The EUVP\_Un dataset includes 3140 images, with each subcategory for the degradation types having 1046-1048 images, totaling 18,840 images.

The second table summarizes the overall counts of referenced and degraded images: there are 13,053 referenced images and 78,318 degraded images. This data shows how images are categorized and distributed among different degradation types and their subcategories within your datasets, helping to understand the distribution and quantity of images available for each type of degradation.

\begin{table}[!h]
\centering
\caption{Dataset statistics for various types of image degradation}
\resizebox{\textwidth}{!}{
\begin{tabular}{|l|ccccccccccccccccccc}
\toprule
Dataset & \multicolumn{3}{c}{Illumination} & \multicolumn{3}{c}{Contrast} & \multicolumn{3}{c}{Hazy} & \multicolumn{3}{c}{Blurry} & \multicolumn{3}{c}{Noisy} & \multicolumn{3}{c}{Reddish/ Greenish/ Bluish} & Total \\
\cmidrule(lr){2-4} \cmidrule(lr){5-7} \cmidrule(lr){8-10} \cmidrule(lr){11-13} \cmidrule(lr){14-16} \cmidrule(lr){17-19}
& a & b & c & a & b & c & a & b & c & a & b & c & a & b & c & a & b & c & \\
\midrule
UIEB (890) & 296 & 296 & 298 & 296 & 296 & 298 & 296 & 296 & 298 & 296 & 296 & 298 & 296 & 296 & 298 & 296 & 296 & 298 & 7120 \\
EUVP\_P (U\_Dark) (3138) & 1046 & 1046 & 1046 & 1046 & 1046 & 1046 & 1046 & 1046 & 1046 & 1046 & 1046 & 1046 & 1046 & 1046 & 1046 & 1046 & 1046 & 1046 & 25,104 \\
EUVP\_P (U\_ImageNet) (3700) & 1233 & 1233 & 1234 & 1233 & 1233 & 1234 & 1233 & 1233 & 1234 & 1233 & 1233 & 1234 & 1233 & 1233 & 1234 & 1233 & 1233 & 1234 & 29,600 \\
EUVP\_P (U\_Scenes) (2185) & 728 & 728 & 729 & 728 & 728 & 729 & 728 & 728 & 729 & 728 & 728 & 729 & 728 & 728 & 729 & 728 & 728 & 729 & 17,480 \\
EUVP\_Un (3140) & 1046 & 1046 & 1048 & 1046 & 1046 & 1048 & 1046 & 1046 & 1048 & 1046 & 1046 & 1048 & 1046 & 1046 & 1048 & 1046 & 1046 & 1048 & 25,120 \\
\bottomrule
\end{tabular}}
\end{table}

This table summarizes the overall counts of referenced and degraded images as shown in Table \ref{tab:totalrd} :
\begin{table}[!h]
\centering
\caption{Total Referenced and Degraded Images}
\begin{tabular}{l r}
\toprule
Category & Count \\
\midrule
Total Referenced Image & 13,053 \\
Total Degraded Image & 1,04,424 \\
\bottomrule
\end{tabular}
\label{tab:totalrd}
\end{table}

\section{Methodology}
The proposed Iterative framework for Degradation Aware Underwater Image Enhancement (IDA-UIE) progressively enhances the input image. In each iteration, a degradation classifier network $\mathbf{\Phi}_{DC}$ identifies the dominant degradation condition in the image. Being \emph{degradation aware}, helps in choosing the corresponding deep network for enhancing the image to remove the effect of degradation. The removal of the present dominant degradation might reveal the efect of another degradation. Thus, the output image is again processed in the next iteration for further enhancement. The process can be continued till (a) the classifier flags the absence of any degradation or (b) a maximum number of iterations are completed. This work uses the second criterion to limit the maximum number of floating point operations. Here, IDA-UIE is operated with a maximum of 3 iterations. The sub-networks used for degradation classification and sub-sequent enhancement are described next.

\subsection{Design of Degradation Classification and Enhancement Networks}
\textbf{Degradation Classification Network --} This network $\mathbf{\Phi}_{DC}$ identifies the category (one of eight classes) of dominant degradation in an underwater image. Additionally, it recognizes the absence of degradation. Thus, it is a $8+1=9$ categtory classifier. It is trained on the UIEB-D8 and EUVP-D8 datasets.
The custom neural network in Fig \ref{fig:1st_ablation} using PyTorch for image classification with a default of 9 classes. The network starts with an initial $3\times3$ convolution layer, followed by two parallel paths: a $1\times1$ convolution and a $3\times3$ convolution. The outputs of these paths are concatenated and passed through another $1\times1$ convolution layer. This result is added to the initial convolution output, similar to a residual connection.

\begin{figure}[!h]
  \centering
  \begin{minipage}[b]{0.55\textwidth}
    \centering
    \includegraphics[width=1.0\columnwidth,height=0.45\columnwidth]{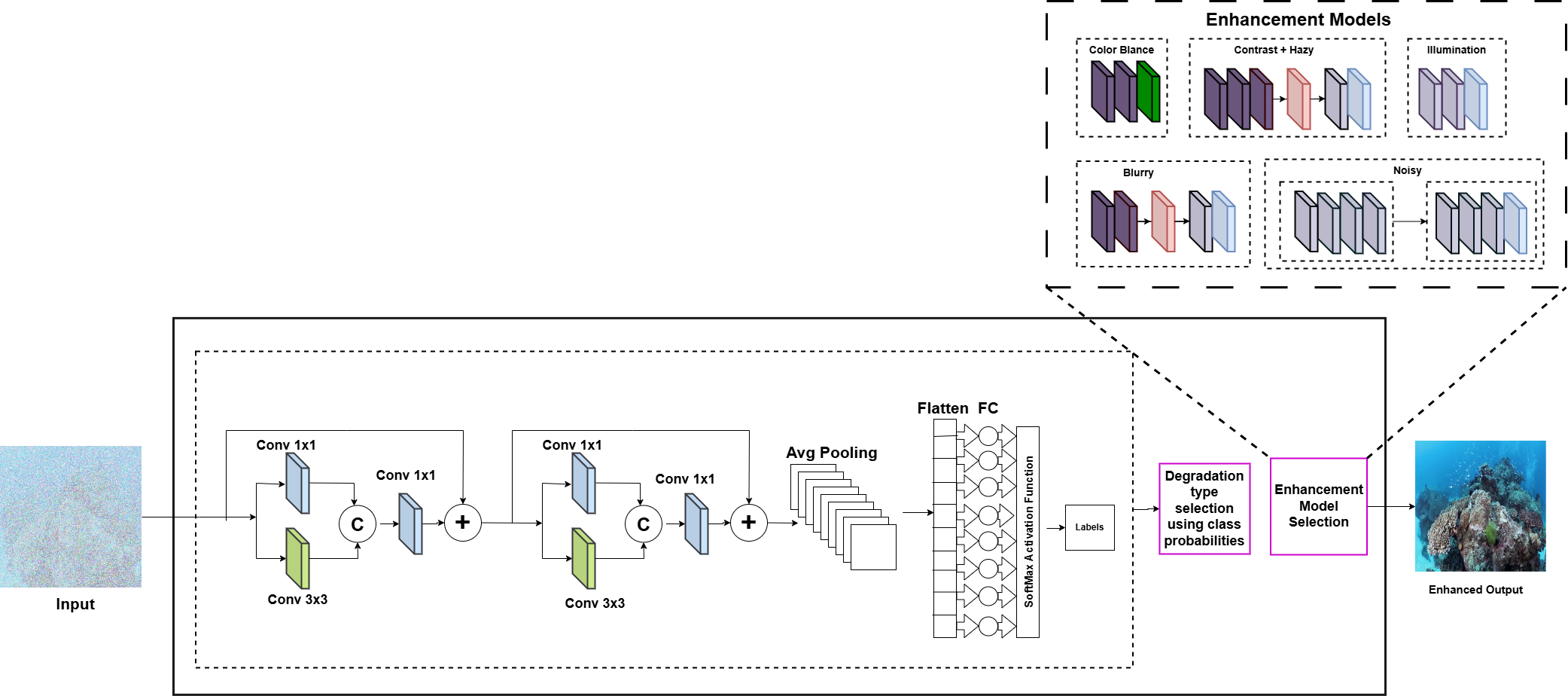}
    \caption{A block diagram of the proposed IDA-UIE model includes a classifier and enhancement model, incorporating degradation selection and enhancement model selection modules. It outputs an enhanced image along with performance metrics.}
    \label{fig:FinalModel}
  \end{minipage}
  \hspace{0.05\textwidth}
  \begin{minipage}[b]{0.35\textwidth}
    \centering
    \resizebox{\textwidth}{!}{%
    \begin{tabular}{|l|c|}
        \toprule
        \textbf{Degradation Type} & \textbf{F1 Score} \\
        \midrule
        No Degradation & 0.9350 \\
        Bluish & 0.9907 \\
        Blurry & 0.9898 \\
        Contrast & 0.9049 \\
        Greenish & 0.9980 \\
        Hazy & 0.9554 \\
        Illumination & 0.9936 \\
        Noisy & 0.9929 \\
        Reddish & 0.9827 \\
        \midrule
        \textbf{GFLOPs} & \textbf{15.1666} \\
        \textbf{Number of Parameters} & \textbf{0.2250 M} \\
        \textbf{Test Accuracy} & \textbf{97.63\%} \\
        \bottomrule
    \end{tabular}}
    \captionof{table}{F1 Scores for different degradation types of the proposed IDA-UIE model}
    \label{tab:f1_scores}
  \end{minipage}
\end{figure}

Next, the network uses weighted average pooling to reduce the feature maps to a $1x1$ spatial dimension. The pooled features are then flattened and passed through a fully connected layer to produce the final classification output. This architecture combines convolutional layers, parallel processing paths, and pooling to effectively extract and classify features from the input image.
The network architecture has a cascade of two modules in Fig \ref{fig:FinalModel}, each containing $1 \times 1$ and $3 \times 3$ convolution kernels in parallel with residual connections. The convolution layer output is flattened and processed by fully connected layers for the final classification. A Winner-Take-All strategy is applied to select the dominant degradation. Accordingly, a suitable deep network is selected for image enhancement. This is an iterative process which checks for different degradations. If no degradation is detected, iteration stops. The network is shown in Figure~\ref{fig:FinalModel} and Table \ref{tab:f1_scores}.

\begin{figure}[!h]
  \centering
  \begin{minipage}[b]{0.55\textwidth}
    \centering
    \includegraphics[width=\textwidth]{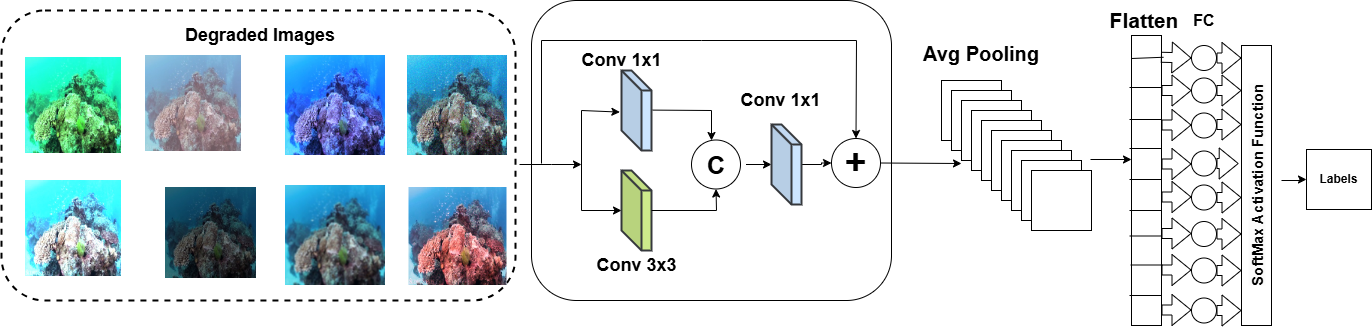} 
    \caption{Degradation Classifier Architecture through Ablation Study}
    \label{fig:1st_ablation}
  \end{minipage}
  \hspace{0.05\textwidth}
  \begin{minipage}[b]{0.35\textwidth}
    \centering
    \resizebox{\textwidth}{!}{%
    \begin{tabular}{|l|c|}
        \toprule
        \textbf{Degradation Type} & \textbf{F1\_Score} \\
        \midrule
        No Degradation & 0.824 \\
        Bluish & 0.757 \\
        Blurry & 0.869 \\
        Contrast & 0.802 \\
        Greenish & 0.924 \\
        Hazy & 0.816 \\
        Illumination & 0.820 \\
        Noisy & 0.843 \\
        Reddish & 0.826 \\
        \midrule
        GFLOPs & 1.7448 \\
        Number of Parameters & 0.0280 \\
        Test Accuracy & 80.14\%  \\
        \bottomrule
    \end{tabular}}
   
    \captionof{table}{F1 Score for different degradation types as described in Ablation.}
    \label{tab:f1score}
  \end{minipage}
\end{figure}

\subsubsection{Ablation Study}
\textbf{Training of Models:}

\begin{figure}[!h]
  \centering
  \begin{minipage}[b]{0.45\textwidth}
    \centering
    \includegraphics[width=\textwidth]{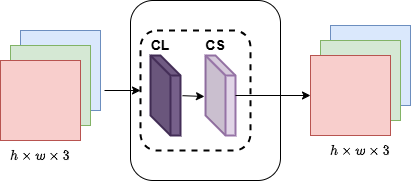} 
    \caption{Model 1: Evolving Network Architecture through Ablation Study}
    \label{fig:1st_ablation}
  \end{minipage}
  \hspace{0.05\textwidth}
  \begin{minipage}[b]{0.45\textwidth}
    \centering
    \resizebox{\textwidth}{!}{%
    \begin{tabular}{|l|c|c|}
        \hline
        \textbf{Degradation Type} & \textbf{MSE} & \textbf{PSNR} \\
        \hline
        Illumination & 0.01 & 24.97 dB \\
        Hazy & 2.06 & 10.85 dB \\
        Blurry & 0.23 & 23.85 dB \\
        Noisy & 0.71 & 18.36 dB \\
        Contrast & 0.501 & 21.86 dB \\
        Color Balance & 0.005 & 29.78 dB \\
        \hline
        GFLOPs: 0.4530 & Number of Parameters: 0.0070 M \\
        \hline
    \end{tabular}}
    \captionof{table}{A table of the enhancement model designed to address degradations through ablation study based on Model 1.}
    \label{tab:1stAblation}
  \end{minipage}
\end{figure}

In the first ablation study (Figure \ref{fig:1st_ablation}), we designed a network constructed using a single convolutional layer, represented as $CL(3\times3\times3@128;1,1)$. This layer aims to extract and assimilate information from the input image. A LeakyReLU activation function follows the initial convolution operation, enhancing the network's ability to capture non-linear features. Here, $CL(m \times n \times k @ q; s, p)$ refers to $q$ number of $m \times n \times k$ convolution kernels with stride $s$ and padding $p$, followed by LeakyReLU. Following this, we employed another layer, $CS(128\times128\times3@3;1,1)$, which involves a convolution followed by a Sigmoid activation to produce the final output. Despite these efforts, the results shown in Table \ref{tab:1stAblation} indicate that this architecture did not yield satisfactory performance.

\begin{figure}[htbp]
  \centering
  \begin{minipage}[b]{0.45\textwidth}
    \centering
    \includegraphics[width=\textwidth]{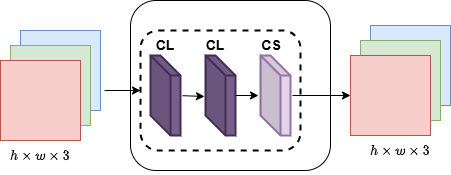} 
    \caption{Model 2: Evolving Network Architecture through Ablation Study. }
    \label{fig:2nd ablation}
  \end{minipage}
  \hspace{0.05\textwidth}
  \begin{minipage}[b]{0.45\textwidth}
    \centering
    \resizebox{\textwidth}{!}{%
     \begin{tabular}{|l|c|c|}
        \toprule
        \textbf{Degradation Type} & \textbf{MSE} & \textbf{PSNR} \\
        \midrule
        Illumination & 0.06 & 22.57 dB \\
        Hazy & 2.81 & 10.95 dB \\
        Blurry & 0.65 & 20.85 dB \\
        Noisy & 0.35 & 22.53 dB \\
        Contrast & 0.767 & 21.51 dB \\
        Color Balance & 0.01 & 25.98 dB \\
        \midrule
        GFLOPs:  20.0068 & Number of Parameters: 0.3057 M \\
        \bottomrule
    \end{tabular}}
    \captionof{table}{A table of the enhancement model designed to address degradations through ablation study based on Model 2.}
    \label{tab:2ndablation}
  \end{minipage}
\end{figure}

For the second ablation study (Figure \ref{fig:2nd ablation}), we expanded the network by incorporating additional convolutional layers, specifically $CL(3\times3\times3@128;1,1)$ and $CL(128\times128\times3@256;1,1)$. These layers were designed to further extract and assimilate information from the input image. As before, a LeakyReLU activation function was applied after each convolution operation to enhance feature capture. Subsequently, we added a layer $CS(256\times256\times3@3;1,1)$, involving a convolution followed by a Sigmoid activation. However, despite these modifications, the results presented in Table \ref{tab:2nd Ablation} still showed that the architecture did not achieve satisfactory performance.

\begin{figure}[htbp]
  \centering
  \begin{minipage}[b]{0.45\textwidth}
    \centering
    \includegraphics[width=\textwidth]{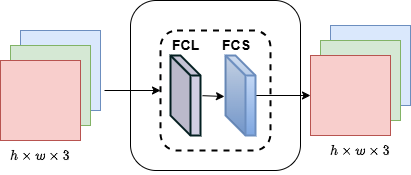} 
    \caption{Model 3: Evolving Network Architecture through Ablation Study }
    \label{fig:3rd ablation}
  \end{minipage}
  \hspace{0.05\textwidth}
  \begin{minipage}[b]{0.45\textwidth}
    \centering
    \resizebox{\textwidth}{!}{%
     \begin{tabular}{|l|c|c|}
        \toprule
        \textbf{Degradation Type} & \textbf{MSE} & \textbf{PSNR} \\
        \midrule
        Illumination & 3.09 & 14.07 dB \\
        Hazy & 3.90 & 14.08 dB \\
        Blurry & 3.90 & 14.08 dB \\
        Noisy & 3.91 & 14.07 dB \\
        Contrast & 3.91 & 14.07 dB \\
        Color Balance & 3.91 & 14.07 dB \\
        \midrule
        GFLOPs  0.0503 & Number of Parameters: 50.52 M \\
        \bottomrule
    \end{tabular}}
    \captionof{table}{A table of the enhancement model designed to address degradation through ablation study based on Model 3.}
    \label{tab:3rdablation}
  \end{minipage}
\end{figure}

In the third ablation study (Figure \ref{fig:3rd ablation}), we explored the use of fully connected layers ($FCL$) encapsulated with LeakyReLU activations. The output from this setup was connected to a fully connected layer with Sigmoid activation ($FCS$) and then unflattened to reconstruct the enhanced image matching the original image dimensions $(h \times w \times 3)$. While fully connected layers are capable of modeling complex, non-linear transformations necessary to enhance various degradations, the results in Table \ref{tab:3rdablation} indicated that this approach also did not yield satisfactory performance.

\begin{figure}[htbp]
  \centering
  \begin{minipage}[b]{0.45\textwidth}
    \centering
    \includegraphics[width=\textwidth]{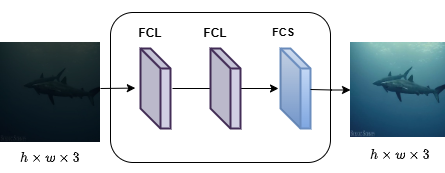} 
    \caption{Model 4: Evolving Network Architecture through Ablation Study }
    \label{fig:illumination}
  \end{minipage}
  \hspace{0.05\textwidth}
  \begin{minipage}[b]{0.45\textwidth}
    \centering
    \resizebox{\textwidth}{!}{%
     \begin{tabular}{|l|c|c|}
        \toprule
        \textbf{Degradation Type} & \textbf{MSE} & \textbf{PSNR} \\
        \midrule
        Color Balance & 0.010 & 17.71 dB \\
        Hazy & 0.017 & 17.08 dB \\
        Blurry & 0.010 & 18.62 dB \\
        Noisy & 0.010 & 17.71 dB \\
        Contrast & 0.014 & 18.28 dB \\
        \textbf{Illumination} & \textbf{5.41e-06} & \textbf{52.66 dB} \\
        \midrule
        GFLOPs:  0.0503 & Number of Parameters: 50.54 M \\
        \bottomrule
    \end{tabular}}
    \captionof{table}{A table of the enhancement model designed to address degradation through ablation study based on Model 4.}
    \label{tab:illumination}
  \end{minipage}
\end{figure}

In the fourth ablation study (Figure \ref{fig:illumination}), we designed a deep network $\mathbf{\Phi}_{IC}$ specifically for enhancing images with low illumination. This network is constructed through a cascade of two fully connected layers ($FCL$) encapsulated with LeakyReLU. The output from this setup is connected to a fully connected layer with Sigmoid activation ($FCS$) and then unflattened to match the original image dimensions $(h \times w \times 3)$. Low illumination correction often involves adjusting the overall brightness and contrast of the image, which can be efficiently learned by fully connected layers as they consider all pixel values at once. Fully connected layers can model complex, non-linear transformations that might be needed to enhance the illumination of the entire image, especially when the correction requires considering the entire image context. This approach has shown good results in illumination correction, as evidenced in Table \ref{tab:illumination}.

The illumination-specific network demonstrated promising results, guiding the development of more advanced network structures that integrate the strengths of fully connected layers.

\begin{figure}[htbp]
  \centering
  \begin{minipage}[b]{0.45\textwidth}
    \centering
    \includegraphics[width=\textwidth]{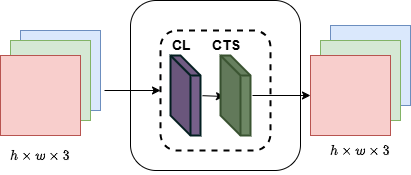} 
    \caption{Model 5: Evolving Network Architecture through Ablation Study }
    \label{fig:4th ablation}
  \end{minipage}
  \hspace{0.05\textwidth}
  \begin{minipage}[b]{0.45\textwidth}
    \centering
    \resizebox{\textwidth}{!}{%
     \begin{tabular}{|l|c|c|}
        \toprule
        \textbf{Degradation Type} & \textbf{MSE} & \textbf{PSNR} \\
        \midrule
        Illumination & 0.04 & 24.57 dB \\
        Hazy & 2.06 & 10.85 dB \\
        Blurry & 0.82 & 20.85 dB \\
        Noisy & 0.23 & 22.36 dB \\
        Contrast & 0.567 & 21.64 dB \\
        Color Balance & 0.06 & 23.28 dB \\
        \midrule
        GFLOPs & 0.4530 & Number of Parameters: 0.0070 M \\
        \bottomrule
    \end{tabular}}
    \captionof{table}{A table of the enhancement model designed to address degradation through ablation study based on Model 5.}
    \label{tab:4th ablation}
  \end{minipage}
\end{figure}

\begin{figure}[htbp]
  \centering
  \begin{minipage}[b]{0.45\textwidth}
    \centering
    \includegraphics[width=\textwidth]{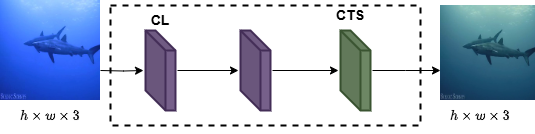} 
    \caption{Model 6: Evolving Network Architecture through Ablation Study }
    \label{fig:colorbalance}
  \end{minipage}
  \hspace{0.05\textwidth}
  \begin{minipage}[b]{0.45\textwidth}
    \centering
    \resizebox{\textwidth}{!}{%
     \begin{tabular}{|l|c|c|}
        \toprule
        \textbf{Degradation Type} & \textbf{MSE} & \textbf{PSNR} \\
        \midrule
        Illumination & 0.05 & 24.27 dB \\
        Hazy & 1.02 & 17.08 dB \\
        Blurry & 0.01 & 22.62 dB \\
        Noisy & 0.019 & 22.35 dB \\
        Contrast & 0.507 & 19.46 dB \\
        \textbf{Color Balance} & \textbf{0.000035} & \textbf{36.45 dB} \\
        \midrule
        GFLOPs & 0.166 & Number of Parameters: 0.4045 M \\
        \bottomrule
    \end{tabular}}
    \captionof{table}{A table of the enhancement model designed to address degradation through ablation study based on Model 6.}
    \label{tab:colorbalance}
  \end{minipage}
\end{figure}

In the fifth ablation study, we constructed a network (Figure \ref{fig:colorbalance}) using a convolutional layer represented as $CL(3\times3\times3@64;1,1)$. This layer aims to extract and assimilate information from the input image. Following the convolution operation, a LeakyReLU activation function is applied. Subsequently, a transposed convolutional layer, denoted as $CTS(64\times64\times3@3;1,1)$, is employed, which includes a Sigmoid activation. Despite these efforts, the results shown in Table \ref{tab:4th ablation} indicate that this architecture did not yield satisfactory performance.

In the sixth ablation study, we further refined the network $\mathbf{\Phi}_{CB_{RGB}}$ (Figure \ref{fig:colorbalance}) to address color imbalances in images. This network is constructed through a cascade of convolutional layers: $CL(3\times3\times3@64;1,1)$ and $CL(3\times3\times64@64;1,1)$. These layers aim to extract and assimilate information from the input image, with a LeakyReLU activation function applied after each convolution operation. Following this, a transposed convolutional layer, $CTS(64\times64\times3@3;1,1)$, is employed, incorporating a Sigmoid activation as shown in Table \ref{tab:colorbalance}.

The $\mathbf{\Phi}_{CB_{RGB}}$ network is particularly well-suited for color correction tasks due to its simple yet effective architecture. It leverages convolutional operations to capture local color relationships and uses non-linear activations to learn complex color mappings. The transposed convolutional layer helps maintain the original image resolution. This end-to-end learning approach makes the network adaptable to various color correction challenges, demonstrating its potential in addressing color imbalances effectively.

Through these ablation studies, we observed the varying efficacy of different architectures in tackling specific image enhancement challenges. While some architectures did not perform satisfactorily, the insights gained guided the refinement and development of more advanced network structures capable of addressing complex underwater image enhancement tasks.

\begin{figure}[htbp]
  \centering
  \begin{minipage}[b]{0.45\textwidth}
    \centering
    \includegraphics[width=\textwidth]{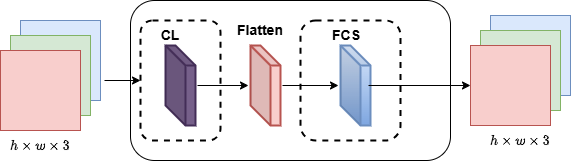} 
    \caption{Model 7: Evolving Network Architecture through Ablation Study }
    \label{fig:6th ablation}
  \end{minipage}
  \hspace{0.05\textwidth}
  \begin{minipage}[b]{0.45\textwidth}
    \centering
    \resizebox{\textwidth}{!}{%
     \begin{tabular}{|l|c|c|}
        \toprule
        \textbf{Degradation Type} & \textbf{MSE} & \textbf{PSNR} \\
        \midrule
        Illumination & 0.02 & 22.43 dB \\
        Hazy & 37.9 & 4.20 dB \\
        Color Balance & 5.79 & 8.79 dB \\
        Noisy & 5.91 & 12.27 dB \\
        Contrast & 38.9 & 4.09 dB \\
        Blurry & 2.58 & 10.88 dB \\
        \midrule
        GFLOPs & 0.088 & Number of Parameters: 72.29 M \\
        \bottomrule
    \end{tabular}}
    \captionof{table}{A table of the enhancement model designed to address degradation through ablation study based on Model 7.}
    \label{tab:6th ablation}
  \end{minipage}
\end{figure}

In the seventh ablation study, the deep network, as shown in Figure \ref{fig:6th ablation}, utilizes an encoder-decoder architecture. The encoder includes convolutional layers $CL(3\times3\times3@32;1,1)$ and one fewer $CL(3\times3\times32@64;2,1)$ layer, reducing the parameter count. After encoding, the feature map is flattened and passed through a fully connected layer with Sigmoid activation ($FCS$), reducing the dimensionality to 100. The decoder reconstructs the image from this latent representation using fully connected layers followed by transposed convolutional layers. Initially, the latent vector (size 100) is expanded to 500 dimensions using $FCS$, and a Sigmoid activation function scales the pixel values between 0 and 1, restoring the image to its original dimensions $(h\times w\times3)$. Despite these efforts, the results, as shown in Table \ref{tab:6th ablation}, indicate unsatisfactory performance.

The advantages of using convolution layers in the encoder include:
\begin{enumerate}
    \item \textbf{Spatial Hierarchies:} The convolutional encoder captures essential spatial features, making it effective for dehazing tasks that rely on understanding spatial dependencies.
    \item \textbf{Compact Representation:} The output of convolution layers represents a high-level, compact representation of the input image, transforming it into a manageable latent space that facilitates efficient reconstruction.
    \item \textbf{Detail Preservation:} Fully connected layers alone were insufficient for capturing the spatial details required for high-quality dehazing and contrast correction, leading to suboptimal performance.
\end{enumerate}

\begin{figure}[htbp]
  \centering
  \begin{minipage}[b]{0.45\textwidth}
    \centering
    \includegraphics[width=\textwidth]{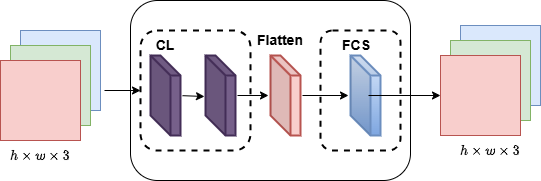} 
    \caption{Model 8: Evolving Network Architecture through Ablation Study }
    \label{fig:7th ablation}
  \end{minipage}
  \hspace{0.05\textwidth}
  \begin{minipage}[b]{0.45\textwidth}
    \centering
    \resizebox{\textwidth}{!}{%
     \begin{tabular}{|l|c|c|}
        \toprule
        \textbf{Degradation Type} & \textbf{MSE} & \textbf{PSNR} \\
        \midrule
        Illumination & 0.064 & 28.32 dB \\
        Hazy & 0.165 & 25.82 dB \\
        Color Balance & 0.100 & 26.98 dB \\
        Noisy & 0.130 & 26.38 dB \\
        Contrast & 0.176 & 25.89 dB \\
        Blurry & 0.071 & 28.42 dB \\
        \midrule
        GFLOPs & 0.138 & Number of Parameters: 46.0913 M \\
        \bottomrule
    \end{tabular}}
    \captionof{table}{A table of the enhancement model designed to address degradation through ablation study based on Model 8.}
    \label{tab:7th ablation}
  \end{minipage}
\end{figure}

In the eighth ablation study, the deep network (Figure \ref{fig:7th ablation} and Table \ref{tab:7th ablation}) employs an encoder-decoder framework. The encoder includes convolutional layers $CL(3\times3\times3@32;1,1)$. This network includes one fewer $CL(3\times3\times32@64;2,1)$ layer, reducing the parameter count. Once the image is encoded, the feature map is flattened and passed through a fully connected layer with Sigmoid activation ($FCS$), reducing the dimensionality to 100. The decoder then reconstructs the image from this latent representation using fully connected layers followed by transposed convolutional layers. The latent vector (size 100) is initially expanded to 500 dimensions using $FCS$, followed by a Sigmoid activation function to scale the pixel values between 0 and 1, restoring the output to the original image dimensions $(h\times w\times3)$. However, the results shown in Table \ref{tab:7th ablation} indicate unsatisfactory performance.

\begin{figure}[htbp]
  \centering
  \begin{minipage}[b]{0.45\textwidth}
    \centering
    \includegraphics[width=\textwidth]{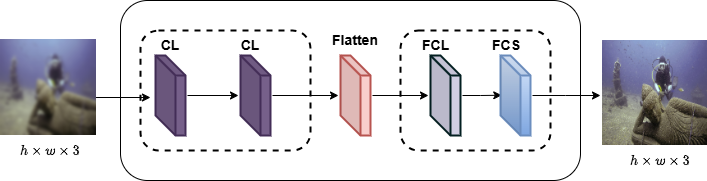} 
    \caption{Model 9: Evolving Network Architecture through Ablation Study }
    \label{fig:Blurry}
  \end{minipage}
  \hspace{0.05\textwidth}
  \begin{minipage}[b]{0.45\textwidth}
    \centering
    \resizebox{\textwidth}{!}{%
     \begin{tabular}{|l|c|c|}
        \toprule
        \textbf{Degradation Type} & \textbf{MSE} & \textbf{PSNR} \\
        \midrule
        Illumination & 0.001 & 38.55 dB \\
        Hazy & 0.00276 & 34.04 dB \\
        Color Balance & 0.0014 & 38.45 dB \\
        Noisy & 0.0002 & 39.05 dB \\
        Contrast & 0.001 & 38.67 dB \\
        \textbf{Blurry} & \textbf{4.06e-05} & \textbf{40.02 dB} \\
        \midrule
        GFLOPs & 0.571 & Number of Parameters: 203.430 M \\
        \bottomrule
    \end{tabular}}
    \captionof{table}{A table of the enhancement model designed to address degradation through ablation study based on Model 9. }
    \label{tab:Blurry}
  \end{minipage}
\end{figure}

In the ninth ablation study, the deep network $\mathbf{\Phi}_{DB}$ (Figure \ref{fig:Blurry} and Table \ref{tab:Blurry}) gives satisfactory result for blurry dataset, realized in an encoder-decoder framework. The encoder includes convolutional layers $CL(3\times3\times3@32;1,1)$. This network includes one fewer $CL(3\times3\times32@64;2,1)$ layer, reducing the parameter count. Once the image is encoded, the feature map is flattened and passed through a fully connected layer ($FCL$), reducing the dimensionality to 100. The decoder then reconstructs the image from this latent representation using fully connected layers followed by transposed convolutional layers. Initially, the latent vector (size 100) is expanded to 500 dimensions using $FCL$, followed by another fully connected layer that adjusts the output to the original image dimensions $(h\times w\times3)$. Finally, a Sigmoid activation function is applied to scale the pixel values between 0 and 1, ensuring the deblurred image maintains proper intensity levels.


\begin{figure}[htbp]
  \centering
  \begin{minipage}[b]{0.45\textwidth}
    \centering
    \includegraphics[width=\textwidth]{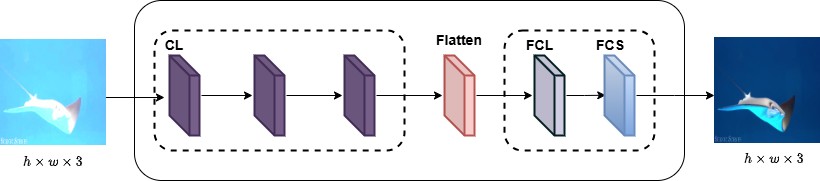} 
    \caption{Model 10: Evolving Network Architecture through Ablation Study }
    \label{fig:contrasthazy}
  \end{minipage}
  \hspace{0.05\textwidth}
  \begin{minipage}[b]{0.45\textwidth}
    \centering
    \resizebox{\textwidth}{!}{%
     \begin{tabular}{|l|c|c|}
        \toprule
        \textbf{Degradation Type} & \textbf{MSE} & \textbf{PSNR} \\
        \midrule
        Color Balance & $3.185 \times 10^{-5}$ & 40.96 dB \\
        Blurry & $2.098 \times 10^{-5}$ & 38.92 dB \\
        Noisy & $2.28 \times 10^{-5}$ & 40.38 dB \\
        Illumination & $2.18 \times 10^{-5}$ & 42.46 dB \\
        \textbf{Contrast} & \textbf{$1.49 \times 10^{-5}$} & \textbf{46.39 dB} \\
       \textbf{ Hazy} & \textbf{$1.16 \times 10^{-5}$} & \textbf{46.03 dB} \\
        \midrule
        GFLOPs & 0.822 & Number of Parameters: 151.07 M \\
        \bottomrule
    \end{tabular}}
    \captionof{table}{A table of the enhancement model designed to address degradation through ablation study based on Model 10.}
    \label{tab:contrasthazy}
  \end{minipage}
\end{figure}

\textbf{Deep Network for Dehazing and Enhancing High-contrast Images -- }Ideally, two networks, $\mathbf{\Phi}_{DH}$ for dehazing and $\mathbf{\Phi}_{CE}$ for contrast enhancement, would be designed to handle these respective tasks. However, the $\mathbf{\Phi}_{DH}$ network often produces noisy images for high contrast and hazy datasets. To address this, we have designed a single network, $\mathbf{\Phi}_{DHCE}$, which performs both dehazing and contrast enhancement (shown in Figure~\ref{fig:contrasthazy} and Table \ref{tab:contrasthazy}).

The model architecture comprises convolutional layers ($CL$) in the encoder. The initial layer is represented as $CL(3 \times 3 \times 3 @ 32;1,1)$, where each convolution ($C$) is paired with LeakyReLU ($L$) for non-linearity. This encoder downsamples the input image, producing a compressed latent representation that retains essential details while eliminating noise. This representation is further refined with two additional layers: $CL(3 \times 3 \times 32 @ 64;1,1)$ and $CL(3 \times 3 \times 64 @ 128;1,1)$, each followed by LeakyReLU activation.

The encoded feature map is then flattened and passed through a fully connected layer ($FCL$), reducing the dimensionality to 100. The decoder reconstructs the dehazed and enhanced image from this latent representation. The decoder includes fully connected layers followed by transposed convolutional layers. Initially, the latent vector (size 100) is expanded to 500 dimensions using $FCL$. This is followed by another fully connected layer, which expands the output to match the original image dimensions $(h \times w \times 3)$. Finally, a sigmoid activation function scales the pixel values between 0 and 1, ensuring a properly reconstructed image that is both dehazed and enhanced.




\begin{figure}[htbp]
  \centering
  \begin{minipage}[b]{0.45\textwidth}
    \centering
    \includegraphics[width=\textwidth]{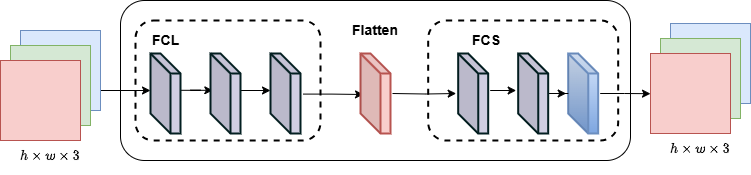} 
    \caption{Model 11: Evolving Network Architecture through Ablation Study }
    \label{fig:8th ablation}
  \end{minipage}
  \hspace{0.05\textwidth}
  \begin{minipage}[b]{0.45\textwidth}
    \centering
    \resizebox{\textwidth}{!}{%
     \begin{tabular}{|l|c|c|}
        \toprule
        \textbf{Degradation Type} & \textbf{MSE} & \textbf{PSNR} \\
        \midrule
        Illumination & 4.14 & 13.82 dB \\
        Hazy & 1.02 & 12.78 dB \\
        Blurry & 0.07 & 23.82 dB \\
        Color Balance & 0.02 & 25.24 dB \\
        Contrast & 0.56 & 19.38 dB \\
        Noisy & 0.00056 & 36.72 dB \\
        \midrule
        GFLOPs & 0.05055 & Number of Parameters: 50.54 M \\
        \bottomrule
    \end{tabular}}
    \captionof{table}{A table of the enhancement model designed to address degradation through ablation study based on Model 11.}
    \label{tab:8th ablation}
  \end{minipage}
\end{figure}

In the eleventh ablation study (Figure \ref{fig:8th ablation} and Table \ref{tab:8th ablation}), the network architecture begins with an encoder that vectorizes the incoming $h \times w \times c$ image. This vectorized image is then sequentially processed through four fully connected layers ($FCL$s). The input dimensionality is progressively reduced: starting from $h \times w \times c$ to 128 neurons in the first layer, then 64 neurons in the second layer, and finally 32 neurons in the third layer, creating a compact latent space representation. The decoder's objective is to reconstruct the image from this compressed latent space, mirroring the encoder's structure by expanding the 32-dimensional latent vector back to the original image size $h \times w \times c$. A Sigmoid activation function is applied to the final output to ensure pixel values are scaled between 0 and 1.

The choice of a network comprising fully connected layers for this task is driven by several key reasons:

\begin{enumerate}
    \item \textbf{Effective Noise Reduction:} Fully connected layers in the encoder-decoder architecture inherently filter out noise during training. They achieve this by learning a compressed representation of the input data that focuses on essential features while disregarding noise, which tends to be less structured and influential in the learning process.
    \item \textbf{Direct Feature Mapping:} Unlike convolutional layers that excel at capturing spatial hierarchies and local features, fully connected layers treat each pixel uniformly across the image. This uniform treatment allows them to effectively learn and map the relationship between noisy input and clean output without heavily relying on spatial dependencies.
    \item \textbf{Compact Representation:} By reducing the dimensionality of the input through sequential linear transformations in the encoder, the model learns to encapsulate relevant image features in a more condensed form. This latent representation tends to minimize noise components, leading to clearer and more refined reconstructions in the decoder.
    \item \textbf{Flexibility and Reconstruction Quality:} The fully connected layers in the decoder enable flexible and nonlinear reconstruction of the denoised image. This capability ensures that the model can generate smooth and visually appealing outputs by effectively filling in missing or distorted information caused by noise.
    \item \textbf{Proven Effectiveness:} Empirical evidence and research in image processing tasks, including denoising, demonstrate that fully connected autoencoders can achieve impressive results. This is also evident from our experiments on the datasets used. They significantly reduce noise levels while preserving important image details, making them a reliable choice for enhancing image quality.
\end{enumerate}

\begin{figure}[htbp]
  \centering
  \begin{minipage}[b]{0.45\textwidth}
    \centering
    \includegraphics[width=\textwidth]{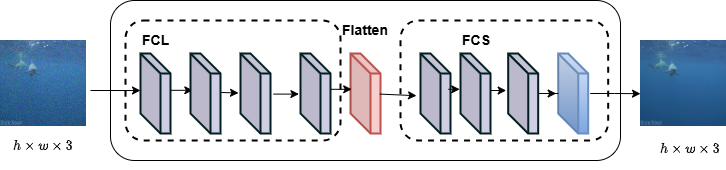} 
    \caption{Model 12: Evolving Network Architecture through Ablation Study }
    \label{fig:Noisy}
  \end{minipage}
  \hspace{0.05\textwidth}
  \begin{minipage}[b]{0.45\textwidth}
    \centering
    \resizebox{\textwidth}{!}{%
     \begin{tabular}{|l|c|c|}
        \toprule
        \textbf{Degradation Type} & \textbf{MSE} & \textbf{PSNR} \\
        \midrule
        Illumination & 4.14 & 13.82 dB \\
        Hazy & 0.08 & 20.78 dB \\
        Blurry & 0.02 & 25.38 dB \\
        Color Balance & 0.02 & 26.44 dB \\
        Contrast & 0.05 & 22.38 dB \\
        \textbf{Noisy} & \textbf{$1.65 \times 10^{-6}$} & \textbf{48.72 dB} \\
        \midrule
        \textbf{GFLOPs} & \textbf{0.05055} & \textbf{Number of Parameters: 50.54 M} \\
        \bottomrule
    \end{tabular}}
    \captionof{table}{A table of the enhancement model designed to address degradation through ablation study based on Model 12.}
    \label{tab:noisy}
  \end{minipage}
\end{figure}

\textbf{Deep Network for Denoising --} The deep network ($\mathbf{\Phi}_{DN}$) (Figure \ref{fig:Noisy} and Table \ref{tab:noisy}) for denoising the data follows a similar architecture. The encoder vectorizes the incoming $h \times w \times c$ image and processes it through a sequence of four $FCL$s. The input dimensionality is progressively reduced: from $h \times w \times c$ to 128 neurons in the first layer, 64 neurons in the second, 32 neurons in the third, and finally 16 neurons in the fourth layer, creating a compact latent space representation. The decoder then reconstructs the image from this compressed latent space, mirroring the encoder's structure by expanding the 16-dimensional latent vector back to the original image size $h \times w \times c$. Finally, a Sigmoid activation function is applied to the output, ensuring the pixel values are scaled between 0 and 1.

\begin{figure}[!h]
    \centering
    \includegraphics[width=\textwidth]{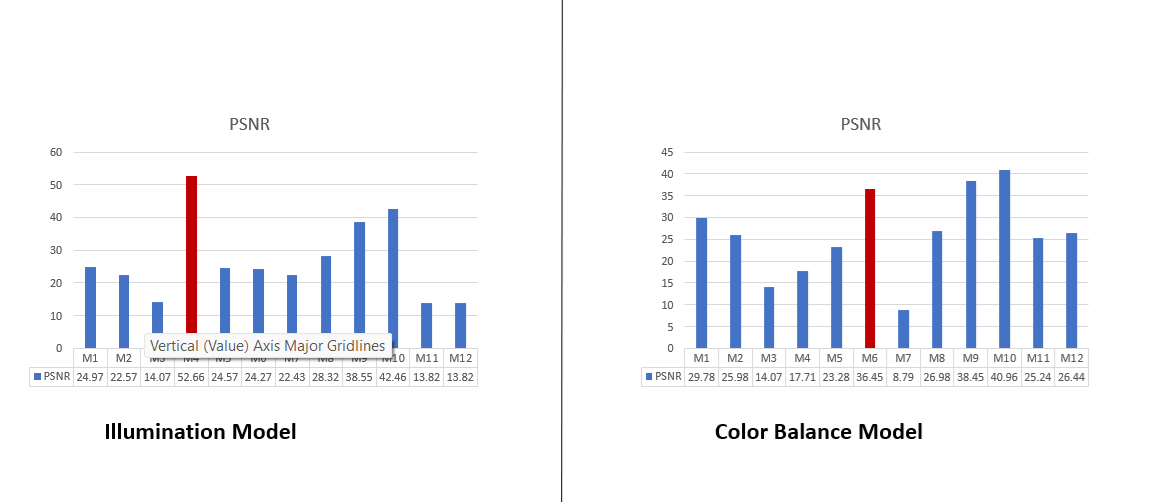}
    \includegraphics[width=\textwidth]{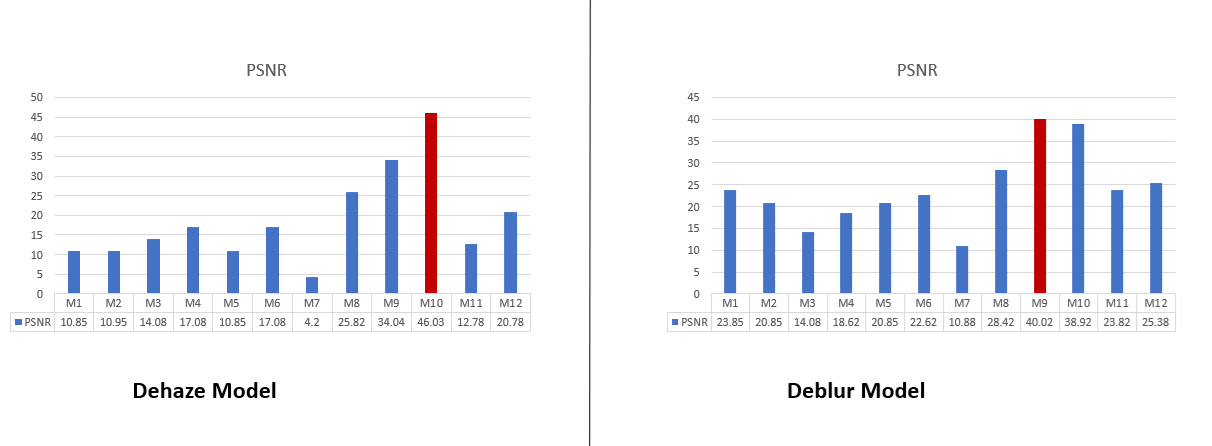}
    \includegraphics[width=\textwidth]{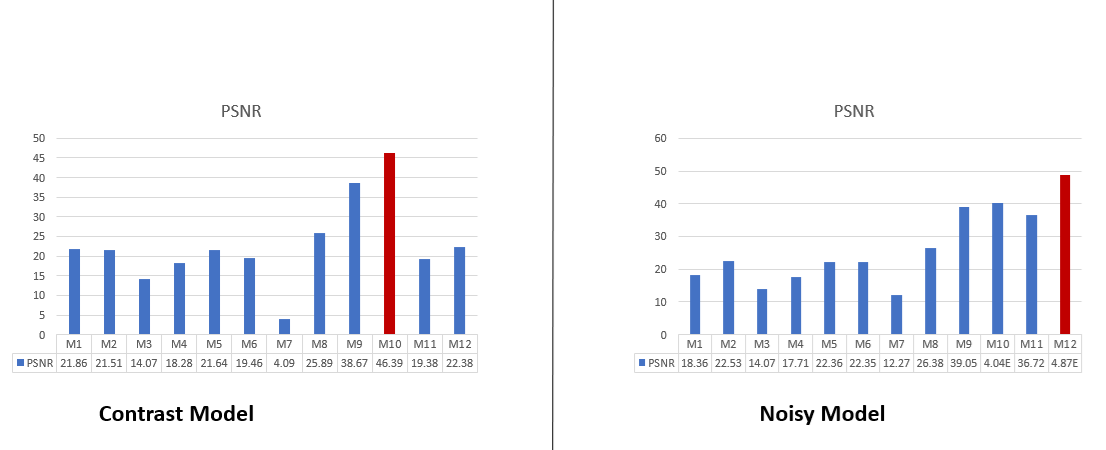}
    \caption{ Selection of the model based on Peak Signal-to-Noise Ratio (PSNR), illustrating how different models perform in terms of image quality enhancement, with higher PSNR values indicating better performance.}
    \label{fig:psnrvsfre}
\end{figure}

This figure Fig. \ref{fig:psnrvsfre} illustrates the selection of models based on their Peak Signal-to-Noise Ratio (PSNR) values. PSNR is a metric used to measure the quality of reconstructed images compared to their original versions, with higher values indicating better performance. The figure compares various models, highlighting how each performs in terms of PSNR, thereby guiding the selection of the most effective model for image quality enhancement.

\section{Experimental Results and Discussion}
\textbf{Baseline Methods --} The proposed approach IDA-UIE is benchmarked on the UIEB~\cite{li2019underwater} and EUVP~\cite{9001231} datasets against nine state-of-art methods. IDA-UIE is compared with WaterNet \cite{8917818}, Fusion-based \cite{6247661}, MSSCE-GAN \cite{10455905}, Deep Wavenet \cite{sharma2023wavelength} on UIEB dataset. IDA-UIE is compared with UGAN \cite{fabbri2018enhancing}, UGAN-P \cite{fabbri2018enhancing}, Funie-GAN \cite{islam2020fast}, Funie-GAN-UP \cite{islam2020fast}, Deep SESR \cite{islam2020simultaneous}, Deep WaveNet \cite{sharma2023wavelength} on EUVP dataset.

\textbf{Evaluation Metrics --} This work has incorporated both reference and reference-less image quality metrics for quantitative performance analysis. The following evaluation metrics are used -- Mean-Squared Error (MSE), Peak Signal-to-Noise Ratio (PSNR), SSIM, Underwater Image Quality Measure (UIQM) \cite{panetta2015human}, Natural Image Quality Evaluator (NIQE) \cite{mittal2012making}, Patch-based Contrast Quality Index (PCQI) \cite{wang2015patch}, Underwater Image Sharpness Measure (UISM) \cite{panetta2015human}, Average Entropy (E), Average Gradient, Underwater Image Contrast Measure (UIConM) \cite{panetta2015human}, and Underwater Color Image Quality Evaluation (UCIQE) \cite{yang2015underwater}. Additionally, the sub-network sizes (parameters in millions) and associated floating point operations (in GFLOPs).

\subsection{Quantitative Performance Analysis --} In terms of qualitative evaluation,  presented the results obtained by applying the aforementioned methods to a single image from the UIEB and EUVP datasets and analyzed the histograms to assess the effects of enhancement. The degradation classifier was trained on the combined UIEB-D8 and EUVP-X-D8 datasets. It achieved an overall accuracy of $97.63\%$. The first performance analysis experiment studies the proportion of images categorized in different dominant degradation conditions (or absence of degradation) for UIEB and EUVP test sets. The results are reported in Table~\ref{tabissue} in terms of percentage for all three iterations. The second experiment evaluated the performance of the individual sub-networks. The image enhancement sub-networks were trained on the combined training subsets of UIEB-D8 and EUVP-UWD-D8 datasets. Their performances were validated on the combined test-sets of UIEB-D8 and EUVP-UWD-D8. The results of this experiment are reported in Table~\ref{DegradationModels}. The network sizes (parameters in millions), floating point operations (in GFLOPs) are reported along with enhancement performance (in terms of MSE and PSNR). The third experiment compares the performance of the proposed model IDA-UIE with four baseline approaches. The results are reported in terms of PSNR and SSIM in Table 3. The fourth experiment presents the comparative performance IDA-UIE and six state-of-art approaches. The results are reported in Table~\ref{EUVP Result} in terms of eleven different evaluation metrics. 

\begin{table}[!h]
\caption{The degradation classifier identifies the necessity of either of illumination correction (IC), contrast enhancement (CE), dehazing (DH), deblurring (DB), denoising (DN), color imbalance correction in red (CBR), green (CBG) or blue (CBG) channel. Additionally, it may also detect the case no further enhancement (NE). The table presents the proportion of images (reported in percentage) from the UIEB and EUVP test sets that are detected for the different kinds of dominant degradation correction (or NE) in all three iterations.}
\label{tabissue}
\resizebox{\textwidth}{!}{
\begin{tabular}{|l|ll|ll|ll|ll|ll|ll|ll|ll|ll|}
\hline
\multirow{2}{*}{}    & \multicolumn{2}{l|}{IC}{} & \multicolumn{2}{l|}{\textbf{CE}} & \multicolumn{2}{l|}{\textbf{DH}} & \multicolumn{2}{l|}{\textbf{DB}} & \multicolumn{2}{l|}{\textbf{DN}} & \multicolumn{2}{l|}{\textbf{CBR}} & \multicolumn{2}{l|}{\textbf{CBG}} & \multicolumn{2}{l|}{\textbf{CBB}} & \multicolumn{2}{l|}{\textbf{NE}}                  \\ \cline{2-19} 
                     & \multicolumn{1}{l|}{\textbf{UIEB}} & \textbf{EUVP} & \multicolumn{1}{l|}{\textbf{UIEB}} & \textbf{EUVP} & \multicolumn{1}{l|}{\textbf{UIEB}} & \textbf{EUVP} & \multicolumn{1}{l|}{\textbf{UIEB}} & \textbf{EUVP} & \multicolumn{1}{l|}{\textbf{UIEB}} & \textbf{EUVP} & \multicolumn{1}{l|}{\textbf{UIEB}}  & \textbf{EUVP} & \multicolumn{1}{l|}{\textbf{UIEB}}  & \textbf{EUVP} & \multicolumn{1}{l|}{\textbf{UIEB}}  & \textbf{EUVP} & \multicolumn{1}{l|}{\textbf{UIEB}}  & \textbf{EUVP}  \\ \hline
\textbf{Iteration 1} & \multicolumn{1}{l|}{11.79\%}           & 12.04\%          & \multicolumn{1}{l|}{11.11\%}           & 11.77\%          & \multicolumn{1}{l|}{9.92\%}           & 11.69\%          & \multicolumn{1}{l|}{10.70\%}           & 11.71\%          & \multicolumn{1}{l|}{10.48\%}           & 11.76\%          & \multicolumn{1}{l|}{10.42\%}            & 12.36\%          & \multicolumn{1}{l|}{10.58\%}            & 11.71\%          & \multicolumn{1}{l|}{10.58\%}            & 11.61\%         & \multicolumn{1}{l|}{14.38\%}  & 5.37\% \\ \hline
\textbf{Iteration 2} & \multicolumn{1}{l|}{11.11\%}           & 12.13\%          & \multicolumn{1}{l|}{11.04\%}           & 11.87\%          & \multicolumn{1}{l|}{10.86\%}           & 11.65\%          & \multicolumn{1}{l|}{10.70\%}           & 11.87\%          & \multicolumn{1}{l|}{10.86\%}           & 12.00\%          & \multicolumn{1}{l|}{11.01\%}            & 8.90\%          & \multicolumn{1}{l|}{10.04\%}            & 11.78\%          & \multicolumn{1}{l|}{10.98\%}            & 11.99\%          & \multicolumn{1}{l|}{13.35\%}  & 4.79\% \\ \hline
\textbf{Iteration 3} & \multicolumn{1}{l|}{10.42\%}           & 18.56\%         & \multicolumn{1}{l|}{10.73\%}           & 11.58\%          & \multicolumn{1}{l|}{9.89\%}           & 11.78\%          & \multicolumn{1}{l|}{10.73\%}           & 11.67\%          & \multicolumn{1}{l|}{11.01\%}           & 11.85\%          & \multicolumn{1}{l|}{10.36\%}            & 11.51\%          & \multicolumn{1}{l|}{10.79\%}            & 11.78\%          & \multicolumn{1}{l|}{10.92\%}            & 11.75\%          & \multicolumn{1}{l|}{15.01\%}  & 6.11\% \\ \hline
\end{tabular}}
\end{table}

\begin{table}[!h]
    \centering
\caption{Individual performance of the image enhancement sub-networks.The network parameters (in millions (M), floating point operations (in GFLOPs) and performance (in terms of MSE and PSNR) are reported.}
\label{DegradationModels}
\begin{tabular}{|l|l|l|l|l|}
\hline
\textbf{Degradation} & \textbf{Parameter(M)} & \textbf{GFLOPs} & \textbf{MSE} & \textbf{PSNR} \\ \hline
Bluish                    & 0.04                  & 0.166            & 0.00058       & 36.78         \\ \hline
Reddish                   & 0.04                  & 0.166            & 0.00052       & 36.44         \\ \hline
Greenish                  & 0.04                  & 0.166            & 0.00009       & 37.67         \\ \hline
Noisy                     & 50.55                & 0.050           & 4.65e-06    & 48.72         \\ \hline
Contrast                  & 151.07                 & 0.822            & 0.00006     & 39.81         \\ \hline
Blurry                    & 203.43                 & 0.571            & 0.00006      & 38.02        \\ \hline
Illumination              & 50.54                 & 0.050            & 4.88e-06       & 49.33         \\ \hline
Hazy                      & 151.07                 & 0.822            & 0.00008       & 40.27         \\ \hline
\end{tabular}
\end{table}

\begin{table}[!h]
   \centering
\caption{Comparison of the proposed model IDA-UIE with four state-of-art approaches on UIEB dataset}
\label{comparision_e}
\begin{tabular}{|l|l|l|l|l|}
\hline
\textbf{Method}  & \textbf{GFLOPs} & \textbf{PSNR}  & \textbf{SSIM} \\ \hline
WaterNet \cite{8917818}    &12.37     & 19.11          & 0.79          \\ \hline
Fusion-based \cite{6247661}  &34.98     & 21.23          & 0.78          \\ \hline
MSSCE-GAN \cite{10455905}  &192      & 21.62          & 0.81          \\ \hline
Deep Wavenet \cite{sharma2023wavelength}  &18.15   & 21.68          & 0.80          \\ \hline
\textbf{IDA-UIE (ours)} & \textbf{16.83} & \textbf{28.87} & \textbf{0.90} \\ \hline
\end{tabular}
\end{table}
\subsection{Qualitative Performance Analysis --} 
\begin{figure}
     \begin{subfigure}[b]{0.48\textwidth}
         \centering
         \includegraphics[width=\linewidth]{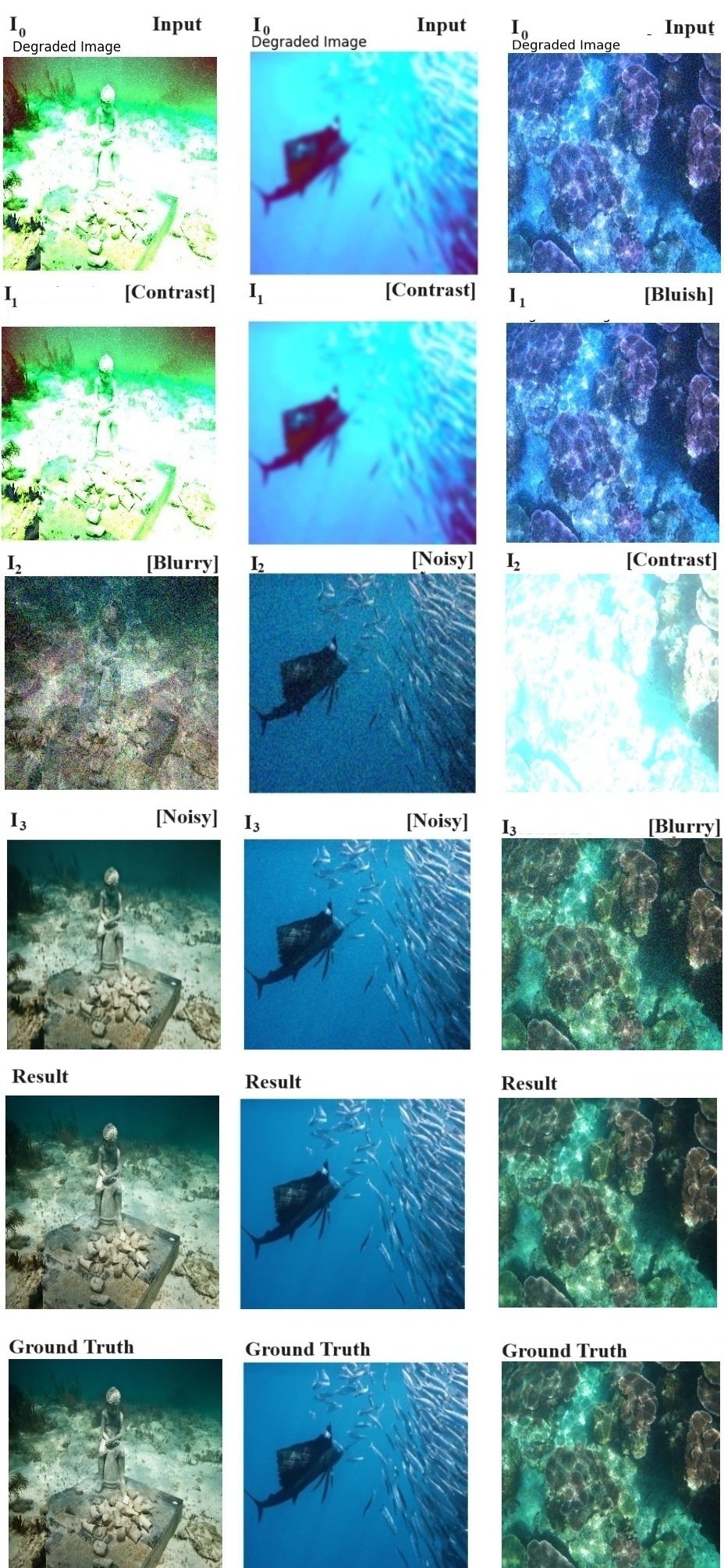}
         \caption{}
         \label{fig:Model}
     \end{subfigure}
     \hfill
     \begin{subfigure}[b]{0.48\textwidth}
         \centering
         \includegraphics[width=\linewidth]{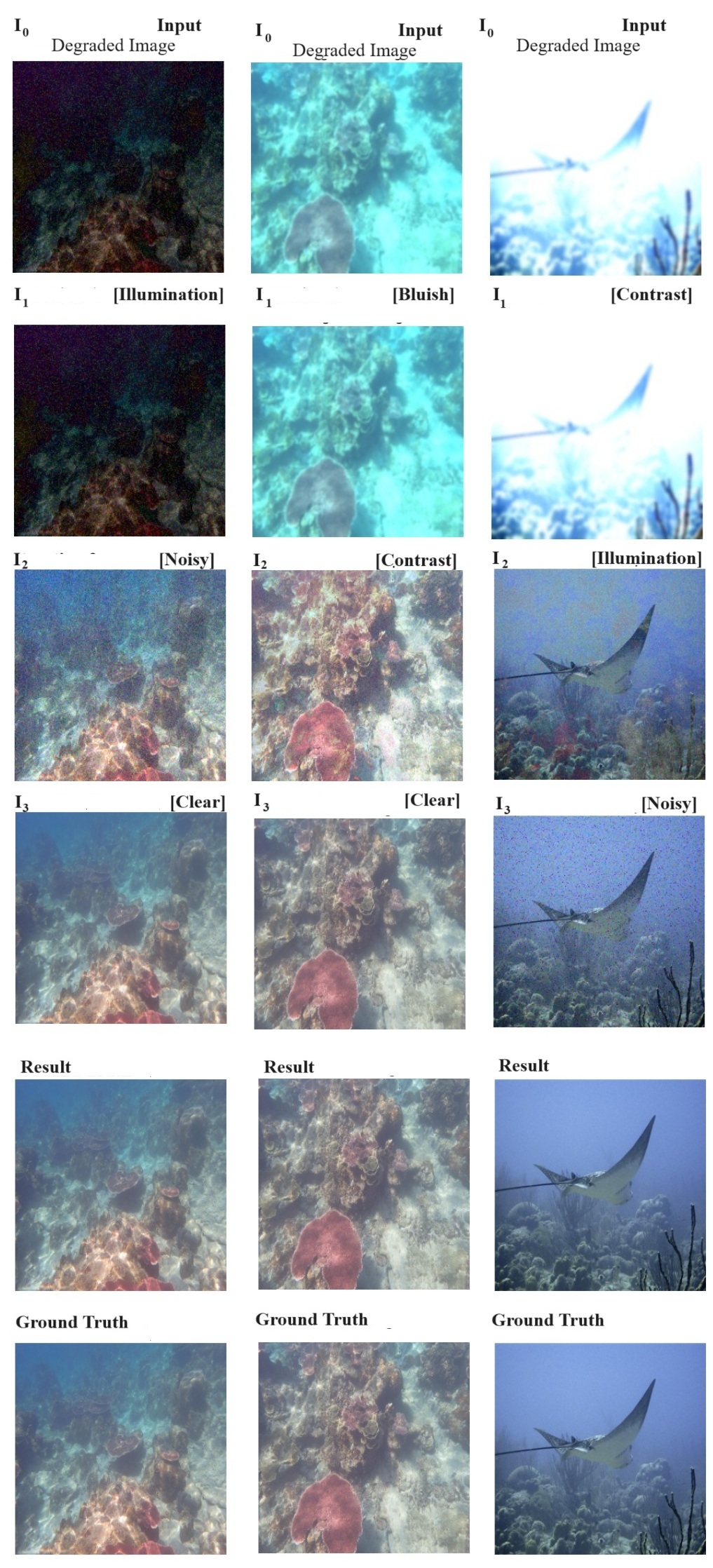}
         \caption{}
         \label{fig:UIEB_S}
     \end{subfigure}
     
        \caption{Each column corresponds to different sample image from the UIEB dataset. The first row shows the input degraded image. This is fed as input to first iteration. The dominant degradation identified in each iteration is shown in square braces. The second and third rows show the images enhanced in respective iterations. The last row shows the ground-truth good quality image and is compared with the output of third iteration (third row).}
        \label{fig:UIEBModel}
\end{figure}

\begin{figure}
     \begin{subfigure}[b]{0.48\textwidth}
         \centering
         \includegraphics[width=\linewidth]{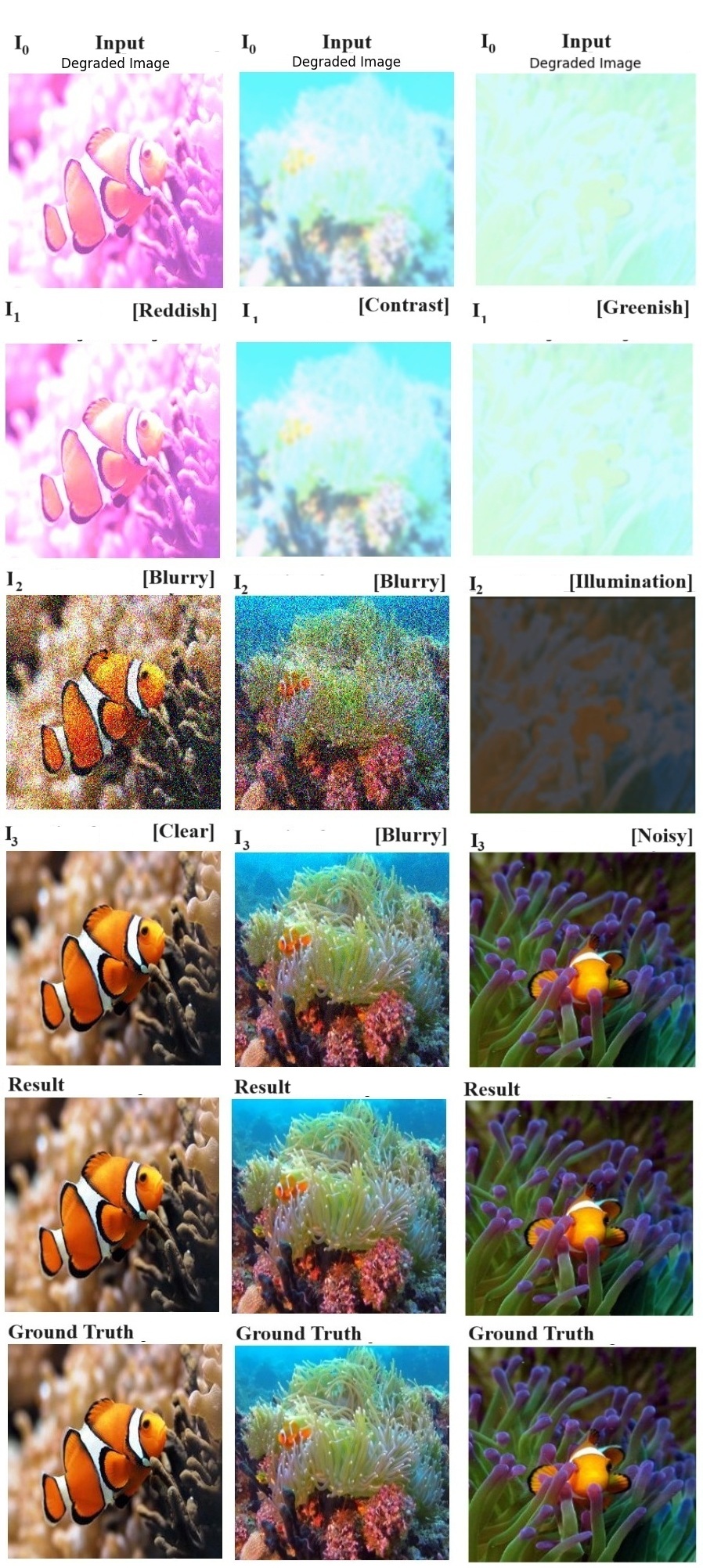}
         \caption{}
         \label{fig:Model}
     \end{subfigure}
     \hfill
     \begin{subfigure}[b]{0.48\textwidth}
         \centering
         \includegraphics[width=\linewidth]{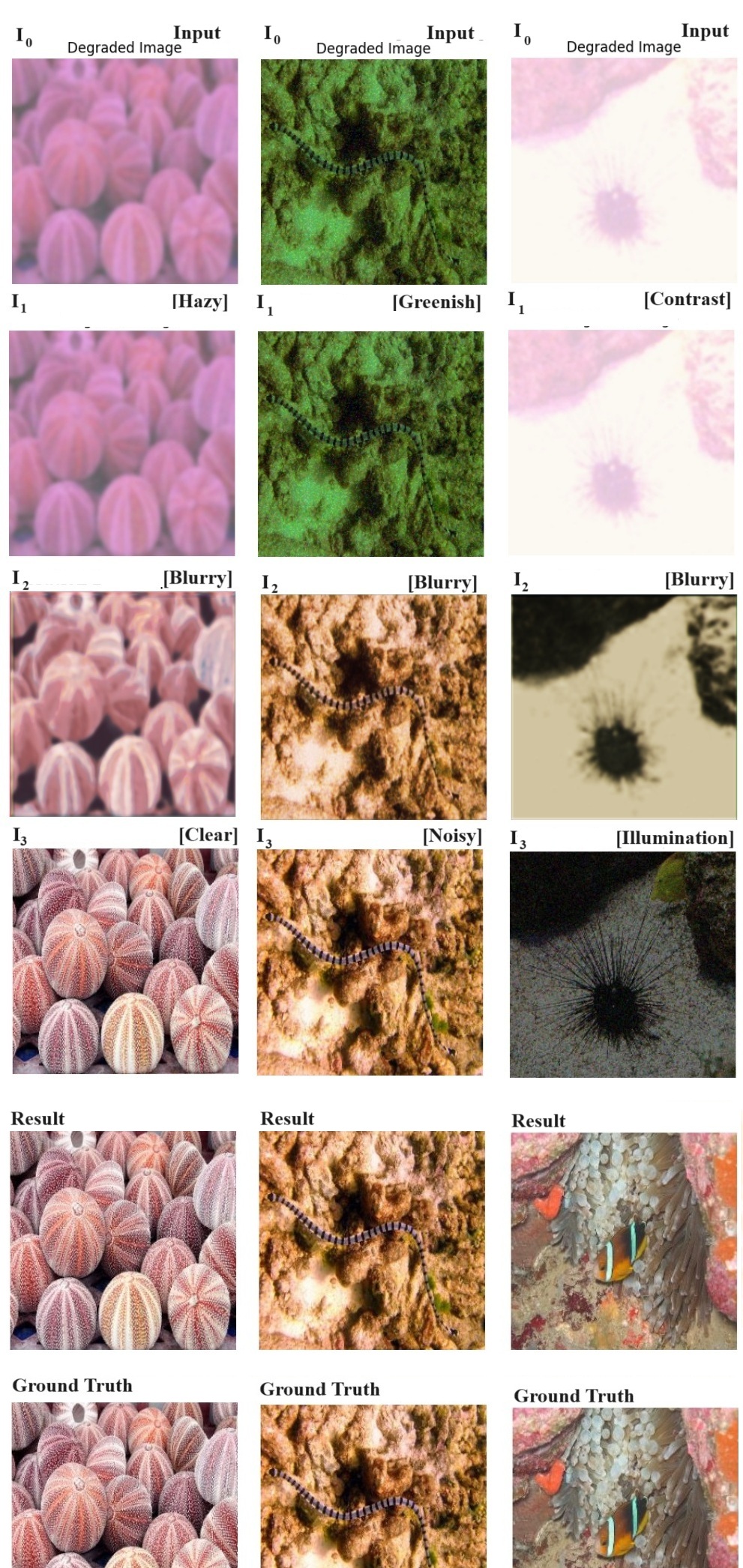}
         \caption{}
         \label{fig:EUVP_S}
     \end{subfigure}
     
        \caption{Each column corresponds to different sample image from the EUVP dataset. The first row shows the input degraded image. This is fed as input to first iteration. The dominant degradation identified in each iteration is shown in square braces. The second and third rows show the images enhanced in respective iterations. The last row shows the ground-truth good quality image and is compared with the output of third iteration (third row).}
        \label{fig:EUVPModel}
\end{figure}

The qualitative performance analysis of the proposed model IDA-UIE are presented in Figures~\ref{fig:UIEBModel} and~\ref{fig:EUVPModel}. Sample images from UIEB and EUVP test-sets are progressively enhanced by correcting the dominant degradations in each iteration. The final output obtained after three iterations is visually compared against the ground-truth good quality image.    

\begin{table}[!h]
\caption{Comparison of the proposed model IDA-UIE with state-of-art approaches on EUVP dataset in terms of different performance metrics.}
\label{EUVP Result}
\resizebox{\textwidth}{!}{
\begin{tabular}{|l|l|l|l|l|l|l|l|l|l|l|l|l|}
\hline
\textbf{Method}  & \textbf{GFLOPs}     & \textbf{MSE} & \textbf{PSNR} & \textbf{SSIM} & \textbf{UIQM} & \textbf{NIQE} & \textbf{PCQI} & \textbf{UISM} & \textbf{Entropy} & \textbf{AG} & \textbf{UIConM} & \textbf{UCIQE} \\ \hline
\textbf{UGAN \cite{fabbri2018enhancing}}   &143      & 0.36         & 26.55         & 0.80          & 2.89          & 49.90         & 0.700          & 6.84          & 7.52             & 7.48        & 0.79            & 0.581          \\ \hline
\textbf{UGAN-P \cite{fabbri2018enhancing}}   &143    & 0.36         & 26.54         & 0.80          & 2.93          & 50.17         & 0.704         & 6.83          & 7.54             & 7.58        & 0.79            & .590           \\ \hline
\textbf{Funie-GAN \cite{islam2020fast}}  &70.34  & 0.39         & 26.22         & 0.79          & 2.97          & 50.51         & 0.706         & 6.90          & 7.55             & 8.58        & 0.84            & 0.590          \\ \hline
\textbf{Funie-GAN-UP \cite{islam2020fast}} &70.34 & 0.60         & 25.22         & 0.78          & 2.93          & 52.87         & 0.702         & 6.86          & 7.80             & 7.80        & 0.79            & 0.588          \\ \hline
\textbf{Deep SESR \cite{islam2020simultaneous}} &30   & 0.34         & 27.08         & 0.80          & 3.09          & 55.68         & 0.679         & 7.06          & 7.40             & 7.57        & 0.78            & 0.572          \\ \hline
\textbf{Deep WaveNet \cite{sharma2023wavelength}}& 18.15 & 0.29         & 28.62         & 0.83          & 3.04          & 44.89         & 0.694         & 7.06          & 7.38             & 7.00        & 0.77            & 0.559          \\ \hline
\textbf{IDA-UIE (ours)}  &\textbf{16.83}       & \textbf{0.0005}       & \textbf{33.75}         & \textbf{0.91}          & \textbf{3.89}          & \textbf{40.34}         & \textbf{0.876}         & \textbf{9.34}          & \textbf{9.45}             & \textbf{8.78}        & \textbf{0.89}            & \textbf{0.784}          \\ \hline
\end{tabular}}
\end{table}

\begin{figure}[hbt!]
    \centering
    \includegraphics[width=1.0\textwidth,height=0.5\textwidth]{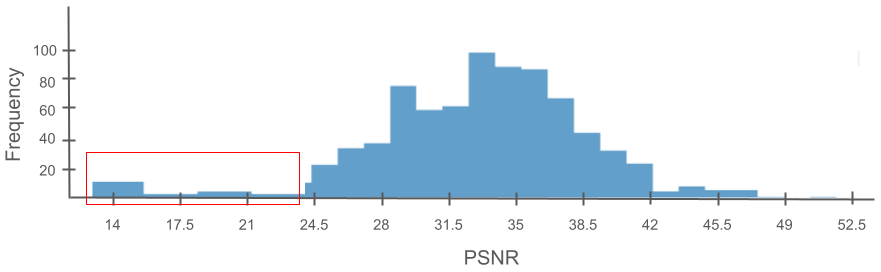}
    \caption{Plot showing PSNR vs Frequency values. The region inside red square depicts the failure cases.}
    \label{fig:psnr}
\end{figure}

The plot Fig \ref{fig:psnr} displays the relationship between Peak Signal-to-Noise Ratio (PSNR) and frequency values, which are used to assess the quality of image enhancement methods. Higher PSNR values typically indicate better image quality. The region inside the red square highlights the failure cases, where the image enhancement method did not perform well. In these instances, the PSNR values are significantly lower, indicating that the enhanced images still contain substantial noise or distortion and thus fail to achieve the desired quality improvements. This analysis helps in identifying specific conditions or frequencies where the enhancement method needs further improvement.
\begin{figure}[hbt!]
    \centering
    \includegraphics[width=1.0\textwidth,height=0.9\textwidth]{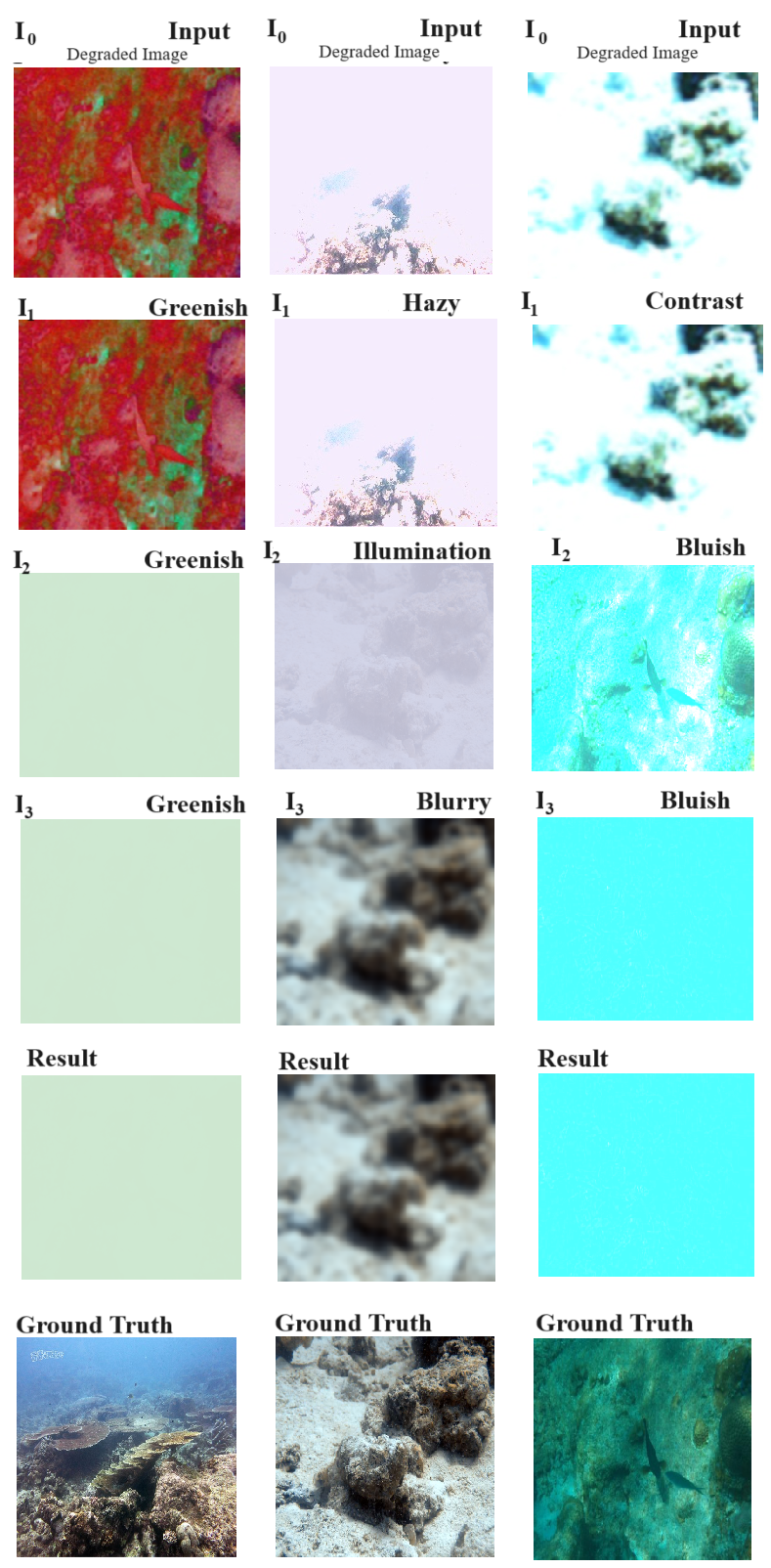}
    \caption{Cascading degradation effect: Initial enhancement failure in iteration 1 adversely impacts subsequent iterations, leading to progressively degraded image quality}
    \label{fig:fail}
\end{figure}

\section{Failure Case}
Due to its severity, the model struggled to eliminate a specific type of degradation. As the enhancement process is sequential, with iteration 2 and iteration 3 depending on the results of iteration 1, any shortcomings in the initial enhancement adversely impact the subsequent iterations. Consequently, the failure to adequately enhance the image in the first iteration propagates through the sequence, leading to progressively degraded results. This issue is illustrated in Figure \ref{fig:fail}, where the cascading effect of the initial enhancement failure is evident in the overall quality of the enhanced images.

\section{Conclusion}
This paper presents an iterative framework for enhancing underwater images with degradation awareness, which identifies and enhances the dominant degradation condition using specific enhancement networks. Unlike single-network approaches, IDA-UIE progressively performs \emph{degradation-aware} enhancements. A classifier identifies one of eight degradation types (including low illumination, low contrast, haziness, blur, noise, and color imbalances), or no degradation, and deploys the corresponding enhancement network. Trained on condition-specific degradations applied to UIEB and EUVP datasets, IDA-UIE outperforms nine state-of-the-art methods on eleven evaluation metrics.

This framework can also be adapted for general image enhancement problems by incorporating condition classifiers and specific enhancement sub-networks, with future research focusing on designing lightweight networks for each component.

\bibliographystyle{plainnat}


\end{document}